\documentclass[journal, compsoc]{IEEEtran}
\usepackage{amsmath,amsfonts}
\usepackage{algorithm}
\usepackage{array}
\usepackage[caption=false,font=normalsize,labelfont=sf,textfont=sf]{subfig}
\usepackage{textcomp}
\usepackage{stfloats}
\usepackage{url}
\usepackage{amsmath,amssymb}
\usepackage{xcolor}
\usepackage{graphicx}
\usepackage{booktabs}
\usepackage{multirow}
\usepackage{tabularx}
\usepackage{array}
\usepackage{pifont}
\usepackage{algpseudocode}
\usepackage{verbatim}
\usepackage{graphicx}
\usepackage[numbers,square,sort&compress]{natbib}
\usepackage{color}
\usepackage{makecell}
\usepackage{bibunits}
\usepackage[colorlinks=true,
            linkcolor=blue,
            citecolor=blue,
            urlcolor=blue]{hyperref}
\newcommand{\eg}{\textit{e.g.}}
\newcommand{\ie}{\textit{i.e.}}
\newcommand{\cmark}{\ding{51}}
\newcommand{\xmark}{\ding{55}}
\usepackage[dvipsnames]{xcolor}
\begin{document}
\title{GeoBridge++: Fact-Guided Geo-Semantic \mbox{Bridging for Unified Cross-View Geo-Localization}}
\author{Zixuan~Song, Jing~Zhang\textsuperscript{\dag},~\IEEEmembership{Senior Member,~IEEE}, Di~Wang\textsuperscript{\dag}, Zhiming~Luo, Wenbin~Liu, Haonan~Guo, En~Wang\textsuperscript{\dag}, Bo~Du\textsuperscript{\dag},~\IEEEmembership{Senior Member,~IEEE}, Liangpei~Zhang,~\IEEEmembership{Fellow,~IEEE}

\thanks{Zixuan~Song is with the College of Computer Science and Technology and Key Laboratory of Symbolic Computation and Knowledge Engineering of Ministry of Education, Jilin University, Changchun 130012, China, and also with Zhongguancun Academy, Beijing 100094, China (e-mail: songzx24@mails.jlu.edu.cn).}

\thanks{Jing~Zhang, Di~Wang, Zhiming~Luo, and Bo~Du are with the School of Computer Science, Wuhan University, Wuhan, China, and also with Zhongguancun Academy, Beijing 100094, China (e-mail: \{jingzhang.cv, d\_wang, thislzm, dubo\}@whu.edu.cn).}
\thanks{Wenbin~Liu and En~Wang are with the College of Computer Science and Technology and Key Laboratory of Symbolic Computation and Knowledge Engineering of Ministry of Education, Jilin University, Changchun 130012, China.
	(e-mail: \{wangen, liuwenbin\}@jlu.edu.cn).}
\thanks{Haonan~Guo and Liangpei~Zhang are with the State Key Laboratory of Information Engineering in Surveying, Mapping and Remote Sensing, Wuhan University, China, and Haonan~Guo is also with Zhongguancun Academy, Beijing 100094, China (e-mail: \{haonan.guo, zlp62\}@whu.edu.cn).}
\thanks{\textsuperscript{\dag}: Corresponding authors.}
}

\markboth{Journal of \LaTeX\ Class Files,~Vol.~14, No.~8, August~2021}%
{Shell \MakeLowercase{\textit{et al.}}: A Sample Article Using IEEEtran.cls for IEEE Journals}

\IEEEtitleabstractindextext{
\rightskip=0pt
\leftskip=0pt
\begin{abstract}
Cross-view geo-localization infers a location by retrieving geo-tagged reference images matching a query image. However, the traditional satellite-centric paradigm limits robustness when high-resolution or up-to-date satellite imagery is unavailable and underexploits complementary cues across views and modalities. To address these challenges, we propose GeoBridge, a novel model that performs bidirectional matching across views and supports language-to-image retrieval. GeoBridge builds on a novel semantic-anchor mechanism that bridges multi-view features through textual descriptions for robust, flexible localization. We further extend GeoBridge to propose GeoBridge++, a fact-guided geo-semantic bridging framework incorporating real-world geographic knowledge to reduce the ambiguity and instability in appearance-dominated supervision. It integrates structured geographic attributes with visual observations to construct factual descriptions and applies targeted guidance based on modality-specific observable content, thereby enhancing geographic discriminability. GeoBridge++ exploits explicit spatial structures encoded by static maps to build a geo-semantic bridge that adaptively aggregates complementary multi-view information and promotes cross-view consistency. In support of this task, we further construct GeoLoc-MM, a million-scale, multi-view, and multi-scale dataset with aligned drone, satellite, street-view, and static-map imagery at six spatial extents per location, enabling systematic evaluation of arbitrary cross-view retrieval, scale robustness, and cross-view generalization. Extensive experiments show that GeoBridge supports robust cross-view and cross-modal geo-localization, while GeoBridge++ achieves consistent improvements across multiple benchmarks. Code and dataset will be released at \url{https://github.com/MiliLab/GeoBridge}.
\end{abstract}

}

\maketitle
\IEEEdisplaynontitleabstractindextext

\section{Introduction}
\IEEEPARstart{C}{ross-view} geo-localization seeks to infer the geographic coordinates of a query image by retrieving visually similar candidates from a database of geo-tagged reference images. By leveraging the rich spatial and semantic cues embedded in imagery, this paradigm enables high-precision localization without external annotations \cite{durgam2024cross, wang2026geoeyes}. It underpins a wide range of applications, including autonomous driving \cite{fervers2023uncertainty,chen2023gnss}, unmanned aerial vehicle (UAV) navigation \cite{wu2025uav, dai2023vision}, geographic information retrieval \cite{xia2024adapting, Transgeo}, disaster monitoring \cite{fan2021design, li2025cross, hypersigma}, and smart city development \cite{sarlin2023orienternet, torii201524, TiMo}. In recent years, cross-view geo-localization has seen growing interest. By learning robust feature mappings across disparate viewpoints, existing methods have substantially improved cross-domain visual correspondence and laid the groundwork for multi-source perception.

\begin{figure*}[!t]
	\begin{center}		\includegraphics[width=0.9\linewidth]{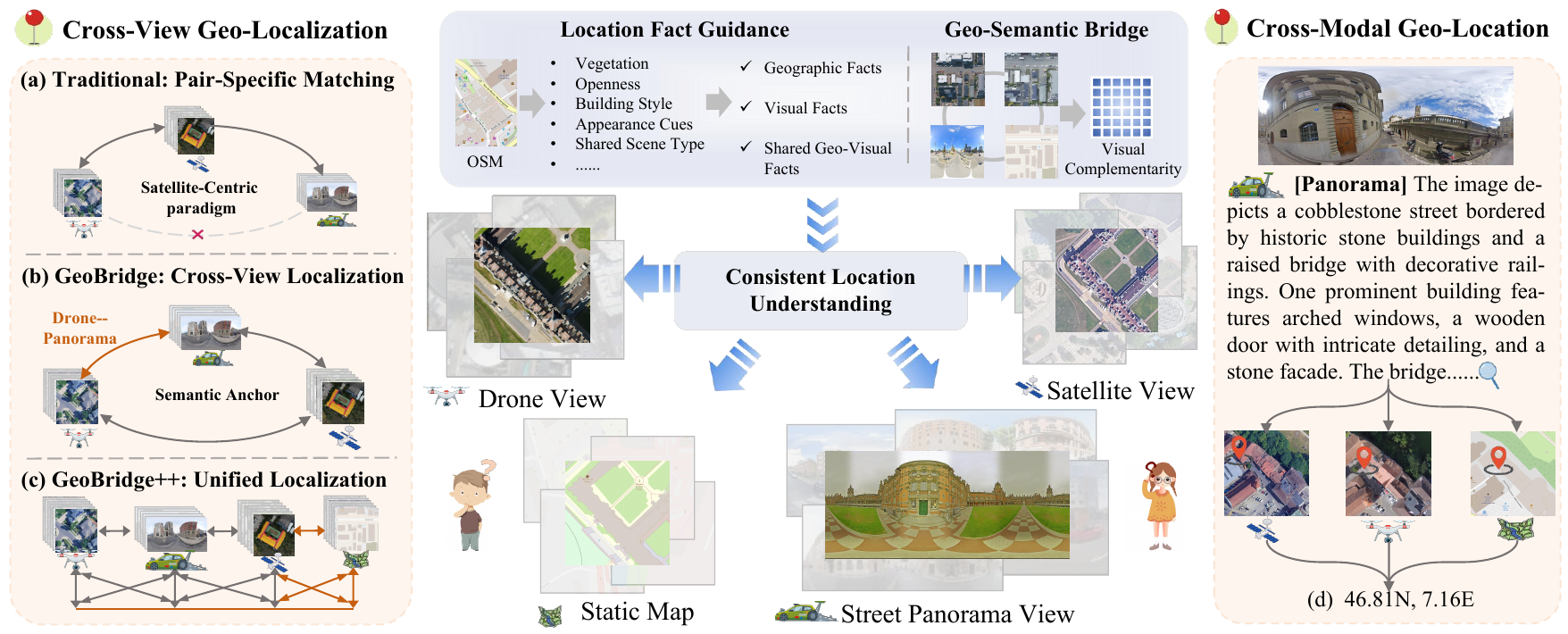}
		\caption{Schematic diagram of GeoBridge and GeoBridge++. Cross-view geo-localization aims to match images with geo-referenced coordinates based on images, while cross-modal geo-localization aims to match images with geo-referenced coordinates based on language descriptions. }
		\label{fig-intro}
	\end{center}
	\vspace{-13pt}
\end{figure*}

Despite this progress, core challenges remain. Owing to the strong heterogeneity between viewpoints, many methods \cite{zheng2020university, Transgeo, wang2021each} adopt the satellite-centric, pair-specific matching strategies for retrieval (see Figure~\ref{fig-intro}a). While effective in certain conditions, this strategy under-utilizes complementary cues across views and becomes fragile when high-resolution or up-to-date satellite imagery is unavailable. An integrated framework that supports arbitrary matching across multiple views can offer alternative localization pathways and markedly improve robustness and generalization under severe viewpoint changes or missing information. Consequently, developing a unified multi-view modeling framework (see Figure~\ref{fig-intro}b and~\ref{fig-intro}c) is crucial for overcoming the limitations of the “satellite-centric” localization and fully exploiting cross-view complementarity.

Meanwhile, rapid advances in large language models (LLMs) open new opportunities for geo-localization \cite{roller2012supervised, li2024georeasoner}. Language naturally encodes rich semantics and spatial knowledge \cite{wang2026text}, as people describe places with directions, landmarks, and topological relations \cite{wang2025geozero}. In practice, language-based rapid localization is especially valuable in low-bandwidth or sensor-constrained settings (\eg, frontline rescue, remote areas, privacy-sensitive scenarios), where text is easier to capture and transmit than images. However, most existing studies \cite{ye2025cross, chu2024towards} overlook the complementarity between language and vision, focusing on single-view description or retrieval. This often leads to semantic hallucinations and spatial inconsistencies. There is therefore a clear need for models that tightly couple linguistic signals with multi-view visual inputs, enhancing spatial understanding and cross-modal alignment so that locations can be inferred robustly from both descriptions and images.

To address these challenges, we propose GeoBridge~\cite{song2026geobridge}, a semantic-anchored, multi-view model for geo-localization. During training, GeoBridge distills drone, street-view, and satellite imagery into a unified, location- and view-aware textual description that serves as a cross-modal semantic bridge. This bridge simultaneously aligns text and each visual view to ensure cross-modal consistency, and aligns the visual views to reinforce cross-view coherence. At inference, the text branch is optional: the model can directly match any pair of drone, street-view, or satellite images, while also supporting language-to-image localization when text is available. By explicitly bridging semantic and visual spaces across views, GeoBridge moves the field beyond reliance on satellite imagery alone. It systematically enables drone-street-view matching, a long-standing gap with clear practical value for UAV navigation, emergency response, and low-altitude logistics. It naturally extends to broader air–ground multi-sensor fusion for closed-loop localization, particularly when satellite data are missing or outdated and drone imagery provides precise coordinates. 

Although GeoBridge establishes an effective semantic bridge across different views, its semantic descriptions remain primarily appearance-driven. Such appearance-driven descriptions mainly characterize visual attributes, including building structures, vegetation distribution, road surfaces, and scene layouts. However, these attributes are highly sensitive to viewpoint, scale, occlusion, seasonal changes, and imaging conditions, which may lead to incomplete or even inaccurate representations of a location. Meanwhile, different locations may exhibit similar visual appearances, introducing semantic ambiguity and reducing location discriminability. Conversely, the same location may appear substantially different across viewpoints and observation conditions, making cross-view semantic consistency difficult to preserve. Excessive reliance on visual appearance may further bias the model toward viewpoint-dependent or environment-dependent cues, thereby weakening its generalization to complex scenes, large viewpoint variations, and diverse geographic observations. This problem becomes more pronounced when unified geo-localization is extended to geographic observations with different information content, spatial range, and representation form. This raises a fundamental question: how can we move beyond limited appearance cues and enable views to form a consistent understanding of a location, thereby faithfully capturing its intrinsic properties while remaining stable under viewpoint and environmental variations.

In light of these limitations, we extend GeoBridge by incorporating real-world geographic information from OpenStreetMap (OSM)~\footnote{\url{https://openstreetmap.org}}, a global geospatial database, together with static maps. These sources provide more stable, objective, and reliable location cues beyond visual appearance. OSM represents physical-world geographic attributes in a structured form, including road categories, building footprints, land-use types, public facilities, natural regions, and waterways. Compared with appearance cues, these attributes are less sensitive to viewpoint and environmental variations, providing complementary structural and functional evidence for distinguishing visually similar locations. In addition, static maps offer a complementary geographic observation by explicitly depicting roads, buildings, functional regions, and their spatial relationships through standardized symbols. Their symbolic representation provides a stable structural reference across heterogeneous observations and naturally connects structured geographic knowledge with real-world visual scenes. Motivated by these properties, we introduce static maps as a new observation modality and extend unified geo-localization to drone, satellite, street-view, and static-map imagery.

Specifically, we propose GeoBridge++, a fact-guided geo-semantic bridging framework for unified cross-view geo-localization, as illustrated in Figure~\ref{fig-intro}. We introduce location facts by integrating structured geographic attributes with real-world visual observations, providing stable and location-discriminative descriptions. Since different modalities vary in observable content and semantic expressiveness, targeted fact guidance is applied according to their respective observable scopes, allowing each modality to preserve its observational characteristics while learning geographically discriminative representations. In parallel, we construct a geo-semantic bridge that adaptively aggregates multi-view representations into a shared latent representation of the location. With static maps introduced as an additional view, the bridge combines their explicit spatial structures with complementary visual evidence from drone, satellite, and street-view observations. This shared representation coordinates the four heterogeneous views without exhaustive pairwise alignment, reducing view-specific appearance bias, redundant supervision, and optimization conflicts while improving scalability. Consequently, cross-view correspondence is established not only through visual similarity, but also through consistent geographic entities, spatial structures, and location facts, enabling unified geo-localization across arbitrary views.

To support this setting, we further construct GeoLoc-MM, a million-scale, multi-view, and multi-scale dataset collected from diverse geographic regions across 36 countries. GeoLoc-MM provides geographically aligned drone, satellite, street-view, and static-map observations, supporting any-view-to-any-view retrieval across all four views. Each location is observed at six distinct spatial extents, capturing substantial appearance and structural variations and enabling comprehensive evaluation of scale robustness and cross-view generalization.

In summary, our contributions are as follows:

\begin{enumerate}

    \item We propose GeoBridge++, a fact-guided geo-semantic bridging framework that moves unified cross-view geo-localization beyond appearance-driven matching by leveraging stable geographic knowledge, improving cross-view consistency and geographic discriminability across diverse views.

    \item We construct GeoLoc-MM, a million-scale, multi-view, and multi-scale geo-localization dataset covering diverse geographic regions across 36 countries. It provides geographically aligned drone, satellite, street-view, and static-map observations at six spatial extents, together with corresponding OSM attributes and factual descriptions, providing a richer resource for unified geo-localization.

    \item We conduct comprehensive experiments on GeoLoc-MM and other public datasets across diverse cross-view and cross-modal geo-localization settings. Extensive results show that GeoBridge++ consistently improves upon competitive methods across different retrieval directions, viewpoints, and scales, while extensive ablation studies validate the effectiveness of the proposed mechanisms.
\end{enumerate}

\section{Related Work}\label{related work}

\subsection{Cross-View Geo-Localization Method}

Cross-view geo-localization aims to establish geographic correspondence between observations captured from substantially different viewpoints. Early studies mainly focus on ground-to-satellite and drone-to-satellite localization, while recent methods further improve cross-view matching through metric learning, spatial alignment, and Transformer-based architectures ~\cite{Transgeo, dai2021transformer, li2024geoformer, lv2024direction, regmi2019bridging, toker2021coming, Sample4geo}. Despite their effectiveness, most existing approaches are developed for predefined view pairs and commonly treat satellite imagery as the central geographic reference, limiting flexibility in exploiting complementary information from multiple observations. These limitations have motivated a transition toward unified multi-view geo-localization. GeoBridge takes a step in this direction by modeling drone, satellite, and street-view observations within a shared retrieval space. GeoBridge++ further extends this setting to static maps and investigates unified geo-localization across heterogeneous visual and symbolic geographic observations.

Beyond visual matching, semantic information provides complementary supervision for learning more stable geographic representations. Recent studies have introduced language into geo-localization, place recognition, and geographic reasoning, either as auxiliary supervision or for cross-modal retrieval~\cite{chu2024towards, ye2025cross, liu2026radar, yuan2021florence, jia2021scaling, li2022blip,wu2026vision, li2023blip}. However, such semantic guidance mainly reflects visible objects, scene categories, and appearance-related attributes, with limited access to intrinsic geographic properties. GeoBridge alleviates this limitation through a shared textual semantic anchor. Since OSM-derived geographic knowledge provides persistent location-specific attributes, GeoBridge++ further combines it with visual evidence to construct factual supervision, providing more stable, cross-view consistent, and geographically discriminative semantic guidance.

\subsection{Cross-View Geo-Localization Datasets}

Datasets play a crucial role in advancing cross-view geo-localization. Existing benchmarks have supported various retrieval settings, including ground-to-aerial, drone-to-satellite, and street-view-to-satellite localization~\cite{CVUSA, CVACT, vo2016localizing, zhu2021vigor, tian2017cross, xu2024uav, zheng2020university}. These datasets provide valuable testbeds for learning robust and viewpoint-invariant representations across visual domains. However, most of them are limited to two-view settings and are usually centered on satellite references, making them less suitable for studying arbitrary-view matching, cross-modal alignment, and multi-source geographic learning. GeoLoc addresses part of this gap by providing aligned drone, street-view, satellite images, and textual descriptions across diverse geographic regions.

More recent datasets have expanded geographic diversity and annotation richness~\cite{zhu2023sues,ye2025cross,dai2023vision}, with several benchmarks further incorporating multi-scale observations to cover variations in spatial extent. In parallel, static maps have been explored as symbolic geographic references, representing roads, blocks, water bodies, green spaces, and points of interest~\cite{zhou2021efficient,mai2022review}. CVGlobal~\cite{ye2024cross} further incorporates aligned map data for street-view-to-map retrieval. Existing benchmarks nevertheless treat multi-scale imagery, map observations, and multi-view retrieval largely as separate settings. GeoLoc-MM unifies these dimensions by aligning drone, satellite, street-view, and static-map observations at six spatial extents, supporting systematic any-view-to-any-view evaluation across views and scales.

\subsection{Comparison to the conference version}

A preliminary version of this paper is presented in~\cite{song2026geobridge}, which introduces the GeoBridge framework and the GeoLoc dataset. This paper substantially extends the previous study in three major aspects. 1) GeoBridge mainly relies on unified textual descriptions generated from multi-view visual observations as semantic anchors, with supervision largely derived from visual content. We further incorporate real-world physical attributes from OSM and propose GeoBridge++, a location fact-guided geo-semantic bridging framework. By incorporating stable geographic knowledge and adaptively coordinating heterogeneous views through a shared location-level representation, GeoBridge++ moves beyond appearance-driven supervision while avoiding exhaustive pairwise alignment across heterogeneous modalities. 2) GeoLoc contains geographically aligned drone, street-view, and satellite observations from 36 countries, together with location descriptions. We further construct GeoLoc-MM by adding static maps, six spatial extents, OSM attributes, and factual descriptions, resulting in a million-scale dataset for cross-view and multi-scale geo-localization evaluation. 3) GeoBridge was evaluated under three-view cross-view and cross-modal retrieval settings. We further extend the task to static maps and broaden the evaluation to four-view retrieval, map-related localization, multi-scale settings, and additional public benchmarks, together with more comprehensive ablation studies and qualitative analyses.

\section{Method}\label{method}

\subsection{Problem Formulation}
We consider four imagery perspectives—drone (\(d\)), satellite (\(s\)), street-view panoramas (\(t\)), and static maps (\(m\)). We aim to achieve large-scale cross-view retrieval for geo-localization: given a query image from any perspective $u\!\in\!\{d,s,t,m\}$, retrieve co-located references from any other coordinate-bearing view $w\!\neq\!u$ and use their coordinates as the location estimate. Departing from the conventional satellite-centric paradigm, we emphasize closed-loop retrieval, where a query from any perspective can be localized via any other coordinate-bearing perspective. GeoBridge operates on the three-view subset \(\mathcal{V}_{\mathrm{G}}=\{d,s,t\}\), whereas GeoBridge++ extends the unified localization setting to the complete four-view set \(\mathcal{V}=\{d,s,t,m\}\). They also support cross-modal retrieval from single-view textual descriptions, enabling localization across diverse scenarios.

\subsection{GeoBridge Method}
\begin{figure}[!t]
	\begin{center}		\includegraphics[width=1\linewidth]{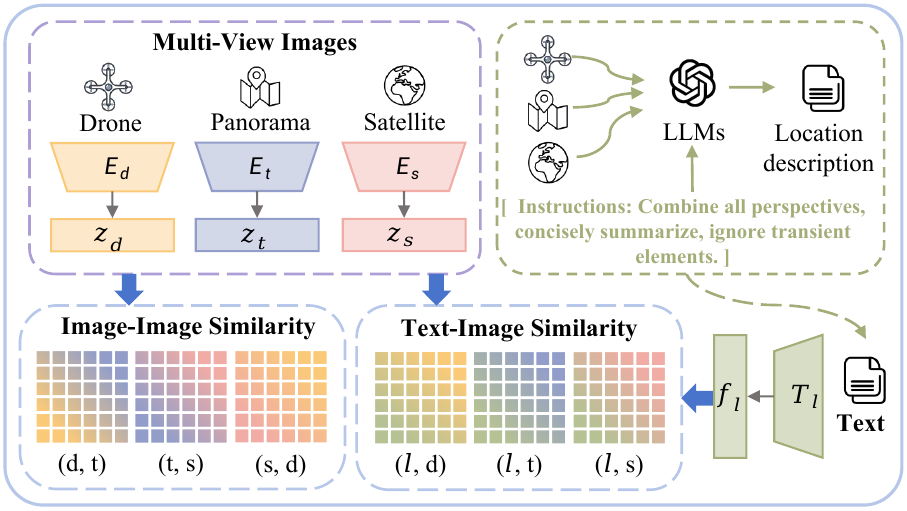}
		\caption{Overall workflow of GeoBridge. A unified location description is generated from aligned drone, street-view panorama, and satellite images and serves as the semantic anchor. GeoBridge jointly performs image-image and text-image alignment to learn a shared representation space for cross-view and cross-modal geo-localization.}
		\label{geo-framework}
	\end{center}
	\vspace{-13pt}
\end{figure}

\begin{figure*}[!t]
	\begin{center}		\includegraphics[width=1\linewidth]{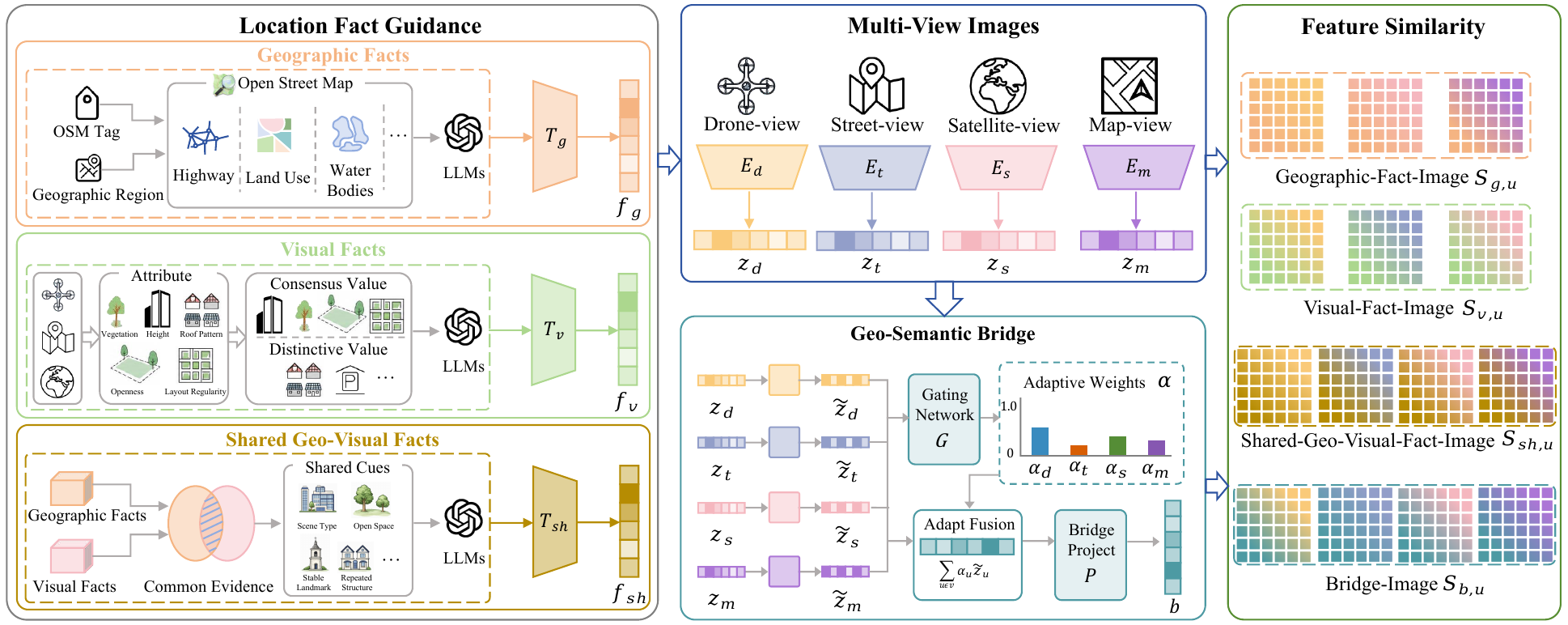}
		\caption{Overall framework of GeoBridge++ with multi-level location fact guidance and geo-semantic bridging. Geographic, visual, and shared geo-visual facts provide modality-aware supervision for different view groups, while the geo-semantic bridge adaptively aggregates drone, street-view, satellite, and static-map representations into a shared location-level representation for unified cross-view alignment.}
		\label{fig-framework}
	\end{center}
	\vspace{-13pt}
\end{figure*}

We introduce a contrastive framework GeoBridge for accurate geo-localization, which is shown in Fig.~\ref{geo-framework}. It uses a unified multi-view textual description ($l$) as the semantic anchor to align representations across three visual perspectives: drone, street-view panoramas, and satellite. The text serves as a semantic bridge, binding view-specific visual embeddings to view-invariant scene semantics. During the training process, language supervision aligns each visual branch to a common embedding space. At inference, the model works text-free for image$\leftrightarrow$image retrieval and naturally supports text$\rightarrow$image search.

GeoBridge employs three view-specific visual encoders \(E_u(\cdot)\), \({u\in\mathcal{V}_{\mathrm{G}}}\), and a shared text encoder $T_l(\cdot)$. All encoders are instantiated from CLIP. Each geographic instance comprises three images captured from different perspectives. For each instance, we craft a concise paragraph that serves as the semantic bridge. This description emphasizes stable, viewpoint-agnostic cues (\eg, roads, intersections, buildings, rivers, landmarks) and their relations, while omitting transient elements and viewpoint-specific phrasing. 

During the training process, each example is a quadruple $\{x_{d},x_{t},x_{s},l\}$. These inputs are encoded by the \(E_u(\cdot)\), \({u\in\mathcal{V}_{\mathrm{G}}}\), and $T_l(\cdot)$ to produce \(z_d,\, z_t,\, z_s,\, f_l, \in \mathbb{R}^{D}, \|z_\cdot\|_2 = 1\), and \(\|f_\cdot\|_2 = 1\). Text sequences are padded or truncated to \(L_{\max}\) before encoding, and all embeddings are mapped to the shared retrieval space.

For each view pair $(u,w)\!\in\!\{(d,t),(t,s),(s,d)\}$ and each text-to-view pair $(l,u)\!\in\!\{(l,d),(l,t),(l,s)\}$, we compute using cosine similarity between embeddings. A learnable temperature coefficient \(\tau\) scales these similarities to control the sharpness. Considering a candidate set of size $N$, the score matrices are computed as:
\begin{equation}
S_{u, w}=  z_u (z_w)^\top/\tau, \quad u\in\mathcal{V}_{\mathrm{G}},
\end{equation}
\begin{equation}
S_{l, u}=  f_l (z_u)^\top/\tau, \quad u\in\mathcal{V}_{\mathrm{G}}.
\end{equation}

Let $y_i$ denote the column index of correct match for query $i$ within the candidate set. We optimize an InfoNCE objective with a temperature-scaled softmax:
\begin{equation}
\mathcal{L}(S)\;=\;-\frac{1}{N}\sum_{i=1}^N \Bigg[\log\frac{\exp\!\big(S(i,y_i)\big)}{\sum_{j=1}^N \exp\!\big(S(i,j)\big)}\Bigg],
\end{equation}
where $i$ indexes queries and $j$ indexes gallery items. Cross-view alignment is enforced by averaging the losses over the three image-to-image pairs:
\begin{equation}
\mathcal{L}_{\mathrm{img}}=\cfrac{1}{3}\Big[\mathcal{L}(S_{d, t})+\mathcal{L}(S_{t, s})+\mathcal{L}(S_{s, d})\Big].
\end{equation}
Language-to-image alignment provides the semantic bridge, the calculation method is as follows:
\begin{equation}
\mathcal{L}_{\text{text}}=\frac{1}{3}\sum_{u\in\mathcal{V}_{\mathrm{G}}}\mathcal{L}\!\left(S_{l, u}\right).
\end{equation}
The total objective aggregates the image-to-image and language-to-image terms with equal weights:
\begin{equation}
\mathcal{L}_{\text{total}}=\mathcal{L}_{\text{img}}+\mathcal{L}_{\text{text}}.
\end{equation}

This joint objective draws image--image and text--image pairs together while pushing impostors apart. At inference, we encode image or text query and retrieve by nearest neighbors. For language-only queries, the description is restricted to a single scene and retrieval other views.

\subsection{GeoBridge++ Method}

GeoBridge++ builds upon GeoBridge by introducing location fact guidance and a geo-semantic bridge for unified cross-view geo-localization across drone, satellite, street-view, and static-map observations. As shown in Figure~\ref{fig-framework}, GeoBridge++ consists of four view-specific image encoders, three fact-text encoders, and a geo-semantic representation constructed from four image views.

For each view \(u\in\mathcal{V}\), the image branch employs an independent encoder \(E_u(\cdot)\), allowing each modality to preserve its own representational characteristics. Specifically, the observations of the location are denoted by
\(\{x_d,x_s,x_t,x_m\}\). The four image encoders are initialized from CLIP, and the normalized image representation \(z_u\), where \( u\in\mathcal{V}\), and \(\|z_\cdot\|_2 = 1 \) of each observation can be obtained.

To provide stable and location-discriminative supervision, GeoBridge++ constructs three types of location facts for different groups of observations: geographic facts (\(q_g\)), visual facts (\(q_v\)), and shared geo-visual facts (\(q_{sh}\)). Geographic facts are derived from OSM data to characterize explicit geographic and spatial structures. Visual facts describe stable and discriminative visual characteristics observed across drone, satellite, and street-view images. Shared geo-visual facts integrate information jointly supported by geographic structure and multi-view visual evidence, thereby providing a location-level factual summary. We employ three CLIP-initialized text encoders, \(T_g(\cdot)\), \(T_v(\cdot)\), and \(T_{sh}(\cdot)\), to obtain the corresponding normalized representations, denoted by \(f_g\), \(f_v\), and \(f_{sh}\) for the geographic, visual, and shared geo-visual facts, respectively.

In parallel, GeoBridge++ adaptively aggregates the four image representations into a geo-semantic bridge representation \(b\). This representation serves as a shared latent location-level target that integrates the explicit spatial structures conveyed by static maps with complementary visual evidence from the remaining views. Rather than imposing exhaustive pairwise constraints among all image modalities, GeoBridge++ coordinates heterogeneous observations through location fact guidance and alignment with the geo-semantic bridge representation.

\subsubsection{Location Fact Guidance}

Although multi-view observations provide rich cues for geo-localization, their observable content differs across viewpoints and representation forms. Visual observations mainly capture appearance and scene structure, whereas static maps explicitly encode symbolic geographic layouts. We therefore construct geographic, visual, and shared geo-visual facts to supervise different view groups according to their semantic relevance. Rather than directly generating a representation from multiple observations, modality-relevant attributes are extracted from each source and organized into factual descriptions. This fact-oriented construction constrains language generation while preserving modality-specific information and promoting geographically discriminative, cross-view consistent representations.

\textbf{Geographic Fact Guidance.} The motivation behind geographic fact guidance is to use geographic structure as a view-invariant anchor. Such structure provides objective spatial evidence that is intrinsically tied to physical locations. Unlike visual appearance, which may vary across viewpoints and imaging conditions, geographic structure captures relatively stable spatial properties of the physical environment, including road networks, building distributions, land-use patterns, public facilities, natural regions, and waterways. Although these properties are represented differently across visual and symbolic observations, drone images, satellite images, and static maps all reflect the underlying spatial organization of roads, buildings, and land-cover regions. Accordingly, geographic facts are used to supervise these three views.

For each geographic location, we first obtain the OSM data within its associated geographic region \(\mathcal{R}\), whose extent is determined by the corresponding observation. We consider the OSM tag set \(\mathcal{K} = \{\texttt{highway},\) \( \texttt{building},\texttt{landuse},\texttt{amenity},\texttt{natural}, \texttt{waterway}\}\), which covers major geographic entities relevant to the structural characteristics of a location.

Rather than directly using the raw and highly fine-grained OSM annotations, we convert them into compact structured geographic facts. Specifically, road tags are normalized into general categories, and their geometries are used to estimate the dominant orientation. Building tags are grouped into coarse categories, with building counts further summarized as density levels. Land-use annotations describe dominant functional regions, amenities provide cues about public facilities, and waterways and selected natural regions capture stable geographic structures. These normalized facts characterize both the composition and spatial organization of a location. For example, \textit{multiple residential roads with predominantly north--south trajectories, numerous apartments, residential land use, a school, and a river can be summarized as local roads with a dominant north--south orientation, dense residential buildings, a residential area, a school, and a river.}

We then employ an LLM to verbalize the structured geographic facts into a concise geographic factual description \(q_g\), constrained to the provided evidence to avoid introducing unobserved appearance cues or transient information. Its normalized representation \(f_g\) is subsequently obtained using the geographic fact encoder \(T_g(\cdot)\). For each view that reflects geographic structure, \(u\in\{d,s,m\}\), we compute the text-to-image similarity matrix $S_{g,u}$ using the geographic fact representation \(f_g\) and image representation \(z_u\):
\begin{equation}
S_{g, u}=  f_g (z_u)^\top/\tau, \quad u\in\{d,s,m\}.
\end{equation}

Then, the geographic fact guidance loss $\mathcal{L}_{g}$ is defined using a text-to-image InfoNCE objective:
\begin{equation}
\mathcal{L}_{g}
=
-\frac{1}{3N}
\sum_{u\in\{d,s,m\}}
\sum_{i=1}^{N}
\log
\frac{\exp(S_{g,u}(i,y_i))}
{\sum_{j=1}^{N}\exp(S_{g,u}(i,j))}.
\end{equation}

This objective aligns drone, satellite, and static-map representations with the geographic facts of the same location while separating samples from different locations. It encourages the model to focus on stable spatial structures and geographic cues rather than relying solely on appearance-level correspondences. Consequently, GeoBridge++ learns a more robust and location-discriminative embedding space despite substantial appearance differences among aerial, overhead, and symbolic map observations.

\textbf{Visual Fact Guidance.} Visual fact guidance aims to identify stable and geographically discriminative cues shared across viewpoints. Although OSM-derived geographic facts provide explicit supervision on geographic structures, they cannot fully capture certain location-specific visual characteristics. Such cues, including roof textures, facade continuity, spatial openness, and recurring architectural patterns, are observable across heterogeneous views. Drone, satellite, and street-view images reveal these salient appearance characteristics from complementary viewpoints and spatial scales, providing visual evidence that is difficult to represent using symbolic geographic data alone. Accordingly, visual facts are used to supervise these three views.

For each location, we independently extract a set of view-specific visual attributes from the drone, satellite, and street-view observations using an LLM. These attributes focus on relatively stable characteristics, including building height and density, vegetation distribution, spatial openness, layout regularity, roof appearance, facade patterns, surface characteristics, and repetitive architectural structures. We then aggregate the view-specific attributes into cross-view consensus visual facts, retaining visual characteristics consistently supported across the three views. In addition, an LLM is employed to identify a discriminative visual fact from multi-view attributes, emphasizing the distinctive and recognizable appearance pattern of the location.

The resulting consensus visual facts, together with the discriminative visual fact, are then provided to an LLM to generate a concise and coherent visual factual description \(q_v\). This process is constrained to visually grounded evidence and excludes temporary objects, environmental conditions, and geographic information already represented by geographic facts. The normalized representation \(f_v\) is then obtained by encoding the visual factual description \(q_v\) with the visual fact encoder \(T_v(\cdot)\).

Visual facts supervise the drone, satellite, and street-view representations. For each visual view \(u\in\{d,s,t\}\), the text-to-image similarity matrix is computed as:

\begin{equation}
S_{v, u}=  f_v (z_u)^\top/\tau, \quad u\in\{d,s,t\}.
\end{equation}

Accordingly, the visual fact guidance loss is defined with a text-to-image InfoNCE objective:

\begin{equation}
\mathcal{L}_{v}
=
-\frac{1}{3N}
\sum_{u\in\{d,s,t\}}
\sum_{i=1}^{N}
\log
\frac{\exp(S_{v,u}(i,y_i))}
{\sum_{j=1}^{N}\exp(S_{v,u}(i,j))}.
\end{equation}

This objective aligns drone, satellite, and street-view representations with stable cross-view visual facts while separating samples from different locations. Complementing the explicit geographic and symbolic spatial structures captured by geographic facts, visual facts provide appearance-based cues that reduce viewpoint-induced discrepancies and improve location discriminability.

\textbf{Shared Geo-Visual Fact Guidance.} Geographic and visual facts characterize a location from complementary perspectives. Since some location properties are reflected by both sources, we further construct shared geo-visual facts as common supervision for all four views.

Specifically, we compare the geographic facts with the visual facts and retain only mutually supported attributes, including building density, spatial openness, layout regularity, vegetation or open-space presence, surface characteristics, and stable structural patterns. For example, building density is retained only when the density estimated from OSM-derived building distributions is consistent with that observed across visual attributes. Attributes supported by only one source or showing clear inconsistencies are discarded. Fine-grained appearance cues remain in visual facts, whereas explicit spatial structures and relationships are retained in geographic facts rather than being duplicated in shared representation. Through this consistency-based filtering process, we obtain a compact set of shared geo-visual facts that reflects location properties jointly expressed by geographic structure and visual appearance.

We then employ an LLM to verbalize the retained shared facts into a concise shared geo-visual factual description \(q_{sh}\). The verbalization is restricted to the filtered evidence and avoids introducing new modality-specific information. The normalized representation \(f_{sh}\) is subsequently obtained by encoding \(q_{sh}\) with the shared fact encoder \(T_{sh}(\cdot)\). Unlike geographic facts and visual facts, shared geo-visual facts are used to supervise all modalities. For each view $u$, we compute the text-to-image similarity matrix as $S_{\mathrm{sh},u}$:
\begin{equation}
S_{sh, u}=  f_{sh} (z_u)^\top/\tau, \quad u\in\mathcal{V}.
\end{equation}

The shared fact guidance loss is then defined using the text-to-image InfoNCE training objective:
\begin{equation}
\mathcal{L}_{sh}
=
-\frac{1}{4N}
\sum_{u\in\mathcal{V}}
\sum_{i=1}^{N}
\log
\frac{\exp(S_{sh,u}(i,y_i))}
{\sum_{j=1}^{N}\exp(S_{sh,u}(i,j))}.
\end{equation}

This objective uses shared geo-visual facts as common textual anchors for drone-view, satellite-view, street-view, and static map representations. By aligning all modalities with the same shared factual representation, GeoBridge++ learns a unified fact-aware embedding space.

\subsubsection{Geo-Semantic Bridge}
\begin{figure}[!t]
	\begin{center}		\includegraphics[width=1\linewidth]{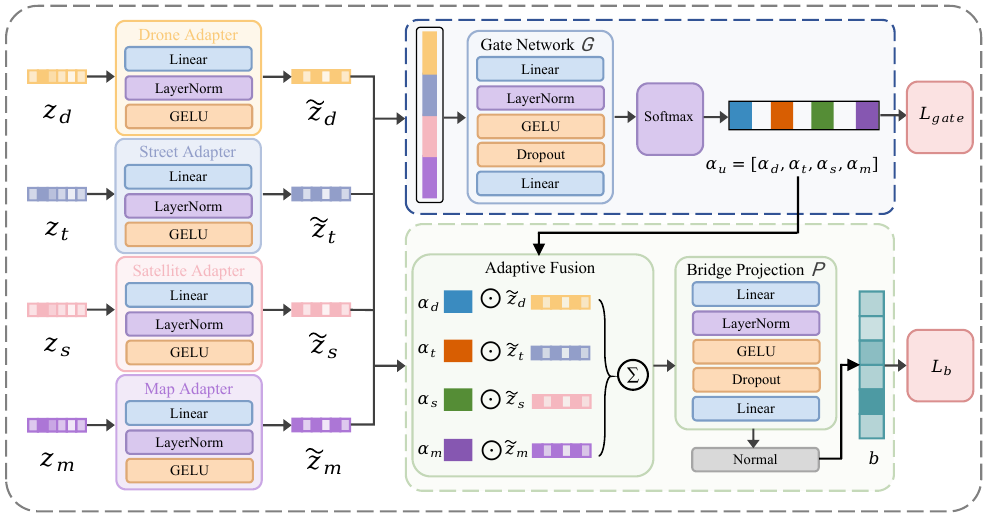}
		\caption{Architecture of the geo-semantic bridge.}
		\label{geo-semantci}
	\end{center}
	\vspace{-13pt}
\end{figure}

Location fact guidance provides stable textual supervision but does not explicitly capture interactions among heterogeneous image observations. Drone, satellite, street-view, and static-map views contain complementary information across different perspectives and representation forms. We therefore construct a geo-semantic bridge that adaptively aggregates the four views into a shared location-level representation, as illustrated in Figure~\ref{geo-semantci}.

Given the view representations \(z_u\), each feature is first transformed by a lightweight view-specific adapter:
\begin{equation}
\tilde{z}_u
=
\operatorname{GELU}
\left(
\operatorname{LN}
\left(
W_u z_u + b_u
\right)
\right),
\quad u\in\mathcal{V},
\end{equation}
where \(W_u\) and \(b_u\) denote the weight matrix and bias of the view-specific linear transformation, respectively. Although the view-specific encoders produce representations of the same dimensionality, their feature distributions may differ because of cross-view heterogeneity. The adapters calibrate these distributions before aggregation while preserving the representational characteristics of each view.

The adapted representations are subsequently concatenated and passed through a gating network to estimate location-dependent fusion weights:
\begin{equation}
\boldsymbol{\alpha} =
\operatorname{Softmax}
\left(
G\left(
[\tilde{z}_d;
 \tilde{z}_s;
 \tilde{z}_t;
 \tilde{z}_m]
\right)
\right),
\label{bridge_gate}
\end{equation}
where \(G(\cdot)\) denotes the gating network, implemented using two linear layers with layer normalization, GELU activation, and dropout between them. Here, $\boldsymbol{\alpha}=
[\alpha_d,\alpha_s,\alpha_t,\alpha_m]$ contains the contribution weights of the four views. These weights satisfy
$\alpha_u\geq 0$ and
$\sum_{u\in\mathcal{V}}\alpha_u=1$. Unlike fixed averaging, the gating mechanism dynamically determines the contribution of each modality for the current location, allowing it to emphasize informative views while suppressing less reliable or redundant features.

The adapted representations are then aggregated using the learned weights and projected into a shared retrieval space. The geo-semantic bridge representation \(b\) is obtained:
\begin{equation}
b =
\operatorname{Norm}
\left(
P
\left(
\sum_{u\in\mathcal{V}}
\alpha_u\tilde{z}_u
\right)
\right),
\label{bridge_feature}
\end{equation}
where \(P(\cdot)\) denotes the bridge projection, which is implemented using two linear layers with layer normalization, GELU activation, and dropout between them. Through location-dependent aggregation, \(b\) integrates complementary structural and appearance information from all four observations while preventing any single modality from consistently dominating the representation.

To prevent the gating network from collapsing onto only a small subset of views, we regularize its batch-averaged output toward a predefined modality prior \(\boldsymbol{\pi}=[\pi_d,\pi_s,\pi_t,\pi_m]\). The regularization is defined as follows:
\begin{equation}
\mathcal{L}_{\mathrm{gate}}
=
\frac{1}{|\mathcal{V}|}
\left\|
\frac{1}{N}
\sum_{i=1}^{N}
\boldsymbol{\alpha}_i
-
\boldsymbol{\pi}
\right\|_2^2.
\label{gate_regularization}
\end{equation}

This batch-level regularization discourages persistent modality collapse while retaining location-dependent adaptive weighting. It therefore encourages all views to participate in bridge construction without imposing identical fusion weights across locations.

To encourage every image modality to approach the shared bridge representation, we employ a bridge-guided InfoNCE objective. For each view \(u\), the bridge-to-image similarity $S_{b,u}$ is computed as:
\begin{equation}
S_{b, u}=  b (z_u)^\top/\tau, \quad u\in\mathcal{V}.
\end{equation}

The bridge-to-image InfoNCE loss is then defined as:
\begin{equation}
\mathcal{L}_{b}
=
-\frac{1}{4N}
\sum_{u\in\mathcal{V}}
\sum_{i=1}^{N}
\log
\frac{
\exp\left(S_{b,u}(i,y_i)\right)
}{
\sum_{j=1}^{N}
\exp\left(S_{b,u}(i,j)\right)
}.
\label{bridge_loss}
\end{equation}

The geo-semantic bridge provides all image modalities with a common location-level alignment target. Rather than requiring exhaustive pairwise constraints among image modalities, it enables complementary information from different views to be coordinated through a shared latent representation. Consequently, the bridge effectively facilitates information exchange across heterogeneous observations and promotes a consistent embedding space for unified any-view-to-any-view geo-localization.

\subsubsection{Multi-Level Contrastive Learning}

GeoBridge++ jointly optimizes the fact-guidance and geo-semantic bridge objectives defined above. Instead of directly imposing pairwise contrastive constraints between image modalities, heterogeneous views are coordinated through factual supervision and the shared geo-semantic bridge.

The final training objective is defined as:
\begin{equation}
\mathcal{L} =
 \mathcal{L}_{g} + \mathcal{L}_{v} + \mathcal{L}_{sh} + \mathcal{L}_{b} + \lambda\mathcal{L}_{\mathrm{gate}} ,
\label{overall_loss}
\end{equation}
where $\lambda$ controls the contribution of the gate regularization term. $\mathcal{L}_{g}$, $\mathcal{L}_{v}$, and $\mathcal{L}_{sh}$ provide complementary semantic supervision, while $\mathcal{L}_{b}$ promotes location-level consistency across heterogeneous views and $\mathcal{L}_{gate}$ stabilizes adaptive fusion. This multi-level objective enables unified cross-view alignment without exhaustive image-to-image supervision. 

During inference, GeoBridge++ supports both any-view-to-any-view image retrieval and text-to-image geo-localization. For image retrieval, query and gallery representations from arbitrary views are ranked by cosine similarity. For text-to-image retrieval, the query is encoded by the shared geo-visual fact encoder $T_{sh}(\cdot)$ and matched against image representations in the shared embedding space.

\section{Datasets Construction}
\begin{table*}[ht]
  \centering
  \caption{Comparison of cross-view geo-localization datasets.
  \emph{Platform}: A = Aerial, D = Drone, G = Street, T = Street-view panorama, S = Satellite, M = Map. 
  \emph{Multi-scale}: multiple scales per location. 
  \emph{GPS-tag}: geo-coordinates provided.
  \emph{Altitudes}: multiple drone heights.}
  \label{tab:datasets}
    \begin{tabularx}{\textwidth}{
    l
    c
    c
    >{\centering\arraybackslash}X
    c
    c
    c
    c
}
    \toprule
    \textbf{Dataset} & \textbf{Year} & \textbf{Platform}& \textbf{Region(s)} & \textbf{Size(k)} & \textbf{Multi-scale} & \textbf{GPS-tag} & \textbf{Altitudes}  \\
    \midrule
    CVUSA~\cite{CVUSA}   & 2015 & T{+}S  & Nationwide (USA)& 44.4 {+} 44.4& \xmark & \cmark & — \\
    Tian \emph{et al.}~\cite{tian2017cross}& 2017& G{+}A& Multi-city (USA) &        35.4 {+} 35.4&   \xmark   & \cmark & — \\
    CVACT~\cite{CVACT}& 2019 & T{+}S & \mbox{City-scale(Canberra, Australia)}&  128.3 {+} 128.3&        \xmark    & \cmark & — \\
    VIGOR~\cite{zhu2021vigor} & 2021 & T{+}A& Multi-city (USA)& 105.2 {+} 90.6&  \xmark   & \cmark & — \\
    University\textendash1652~\cite{zheng2020university}& 2020 & D{+}G{+}S& \mbox{Campus/City, multi-source}& 37.9 {+} 2.6 {+} 0.7&\cmark& \xmark & \xmark \\
    DenseUAV~\cite{dai2023vision} & 2023    & D{+}S & \mbox{City-scale (Zhejiang, China)} & 9.1 {+} 31.7    &    \cmark      & \xmark & \cmark \\
    SUES\textendash200~\cite{zhu2023sues}            & 2023 & D{+}S   & Multi-city (Region)& 80 {+} 1.6 &   \cmark     & \xmark & \cmark \\
    CVGlobal$^{\dagger}$~\cite{ye2024cross}& 2024 &  T{+}S{+}M  &City-scale & 134.2 {+} 134.2 {+} 134.2 &   \xmark    & \cmark & — \\
    \midrule
    GeoLoc~\cite{song2026geobridge} (ours)         & 2025 & D{+}T{+}S& Global (36 countries)& 52.7 {+} 52.7 {+} 52.7 &  \xmark    & \cmark & \,\cmark$^{\ddagger}$ \\
    GeoLoc-MM (ours)         & 2026 & D{+}T{+}S{+}M& Global (36 countries)& 316.1 {+} 52.7 {+} 316.1 {+} 316.1 &    \cmark   & \cmark & \,\cmark\\
    \bottomrule
  \end{tabularx}
  \raggedright
  {\footnotesize
  $^{\dagger}$It covers seven cities—Munich, Toronto, Singapore, Sao Paulo, Brisbane, Cape Town, and Johannesburg—and its map-based setting is limited to street-view-to-map retrieval. $^{\ddagger}$ GeoLoc comprises UAV imagery collected under diverse conditions, with flight altitude determined primarily by the desired ground coverage; five coverage scales are provided: $80\!\times\!80$, $100\!\times\!100$, $120\!\times\!120$, $150\!\times\!150$, and $180\!\times\!180$~(m$^2$). 
  }
    
 \vspace{-13pt}
\end{table*}

\begin{figure}[!t]
	\begin{center}
		\includegraphics[width=1\linewidth]{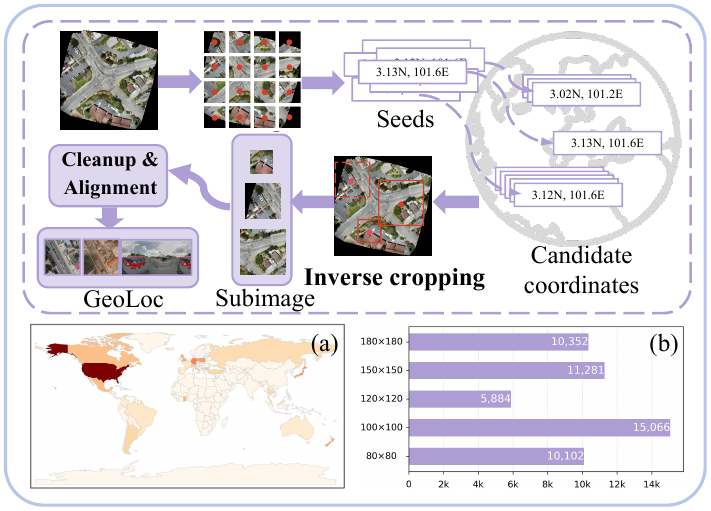}
		\caption{Multi-view data processing for the GeoLoc dataset. (a) Global distribution of multi-view image groups (darker shades indicate higher density). (b) Counts of drone images per ground-footprint bin as a function of covered area (m$^2$).}
		\label{geoloc}
	\end{center}
    \vspace{-20pt}
\end{figure}
\subsection{GeoLoc dataset}
We curate GeoLoc, a cross-view dataset with precise geographic and semantic alignment. The process, as shown in Figure~\ref{geoloc}, comprises four stages: 
(1) drone image acquisition and seed generation, 
(2) cross-source harvesting with inverse cropping, 
(3) duplicate suppression and cleanup, and (4) multi-view alignment. 
We detail the operations below.

\textbf{Drone images acquisition and seed generation.}
We collect globally sourced drone images from OpenAerialMap\footnote{\url{https://map.openaerialmap.org/}} with geo-referencing metadata (projection and coordinates). For each large drone image, we slide a fixed $80\!\times\!80$-pixel window to estimate the ground footprint, recording the pixel center $(x,y)$ together with the latitude and longitude $(\varphi,\lambda)$. These centers serve as seeds for subsequent cross-source retrieval and alignment.

\textbf{Cross-source harvesting with inverse cropping.} For each seed, we query Google Street View\footnote{\url{https://www.google.com/streetview/}} (GSV) to obtain candidate panoramic coordinates and retain only those whose locations fall within the coverage of drone image. Using the GSV location as an anchor, we inverse crop the original drone image according to the ground footprint to obtain a co-located drone subimage. To improve robustness to scale and viewpoint variations, we consider five candidate ground footprints, approximately $80\!\times\!80$, $100\!\times\!100$, $120\!\times\!120$, $150\!\times\!150$, and $180\!\times\!180~\text{m}^2$. For each drone observation, we ask an expert to select an appropriate ground scale according to its acquisition altitude, and the corresponding subimage is generated at the matched output resolution. As a result, the dataset contains drone subimages at diverse spatial scales, while each geographic location is associated with only one drone image at its selected scale.

\textbf{Duplicate suppression and cleanup.}
We perform an initial screening of the drone subimages. If two drone sub-images have more than $50\%$ ground-coverage overlap, or if their latitude and longitude are identical, they are treated as duplicates and only one is retained. Subimages in which more than $1\%$ of the area consists entirely of black or entirely of white pixels (\eg, sensor-strip edges) are discarded.

To further remove low-quality content, we employ a cascade of  three lightweight quality gates:

\begin{itemize}
    \item BH-Gate (Blur--Haze): measures global texture using Laplacian variance and pixel variance, eliminating low-quality blurry sub-images such as those with motion blur, distant scenes under heavy fog, and heavily compressed visual textures.

    \item C-Gate (Global Contrast): eliminates low-quality images with weak edges based on global contrast, filtering drone subimages that are slightly out of focus, severely over- or under-exposed.

    \item UN-Gate (Uniformity--Noise): uses entropy as the primary criterion, combined with variance range and ratios of saturated pixels, to distinguish uniform but informationless pseudo-textures from noise patterns, eliminating images lacking features such as expanses of blue sky, open water, deserts, and grasslands.
\end{itemize} 

In practice, this cascade yields a pool of structurally rich, high-contrast, and semantically useful crops.

\textbf{Tri-view alignment.} For each quality-screened drone subimage, we use its seed coordinates to retrieve Google Street View panoramas. This API returns static, non-interactive panoramic imagery. For each sampled location, we request images using geographic coordinates together with camera parameters. Our use of the API follows service-specific terms\footnote{\url{https://developers.google.com/maps/documentation/streetview/policies}} \footnote{\url{https://cloud.google.com/maps-platform/terms/maps-service-terms/index-20240515}}. We then crop Google Satellite\footnote{\url{https://www.google.com/earth/}} tiles to align with the co-located region covered by the subimage. This yields 52,679 geographically co-located drone--street panorama--satellite triplets spanning 36 countries. We designate 5,351 triplets from non-overlapping cities as a held-out evaluation set. Dataset statistics are summarized in Table~\ref{tab:datasets}. Notably, our method remains effective on public datasets, indicating that its performance is not tied to any proprietary data source. See the supplementary material for details. 

\begin{table}[t]
  \centering
  \caption{Comparison on the University\textendash1652~\cite{zheng2020university} dataset; best results are in bold, second-best are underlined.}
  \label{tab:university}
    \resizebox{\columnwidth}{!}{%

  \begin{tabular}{c c c c c }
    \toprule
    \multirow{2}{*}{\textbf{Method}} &  \multicolumn{2}{c}{\textbf{Drone to Satellite}}& \multicolumn{2}{c}{\textbf{Satellite to Drone}}  \\
    \cmidrule(lr){2-3}\cmidrule(lr){4-5}
    & \textbf{R@1} & \textbf{AP}& \textbf{R@1} & \textbf{AP} \\
    \midrule
    SAIG-D~\cite{SAIG-D}& 78.85&  81.62& 86.45&78.48  \\
    DWDR~\cite{DWDR}&86.41 & 88.41 & 91.30&86.02  \\
    MBF~\cite{MBF}& 89.05&  90.61&92.15 &84.45  \\
    MCCG~\cite{MCCG}& 89.64 &91.32 &94.30 &89.39 \\
    SeGCN~\cite{SeGCN}&  89.18& 90.89& 94.29& 89.65\\
    CCR~\cite{CCR}& 92.54 & 93.78&95.15 & 91.80\\
    MEAN~\cite{MEAN}& 93.55& 94.53& 96.01& 92.08 \\
    DAC~\cite{DAC}& 94.67&  95.50& 96.43 & 93.79\\
    Sample4Geo~\cite{Sample4geo}& 92.65& 93.81& 95.14 & 91.39\\
    \midrule
    GeoBridge (ours) & \underline{95.82} &\underline{97.77} & \underline{97.14}&\underline{95.05} \\
    GeoBridge++ (ours) & \textbf{96.95} &\textbf{98.40} & \textbf{98.29}&\textbf{96.46} \\
    \bottomrule
  \end{tabular}
  }
  \raggedright
  \vspace{-8pt}
\end{table}

\begin{table}[t]
  \centering
  \caption{Comparison on the SUES\textendash200~\cite{zhu2023sues} dataset; best results are in bold, second-best are underlined.}
  \label{tab:sues}
    \resizebox{\columnwidth}{!}{
    \begin{tabular}{c c c c c c c c c }

    \toprule
    \multicolumn{9}{c}{\textbf{Drone to Satellite}} \\
    \midrule
    \multirow{2}{*}{\textbf{Method}} & \multicolumn{2}{c}{\textbf{150m}}& \multicolumn{2}{c}{\textbf{200m}} & \multicolumn{2}{c}{\textbf{250m}} & \multicolumn{2}{c}{\textbf{300m}}  \\
    \cmidrule(lr){2-3}\cmidrule(lr){4-5}\cmidrule(lr){6-7}\cmidrule(lr){8-9}
    & \textbf{R@1} & \textbf{AP}& \textbf{R@1} & \textbf{AP} & \textbf{R@1} & \textbf{AP} & \textbf{R@1} & \textbf{AP} \\
    \midrule
    MBF~\cite{MBF}   & 85.62 & 88.21  & 87.43& 90.02  & 90.65 & 92.53&92.12&93.63 \\
    MCCG~\cite{MCCG}&  82.22&  85.47 & 89.38&  91.41 & 93.82 & 95.04&95.07&96.20 \\
    CCR~\cite{CCR}  & 87.08 &  89.55 & 93.57&  94.90 & 95.42 &96.28 &96.82&97.39 \\
    SeGCN~\cite{SeGCN}&90.80 & 92.32& 91.93& 93.41& 92.53& 93.90&93.33& 94.61\\
     CGSI~\cite{CGSI}& \underline{95.95} & 96.80&97.72 & 98.15&97.60&98.03 & \underline{97.83}& 98.23 \\
    
    \midrule
    GeoBridge (ours)& 95.68& \underline{97.49}& \underline{97.88}& \underline{98.89}&\underline{97.85} & \textbf{98.84}&97.63 & \underline{98.62}\\
    GeoBridge++ (ours)& \textbf{97.68}& \textbf{98.49}& \textbf{98.05}& \textbf{98.91}&\textbf{97.90} & \textbf{98.84}&\textbf{98.40} & \textbf{99.11}\\
    \bottomrule
    \multicolumn{9}{c}{\textbf{Satellite to Drone}} \\
    \midrule
    \multirow{2}{*}{\textbf{Method}} & \multicolumn{2}{c}{\textbf{150m}}& \multicolumn{2}{c}{\textbf{200m}} & \multicolumn{2}{c}{\textbf{250m}} & \multicolumn{2}{c}{\textbf{300m}}  \\
    \cmidrule(lr){2-3}\cmidrule(lr){4-5}\cmidrule(lr){6-7}\cmidrule(lr){8-9}
    & \textbf{R@1} & \textbf{AP}& \textbf{R@1} & \textbf{AP} & \textbf{R@1} & \textbf{AP} & \textbf{R@1} & \textbf{AP} \\
    \midrule
    MBF~\cite{MBF}   & 88.75 & 84.74  & 91.25& 89.95  & 93.75 & 90.65&96.25&91.60 \\
    MCCG~\cite{MCCG}& 93.75 &  89.72 &93.75 &  92.21 & 96.25 &96.14 &\textbf{98.75}& 96.64\\
    CCR~\cite{CCR}&  92.50& 88.54& 97.50&95.22 &97.50 & 97.10&97.50&97.49 \\
    SeGCN~\cite{SeGCN} & 93.75 & 92.45& 95.00& 93.65& 96.25& 94.39&97.50& 94.55\\
    CGSI~\cite{CGSI}& 97.50 & 96.22& 98.75 & 97.62 & 98.75 &\underline{98.01}& \textbf{98.75}& 97.92\\

    \midrule
    GeoBridge (ours) &  \underline{98.99}& \underline{97.98}& \underline{98.98}&\underline{97.98} & \underline{99.01}&97.98&\textbf{98.75} & \underline{97.95} \\
    GeoBridge++ (ours) &  \textbf{99.03}& \textbf{98.24}& \textbf{99.08}&\textbf{98.39} & \textbf{99.11}&\textbf{98.94}&\textbf{98.75} & \textbf{98.84} \\
    \bottomrule
  \end{tabular}
  }
  \raggedright
\vspace{-13pt}
\end{table}

\subsection{GeoLoc-MM dataset}
Building upon GeoLoc, we further construct GeoLoc-MM, a million-scale, multi-view, and multi-scale dataset. It extends GeoLoc by introducing static maps as an additional view and providing complete multi-scale observations for each selected location. Specifically, each location contains one street-view panorama together with spatially aligned drone, satellite, and static-map observations at six scales: $80\times80$, $100\times100$, $120\times120$, $150\times150$, $180\times180$, and $200\times200$~m$^2$. The dataset details are summarized in Table~\ref{tab:datasets}. Further details are in the supplementary material. 

\textbf{Full-scale observation expansion.} For each location, we use its seed coordinates as the common center to construct observations at six predefined scales. Drone images are obtained by inverse cropping the original geo-referenced imagery, while satellite images are collected to cover the same geographic extent. Consequently, the two views are spatially aligned at each scale, while a single street-view panorama is retained as the ground-level observation. This multi-scale expansion provides consistent observations of the same location across varying spatial contexts.

\textbf{Static-map acquisition and alignment.} We further introduce static maps\footnote{\url{https://www.openstreetmap.org}} as an additional geographic observation view. For each location, static-map images are generated at six scales to cover the same geographic extents as the corresponding drone and satellite observations. This scale-wise alignment ensures that drone, satellite, and static-map observations remain spatially consistent while providing complementary visual and structural information.

\textbf{Four-view multi-scale organization.} After scale completion and modality alignment, each location contains one street-view panorama and six images each from the drone, satellite, and static-map views. GeoLoc-MM therefore forms geographically aligned four-view groups with complete multi-scale coverage. Compared with GeoLoc, it extends the original three-view setting to four views and enables systematic and comprehensive evaluation of any-view-to-any-view geo-localization and scale robustness.

\begin{table}[t]
  \centering
  \caption{Comparison on the CVUSA~\cite{CVUSA}  and VIGOR~\cite{zhu2021vigor} datasets for street-to-satellite retrieval; best results are shown in bold, second-best are underlined.}
  \label{tab:street_satellite}
  \resizebox{\columnwidth}{!}{
  \begin{tabular}{c cc cc cc}
    \toprule
    \multirow{3}{*}{\textbf{Method}}
      & \multicolumn{2}{c}{\textbf{CVUSA}}& \multicolumn{2}{c}{\textbf{VIGOR-Same}} & \multicolumn{2}{c}{\textbf{VIGOR-Cross}} \\
      \cmidrule(lr){2-3} \cmidrule(lr){4-5}\cmidrule(lr){6-7}
      & \textbf{R@1} & \textbf{R@1\%} & \textbf{R@1} & \textbf{Hit} & \textbf{R@1} & \textbf{Hit} \\
    \midrule
    TransGeo~\cite{Transgeo}  & 94.08 & 99.77 & 61.48& 73.09& 18.99 & 21.21 \\
    FRGeo~\cite{FRGeo} & 97.06 & 99.85 & 71.26 & 82.41 & 37.54 &40.66 \\
    SAIG-D~\cite{SAIG-D}&96.08&99.86&65.23&74.11&33.05&36.71\\   VimGeo~\cite{VimGeo}&96.19&99.52&55.24&57.43&19.31&20.72\\
    Sample4Geo~\cite{Sample4geo}   & 98.68 & 99.87 & 77.86 & 89.82 & 61.70 & 69.87 \\
    Panorama-BEV~\cite{ye2024cross} & 98.71 & 99.70 & 82.18 & - & 72.19 & - \\
    AuxGeo~\cite{AuxGeo}& 98.80 & 99.85 & 80.34 & \underline{93.78} & 63.94 & 76.25 \\
    
    \midrule
    GeoBridge (ours) & \underline{99.14}&  \underline{99.98}& \underline{85.82} & 92.24 &  \underline{73.87}& \underline{81.51} \\
    GeoBridge++ (ours) & \textbf{99.34}&  \textbf{99.99}& \textbf{89.49} & \textbf{94.92} &  \textbf{78.20}& \textbf{86.97} \\
    \bottomrule
  \end{tabular}
  }
  \vspace{-13pt}
\end{table}

\begin{table}[t]
  \centering
  \caption{Comparison on the CVGlobal~\cite{ye2024cross} dataset for street-to-satellite retrieval; best results are shown in bold, second-best are underlined.}
  \label{CVG:street_satellite}
  \resizebox{\columnwidth}{!}{
  \begin{tabular}{c cc cc cc cc}
    \toprule
    \multirow{3}{*}{\textbf{Method}}
      & \multicolumn{4}{c}{\textbf{Regional val}}& \multicolumn{4}{c}{\textbf{Regional test}} \\
      \cmidrule(lr){2-5} \cmidrule(lr){6-9}
      & \textbf{R@1} & \textbf{R@5} & \textbf{R@10} & \textbf{R@1\%} & \textbf{R@1} & \textbf{R@5} & \textbf{R@10}& \textbf{R@1\%}\\
    \midrule
    GeoDTR~\cite{zhang2023cross}  & 47.42 & 68.43 & 78.11& 99.05& 25.43 & 42.07&51.66&84.74 \\
    SAIG-D~\cite{SAIG-D} & 71.92 & 92.83 & 96.05 & 99.81 & 34.72 &61.53&71.08&91.47 \\
    Sample4Geo~\cite{Sample4geo}   & 97.20 & 99.43 & 99.69 & \underline{99.93} & 84.66 & 92.66 &94.42&98.10\\
    Panorama-BEV~\cite{ye2024cross} & \underline{97.78} & 99.63 & 99.79 &  \underline{99.93} & 85.68 & 92.91&94.77&98.21 \\
    \midrule
    GeoBridge (ours) & 97.38&  \underline{99.64}& \underline{99.85} & \underline{99.93} &  \underline{86.14}& \underline{94.45} & \underline{96.95}& \underline{98.46}\\
    GeoBridge++ (ours) & \textbf{97.82}&  \textbf{99.74}& \textbf{99.89} & \textbf{99.95} &  \textbf{87.32}& \textbf{95.98} & \textbf{97.77}& \textbf{99.36}\\
    \bottomrule
  \end{tabular}
  }
  \vspace{-13pt}
\end{table}

\begin{table}[t]
  \centering
  \caption{Comparison on the CVGlobal~\cite{ye2024cross} dataset for street-to-map retrieval; best results are shown in bold, second-best are underlined.}
  \label{CVG:street_map}
  \resizebox{\columnwidth}{!}{
  \begin{tabular}{c cc cc cc cc}
    \toprule
    \multirow{3}{*}{\textbf{Method}}
      & \multicolumn{4}{c}{\textbf{Regional val}}& \multicolumn{4}{c}{\textbf{Regional test}} \\
      \cmidrule(lr){2-5} \cmidrule(lr){6-9}
      & \textbf{R@1} & \textbf{R@5} & \textbf{R@10} & \textbf{R@1\%} & \textbf{R@1} & \textbf{R@5} & \textbf{R@10}& \textbf{R@1\%}\\
    \midrule
    GeoDTR~\cite{zhang2023cross} &11.95&24.33&34.07&89.32&4.57&10.47&16.07&53.45  \\
    SAIG-D~\cite{SAIG-D} &37.27&70.61&80.01&97.53&4.64&14.80&22.16&61.55 \\
    Sample4Geo~\cite{Sample4geo}   &75.38&92.80&95.67&99.54&46.94&70.46&77.91&91.47 \\
    Panorama-BEV~\cite{ye2024cross}&\underline{81.31} & \underline{95.84} &\underline{97.68} & \underline{99.71} & \underline{49.41} & \underline{71.04} &\underline{78.33} & \underline{91.96}  \\
    \midrule
    GeoBridge++ (ours) & \textbf{83.05}&  \textbf{96.98}& \textbf{98.74} & \textbf{99.84} &  \textbf{58.96}& \textbf{85.78} & \textbf{88.29}& \textbf{93.35}\\
    \bottomrule
  \end{tabular}
  }
  \vspace{-13pt}
\end{table}

\begin{table}[t]
  \centering
  \caption{Comparison on the GeoLoc dataset. D, T, and S denote Drone, Street-View Panorama, and Satellite, respectively. The best results are shown in bold, the second-best results are underlined.}
  \label{tab:GeoLoc}
  \resizebox{\columnwidth}{!}{%
  \begin{tabular}{c*{4}{cc}}
    \toprule
    \multirow{2}{*}{\textbf{Method}}
      & \multicolumn{2}{c}{\textbf{D2S}}
      & \multicolumn{2}{c}{\textbf{S2D}}
      & \multicolumn{2}{c}{\textbf{D2T}}
      & \multicolumn{2}{c}{\textbf{T2D}} \\
      \cmidrule(lr){2-3}\cmidrule(lr){4-5}\cmidrule(lr){6-7}\cmidrule(lr){8-9}
     & \textbf{R@1} & \textbf{AP} & \textbf{R@1} & \textbf{AP} & \textbf{R@1} & \textbf{AP} & \textbf{R@1} & \textbf{AP} \\
    \midrule
    Sample4Geo~\cite{Sample4geo} & 27.27 & 39.69 & 28.70 & 40.32 & 29.51& 31.17& 15.56& 29.68\\
    MEAN~\cite{MEAN} & 21.52& 27.08 & 21.38 &26.97 &13.08&17.76& 1.87&7.74\\
    DAC~\cite{DAC}   & 6.19 & 8.41 & 13.91& 15.16&13.74&15.16 & 19.34 &23.01 \\
    CAMP~\cite{CAMP}   & 19.60 & 24.39 & 14.88 & 18.75 &11.31 &12.34 & 11.46&19.16 \\
    MCCG~\cite{MCCG}   & 12.23 & 13.22 & 14.73& 17.70 & 15.51&19.11 & 12.90&15.75 \\
    CCR~\cite{CCR}   & 12.91 & 15.56 & 13.89 & 16.04 & 10.91&12.34 & 10.81&14.77 \\
    \midrule
    GeoBridge (ours) & \underline{45.05} & \underline{49.05} & \underline{44.81} & \underline{48.76} & \underline{41.22} & \underline{43.54} &\underline{41.15}&\underline{43.41}\\
    GeoBridge++ (ours) &\textbf{58.01} &\textbf{71.01}&\textbf{57.88}&\textbf{70.66}&\textbf{55.86}&\textbf{69.29}&\textbf{59.24}&\textbf{71.64}\\
    \bottomrule
    \multirow{2}{*}{\textbf{Method}}
      & \multicolumn{2}{c}{\textbf{T2S}}
      & \multicolumn{2}{c}{\textbf{S2T}}
      & \multicolumn{2}{c}{\textbf{D2T}}
      & \multicolumn{2}{c}{\textbf{T2D}} \\
      \cmidrule(lr){2-3}\cmidrule(lr){4-5}\cmidrule(lr){6-7}\cmidrule(lr){8-9}
     & \textbf{R@1} & \textbf{AP} & \textbf{R@1} & \textbf{AP} & \textbf{R@1} & \textbf{AP} & \textbf{R@1} & \textbf{AP} \\
    \midrule
    Panorama-BEV~\cite{ye2024cross}   & 10.21 &12.70  & 18.70 & 20.56& 16.32&17.11 & 16.34&18.36 \\
    Sample4Geo~\cite{Sample4geo} & 16.82 & 13.64 & 18.87 & 17.78 & 29.51&31.17 & 15.56&29.68 \\
    AuxGeo~\cite{AuxGeo}& 13.74 &17.09 & 15.61 & 22.96 &9.34&12.69 & 14.70&17.40\\
    FRGeo~\cite{FRGeo}& 17.13 & 18.70& 14.35& 19.69& 13.69&19.34 & 14.38&15.00 \\
    HC-Net~\cite{HC-Net}   & 14.17& 15.05 & 11.79& 12.03 & 12.83&13.31 & 18.79&18.93 \\
    TransGeo~\cite{Transgeo} & 11.21 & 23.77 & 13.74 & 13.43 & 18.70&23.41 & 17.48&23.26 \\
    \midrule
    GeoBridge (ours) & \underline{38.87} & \underline{42.10} & \underline{39.21} & \underline{41.96}& \underline{41.23} & \underline{43.54} &\underline{41.15}&\underline{43.41}\\
    GeoBridge++ (ours) &\textbf{49.69} &\textbf{64.68}&\textbf{45.84}&\textbf{61.35}&\textbf{55.86}&\textbf{69.29}&\textbf{59.24}&\textbf{71.64}\\
    \bottomrule
  \end{tabular}}
  \vspace{-13pt}
\end{table}

\section{Experiment}
\subsection{Experimental setup}

\noindent \textbf{Implementation Details.} We adopt CLIP-L/14~\cite{radford2021learning} as the backbone for both GeoBridge and GeoBridge++, and GPT-4o~\cite{hurst2024gpt} as the LLM to generate textual descriptions and factual descriptions. Models are trained for \(200\) epochs using Adam with a learning rate of \(1\times10^{-5}\), cosine decay, a batch size of \(32\), and \(224\times224\) inputs on eight NVIDIA A800 GPUs. For GeoBridge++, we set \(\boldsymbol{\pi}=[0.25,0.25,0.25,0.25]\) and \(\lambda=0.0005\). All parameters are optimized during pre-training, while only the last three layers are fine-tuned on public benchmarks. For controlled evaluation, we additionally construct a map-augmented version of GeoLoc by introducing aligned static maps. GeoBridge++ is pre-trained and evaluated under this controlled setting, allowing us to analyze the contributions of location fact guidance and geo-semantic bridging without the multi-scale observation expansion introduced in GeoLoc-MM. GeoLoc-MM is then used for large-scale multi-view and multi-scale evaluation. Further details are in the supplementary material.

\textbf{Comparison Methods.} We benchmark our models against representative cross-view geo-localization methods spanning feature matching, representation learning, geometric alignment, and Transformer-based approaches. For drone--satellite retrieval, we adopt SAIG-D~\cite{SAIG-D}, DWDR~\cite{DWDR}, MBF~\cite{MBF}, MCCG~\cite{MCCG}, CCR~\cite{CCR}, SeGCN~\cite{SeGCN}, MEAN~\cite{MEAN}, DAC~\cite{DAC}, Sample4Geo~\cite{Sample4geo} and CGSI~\cite{CGSI}. For street-view--satellite retrieval, TransGeo~\cite{Transgeo}, FRGeo~\cite{FRGeo}, SAIG-D~\cite{SAIG-D}, VimGeo~\cite{VimGeo}, Sample4Geo~\cite{Sample4geo}, Panorama-BEV~\cite{ye2024cross}, and AuxGeo~\cite{AuxGeo} are included, spanning contrastive representation learning, geometry-aware alignment, and BEV-based modeling. For street-view--map retrieval, we compare with GeoDTR~\cite{zhang2023cross}, SAIG-D~\cite{SAIG-D}, Sample4Geo~\cite{Sample4geo}, and Panorama-BEV~\cite{ye2024cross}. Representative methods are further adapted to GeoLoc and GeoLoc-MM for unified cross-view evaluation, including newly introduced retrieval directions without dedicated prior methods. 

\textbf{Evaluation Metrics.} 
To evaluate geo-localization performance across viewpoints and modalities, we use Top-k recall (\(R@k\)) and average precision (\(AP\)) as primary metrics. \(R@k\) measures whether the ground-truth reference appears within the top \(k\) retrieved results, while \(AP\) summarizes ranking quality. We also include task-specific metrics as needed. For street-satellite-view retrieval, we additionally report \(R@1\%\) and \(Hit\). Here, \(R@1\%\) indicates whether the ground-truth lies within the top \(1\%\) of gallery images, and \(Hit\) indicates that the first retrieved reference image covers the query image (including the ground-truth). For cross-modal geo-localization, we complement \(R@k\) with the location recall \(L@k\), which refers to the proportion of retrieved results whose distance to the actual location is below a specified threshold. When geographic coordinates are unavailable, we report \(R@k\) only.

\begin{table*}[t]
  \centering
  \caption{Comparison on the GeoLoc-MM dataset with drone-to-satellite methods extended to other cross-view combinations. D, T, S, and M denote Drone, Street-View Panorama, Satellite, and Static Map, respectively. The best results are shown in bold, the second-best results are underlined.}
  \label{tab:GeoLoc-MM-drone}
  \scriptsize
  \resizebox{\textwidth}{!}{%
  \begin{tabular}{c*{4}{ccccccccc}}
    \toprule
    \multirow{2}{*}{\textbf{Method}}
      &\multirow{2}{*}{\textbf{Scale}}& \multicolumn{2}{c}{\textbf{D2S}}
      & \multicolumn{2}{c}{\textbf{S2D}}
      & \multicolumn{2}{c}{\textbf{D2T}}
      & \multicolumn{2}{c}{\textbf{T2D}} & \multicolumn{2}{c}{\textbf{D2M}}& \multicolumn{2}{c}{\textbf{M2D}}& \multicolumn{2}{c}{\textbf{S2M}}& \multicolumn{2}{c}{\textbf{M2S}}\\
      \cmidrule(lr){3-4}\cmidrule(lr){5-6}\cmidrule(lr){7-8}\cmidrule(lr){9-10}\cmidrule(lr){11-12}\cmidrule(lr){13-14}\cmidrule(lr){15-16}
      \cmidrule(lr){17-18}
     & &\textbf{R@1} & \textbf{AP} & \textbf{R@1} & \textbf{AP} & \textbf{R@1} & \textbf{AP} & \textbf{R@1} & \textbf{AP} & \textbf{R@1} & \textbf{AP}& \textbf{R@1} & \textbf{AP}& \textbf{R@1} & \textbf{AP}& \textbf{R@1} & \textbf{AP}\\
    \midrule
    Sample4Geo~\cite{Sample4geo} &\multirow{5}{*}{80}& \underline{25.95} & \underline{39.17} & \underline{24.20} & \underline{37.61} & \underline{27.51}& \underline{33.07}& \underline{20.27}& \underline{34.82} &14.61&28.05&14.33&27.46&\underline{20.53}&\underline{35.67}&\underline{18.89
    }&\underline{33.75}\\
    MEAN~\cite{MEAN}& & 23.26& 30.32 & 21.44 &30.96 &13.06&17.25& 2.89&7.34&12.61&24.16&12.18&24.09&12.74&24.18&13.11&25.28\\
    CAMP~\cite{CAMP} &  & 20.50& 28.24 & 17.61 & 22.87 &12.31 &14.25 & 13.12&20.71&\underline{17.26}&22.92&17.32&22.10&13.24&18.92&12.90&18.32 \\
    MCCG~\cite{MCCG}  & & 14.38 & 18.91 & 13.60& 17.78& 13.77&19.14 & 13.96&19.78&16.46&\underline{29.50}&\underline{17.52}&\underline{30.71}&15.51&27.46&15.71&29.03 \\
    GeoBridge++(ours)& &\textbf{55.01} &\textbf{69.33}&\textbf{55.69}&\textbf{70.09}&\textbf{58.88}&\textbf{73.02}&\textbf{57.44}&\textbf{71.74}&\textbf{51.46}&\textbf{66.54}&\textbf{47.72}&\textbf{63.31}&\textbf{50.58}&\textbf{66.33}&\textbf{46.51}&\textbf{63.04}\\
    \midrule
    Sample4Geo~\cite{Sample4geo} &\multirow{5}{*}{100}& \underline{25.85} & \underline{39.41} & \underline{25.15} & \underline{38.01} & \underline{29.04}& \underline{33.46}& \underline{20.66}& \underline{34.40}&15.41&28.30&15.38&29.36&\underline{20.82}&\underline{35.16}&\underline{18.37}&\underline{33.50}\\
    MEAN~\cite{MEAN}& & 23.59& 30.54 & 21.37 &30.75 &13.48&17.89& 2.67&7.36&12.45&24.84&12.40&24.64&12.53&24.00&13.90&25.58\\
    CAMP~\cite{CAMP} &  & 23.36 & 32.65 & 20.61 & 27.59 &12.43 &14.58 & 13.75&20.08&\underline{17.41}&22.93&17.01&22.50&14.61&20.81&14.50&20.32 \\
    MCCG~\cite{MCCG}  & & 16.61 & 19.55 & 15.12& 19.89 & 14.50&20.56 & 15.19&22.67&16.42&\underline{29.44}&\underline{19.94}&\underline{30.22}&16.11&28.45&16.67&30.13 \\
    GeoBridge++(ours)& &\textbf{54.21} &\textbf{68.20}&\textbf{53.97}&\textbf{68.49}&\textbf{53.57}&\textbf{67.79}&\textbf{54.08}&\textbf{68.19}&\textbf{51.20}&\textbf{66.46}&\textbf{48.64}&\textbf{63.99}&\textbf{48.71}&\textbf{64.40}&\textbf{45.22}&\textbf{61.20}\\
    \midrule
    Sample4Geo~\cite{Sample4geo} &\multirow{5}{*}{120}& \underline{25.27} & \underline{39.26} & \underline{25.64} & \underline{38.01} & \underline{28.09}& \underline{32.22}& \underline{20.81}& \underline{34.23}&15.50&28.55&14.16&27.42&\underline{19.59}&\underline{32.82}&\underline{17.10}&\underline{31.85}\\
    MEAN~\cite{MEAN}& & 23.74& 32.00 & 21.91 &31.30 &15.40&17.94& 7.13&16.11 &11.16&22.96&11.39&23.20&12.41&23.44&13.84&25.90\\
    CAMP~\cite{CAMP} &  & 23.99 & 32.90 & 20.04 & 27.65 &12.27 &14.39 & 13.78&20.18&\underline{16.91}&22.14&16.63&21.99&13.25&19.59&14.76&19.07 \\
    MCCG~\cite{MCCG}  & & 15.26 & 19.60 & 15.71& 20.09 & 14.52&20.01 &15.42&22.87&16.50&\underline{29.74}&\underline{19.34}&\underline{29.80}&16.62&29.38&15.91&28.64 \\
    GeoBridge++(ours)& &\textbf{53.72} &\textbf{68.26}&\textbf{56.25}&\textbf{70.44}&\textbf{52.25}&\textbf{67.85}&\textbf{54.12}&\textbf{69.20}&\textbf{46.58}&\textbf{62.54}&\textbf{41.71}&\textbf{58.64}&\textbf{51.01}&\textbf{66.93}&\textbf{42.08}&\textbf{59.33}\\
    \midrule
    Sample4Geo~\cite{Sample4geo} &\multirow{5}{*}{150}& \underline{24.90} & \underline{38.45} & \underline{23.32} & \underline{36.10} & \underline{27.44}& \underline{32.86}& \underline{20.33}& \underline{34.75}&13.51&\underline{28.37}&14.91&27.69&\underline{20.82}&\underline{35.74}&\underline{18.39}&\underline{33.12}\\
    MEAN~\cite{MEAN}& & 23.23& 31.74 & 22.01 &31.90 &15.05&17.14& 7.89&16.27&11.52&23.92&11.90&23.83&12.69&23.68&12.71&23.83\\
    CAMP~\cite{CAMP} &  & 22.36 & 30.3 & 19.93 & 27.03&12.00 &14.80 & 14.42&21.84&\underline{17.46}&22.93&16.43&21.51&13.98&19.20&14.47&19.55 \\
    MCCG~\cite{MCCG}  & & 14.26 & 18.32 & 13.11& 18.68 & 13.87&19.28 & 15.57&22.31&15.74&27.64&\underline{18.91}&\underline{27.74}&16.90&29.30&15.93&28.34 \\
    GeoBridge++(ours)& &\textbf{58.64}&\textbf{71.82}&\textbf{59.91}&\textbf{73.46}&\textbf{52.71}&\textbf{68.48}&\textbf{54.34}&\textbf{69.69}&\textbf{47.72}&\textbf{63.61}&\textbf{43.61}&\textbf{60.36}&\textbf{53.41}&\textbf{68.72}&\textbf{44.96}&\textbf{61.42}\\
    \midrule
    Sample4Geo~\cite{Sample4geo} &\multirow{5}{*}{180}& \underline{25.23} & \underline{38.64} & \underline{24.86} & \underline{38.55} & \underline{28.48}& \underline{34.15}& \underline{21.14}& \underline{35.88}&13.76&\underline{27.98}&13.47&26.68&\underline{20.84}&\underline{36.51}&\underline{19.63}&\underline{33.85}\\
    MEAN~\cite{MEAN}& & 24.78& 35.97 & 24.34 &34.99 &14.07&16.72& 9.94&19.85&15.52&27.70&14.53&26.42&13.08&24.17&13.58&24.36\\
    CAMP~\cite{CAMP} &  & 22.32 & 30.83 & 20.32 & 27.17 &14.38 &16.85 & 13.50&20.93&\underline{17.41}&22.34&\underline{17.94}&22.10&12.69&18.80&14.24&18.87 \\
    MCCG~\cite{MCCG} &  & 14.85 & 19.01 & 13.08& 18.75 & 14.97&22.41 & 16.02&22.32&15.13&16.04&18.74&\underline{27.65
    }&17.20&30.91&16.48&28.88 \\
    GeoBridge++(ours)& &\textbf{58.08 }&\textbf{72.09}&\textbf{60.15}&\textbf{73.83}&\textbf{50.49}&\textbf{66.34}&\textbf{52.12}&\textbf{67.31}&\textbf{49.26}&\textbf{64.91}&\textbf{46.86}&\textbf{62.83}&\textbf{51.57}&\textbf{67.36}&\textbf{46.14}&\textbf{62.69}\\
    \midrule
    Sample4Geo~\cite{Sample4geo} &\multirow{5}{*}{200}& \underline{26.29} & \underline{40.89} & \underline{25.82} & \underline{38.45} & \underline{29.79}& \underline{34.42}& \underline{21.13}& \underline{25.37}&14.75&\underline{28.71}&13.80&27.46&\underline{21.19}&\underline{37.01}&\underline{19.71}&\underline{34.43}\\
    MEAN~\cite{MEAN}& & 25.77& 36.98 & 24.80 &35.33 &8.13&18.88& 14.21&17.99&13.91&24.15&13.86&25.32&12.86&23.50&13.44&24.91\\
    CAMP~\cite{CAMP} &  & 20.09 & 29.74 & 18.97 & 26.41 &12.98 &16.49 & 13.16&20.20&\underline{17.69}&22.99&17.06&22.16&13.61&19.37&14.84&20.09 \\
    MCCG~\cite{MCCG} &  & 14.26 & 19.13 & 13.6& 18.44 & 14.79&22.81 & 16.41&22.42&17.55&18.49&\underline{19.34}&\underline{28.52}&17.00&28.67&15.10&28.85 \\
    GeoBridge++(ours)& &\textbf{53.48 }&\textbf{68.09}&\textbf{53.39}&\textbf{68.75}&\textbf{49.46}&\textbf{66.30}&\textbf{52.64}&\textbf{68.24}&\textbf{46.21}&\textbf{62.55}&\textbf{44.90}&\textbf{61.45}&\textbf{50.14}&\textbf{66.48}&\textbf{45.78}&\textbf{62.96}\\
    \bottomrule
  \end{tabular}}
  \vspace{-13pt}
\end{table*}

\begin{figure*}[t]
    \centering

    \subfloat[Street to Satellite\label{fig:sub1}]{%
        \includegraphics[width=0.3\textwidth]{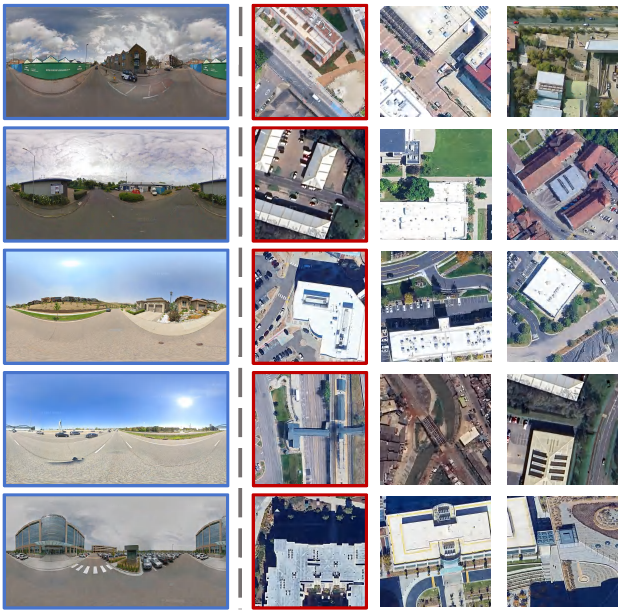}
    }
    \hspace{0.0\textwidth}
    \subfloat[Satellite to Drone\label{fig:sub2}]{%
        \includegraphics[width=0.243\textwidth]{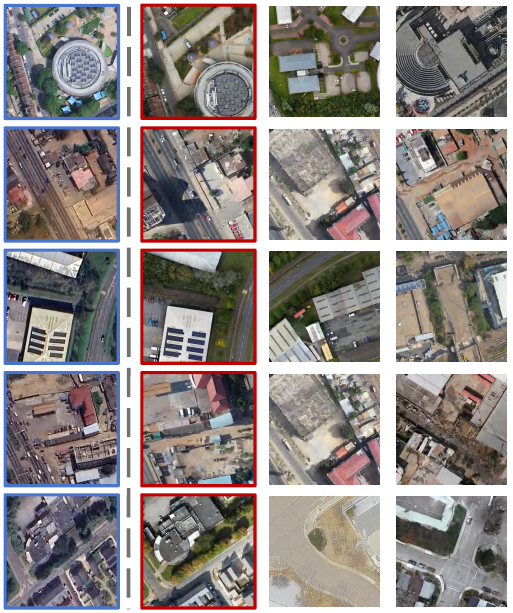}
    }
    \hspace{0.0\textwidth}
    \subfloat[Drone to Street\label{fig:sub3}]{%
        \includegraphics[width=0.400\textwidth]{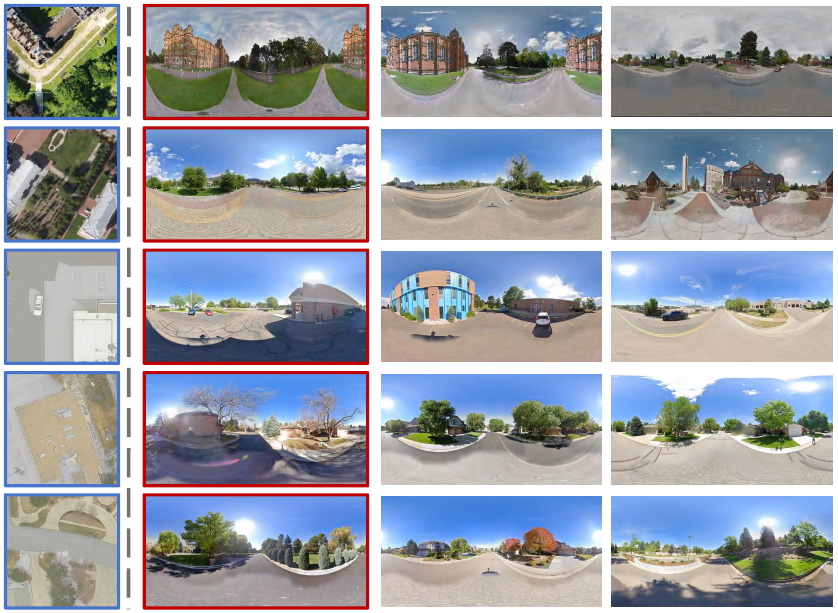}
    }

    \caption{Qualitative results of GeoBridge on GeoLoc. Blue boxes denote queries, 
    and red boxes indicate true matches.}
    \label{fig:triple}
    \vspace{-13pt}
\end{figure*}

\subsection{Cross-View Geo-localization Performance}
\subsubsection{Drone-Satellite Retrieval}

For the drone–satellite matching task, we evaluate GeoBridge and GeoBridge++ on University-1652~\cite{zheng2020university} and SUES-200~\cite{zhu2023sues}, with results reported in Table~\ref{tab:university} and~\ref{tab:sues}. GeoBridge achieves competitive or leading \(R@1\) and \(AP\) performance on both datasets. Although DAC performs strongly, its direct image-to-image alignment is less effective at modeling cross-view semantic consistency and resolving spatial ambiguities. In contrast, GeoBridge uses semantic-anchor mechanism to align multi-view imagery in a shared semantic space, yielding clear gains in bidirectional retrieval on University-1652 and competitive performance across SUES-200 settings. Building upon GeoBridge, GeoBridge++ further improves performance across nearly all evaluation settings on both University-1652 dataset and SUES-200 dataset, with particularly clear gains under the more challenging low-altitude settings. These improvements suggest that geographic and visual facts provide complementary location-level cues that enhance the semantic representations learned by GeoBridge. The consistent gains across datasets, retrieval directions, and flight altitudes further indicate its robustness to variations in viewpoint and observation scale. 

\begin{table*}[t]
  \centering
  \caption{Comparison on the GeoLoc-MM dataset with street-to-satellite methods extended to other cross-view combinations D, T, S, and M denote Drone, Street-View Panorama, Satellite, and Static Map, respectively. The best results are shown in bold, the second-best results are underlined.}
  \label{tab:GeoLoc-MM-street}
  \scriptsize
  \resizebox{\textwidth}{!}{%
  \begin{tabular}{c*{4}{ccccccccc}}
    \toprule
    \multirow{2}{*}{\textbf{Method}}
      &\multirow{2}{*}{\textbf{Scale}}& \multicolumn{2}{c}{\textbf{T2S}}
      & \multicolumn{2}{c}{\textbf{S2T}}
      & \multicolumn{2}{c}{\textbf{D2T}}
      & \multicolumn{2}{c}{\textbf{T2D}} & \multicolumn{2}{c}{\textbf{T2M}}& \multicolumn{2}{c}{\textbf{M2T}}& \multicolumn{2}{c}{\textbf{S2M}}& \multicolumn{2}{c}{\textbf{M2S}}\\
      \cmidrule(lr){3-4}\cmidrule(lr){5-6}\cmidrule(lr){7-8}\cmidrule(lr){9-10}\cmidrule(lr){11-12}\cmidrule(lr){13-14}\cmidrule(lr){15-16}
      \cmidrule(lr){17-18}
     & &\textbf{R@1} & \textbf{AP} & \textbf{R@1} & \textbf{AP} & \textbf{R@1} & \textbf{AP} & \textbf{R@1} & \textbf{AP}& \textbf{R@1} & \textbf{AP}& \textbf{R@1} & \textbf{AP}& \textbf{R@1} & \textbf{AP}& \textbf{R@1} & \textbf{AP} \\
    \midrule
    Panorama-BEV~\cite{ye2024cross}   &\multirow{5}{*}{80}& 12.24 &15.91  & \underline{20.41} & \underline{23.01}& 18.18&22.18 & 16.33&20.62&10.99&13.79&10.85&13.52&14.48&21.01&14.35&20.46 \\
    Sample4Geo~\cite{Sample4geo} && 16.70 & 20.48 & 15.52 & 21.15 & \underline{27.51}&\underline{33.07} & \underline{20.27}&\underline{34.82}&13.13&\underline{26.16}&11.15&23.23&\underline{20.53}&\underline{35.67}&18.89&\underline{33.75} \\
    AuxGeo~\cite{AuxGeo}&& 15.85 &22.63 & 14.76 & 20.86 &7.51&11.85 & 10.55&15.85&\underline{15.17}&18.41&\underline{11.39}&\underline{24.11}&12.47&16.86&12.91&14.13\\
    FRGeo~\cite{FRGeo}&& \underline{18.79} & \underline{24.47}& 16.74& 21.04& 15.33&21.98 & 16.35&17.07&6.31&15.07&6.20&14.62&13.39&19.68&\underline{18.98}&25.11 \\
    GeoBridge++(ours)& &\textbf{50.19 }&\textbf{65.80}&\textbf{51.87}&\textbf{67.23}&\textbf{58.88}&\textbf{73.02}&\textbf{57.44}&\textbf{71.74}&\textbf{51.56}&\textbf{66.58}&\textbf{48.29}&\textbf{63.85}&\textbf{50.58}&\textbf{66.33}&\textbf{46.51}&\textbf{63.04}\\
    \midrule
    Panorama-BEV~\cite{ye2024cross}   &\multirow{5}{*}{100}& 13.26&17.15  & \underline{21.08} & \underline{25.37}& 18.65&22.65 & 16.60&20.27&10.05&13.66&10.90&13.04&14.51&21.48&14.43&20.81 \\
    Sample4Geo~\cite{Sample4geo} && 16.72 & 20.54& 16.98& 23.42 & \underline{29.04}&\underline{33.46} & \underline{20.66}&\underline{34.40}&13.77&\underline{26.51}&11.68&23.84&\underline{20.82}&\underline{35.16}&18.37&\underline{33.50} \\
    AuxGeo~\cite{AuxGeo}&& 15.60 &23.63 & 14.32 & 21.55 &8.07&13.71 & 10.69&15.55&\underline{16.47}&20.07&\underline{12.97}&\underline{24.20}&12.11&16.89&12.09&14.65\\
    FRGeo~\cite{FRGeo}&& \underline{18.18} & \underline{24.51}& 16.45& 21.93& 14.30&19.22 & 16.21&17.19&7.60&16.96&6.44&15.51&12.56&19.91&\underline{18.52}&25.05 \\
    GeoBridge++(ours)& &\textbf{49.03 }&\textbf{64.48}&\textbf{48.04}&\textbf{63.57}&\textbf{53.57}&\textbf{67.79}&\textbf{54.08}&\textbf{68.19}&\textbf{44.06}&\textbf{59.96}&\textbf{41.61}&\textbf{57.93}&\textbf{48.71}&\textbf{64.40}&\textbf{45.22}&\textbf{61.20}\\
    \midrule
    Panorama-BEV~\cite{ye2024cross}   &\multirow{5}{*}{120}& 13.90 &17.87  & \underline{20.16} & 24.93& 19.49&23.88 & 17.41&21.21&11.23&13.88&12.17&16.66&14.97&21.76&14.29&20.67 \\
    Sample4Geo~\cite{Sample4geo} && 17.44 & 21.01 & 16.78 & 23.98 & \underline{28.09}&\underline{32.22} & \underline{20.81}&\underline{34.23}&\underline{13.51}&\underline{26.04}&12.34&\underline{25.61}&\underline{19.59}&\underline{32.82}&17.10&\underline{31.85} \\
    AuxGeo~\cite{AuxGeo}&& 14.83 &22.71 & 15.30 & 22.17 &10.90&14.95 & 14.21&18.28 &12.35&24.96&10.52&22.43&15.20&18.18&10.70&14.66\\
    FRGeo~\cite{FRGeo}&& \underline{19.66} & \underline{26.04}& 18.80& \underline{25.28}& 14.47&20.85 & 18.06&20.88 &9.46&17.54&\underline{13.19}&16.47&14.66&21.32&\underline{17.13}& 24.36\\
    GeoBridge++(ours)& &\textbf{49.52} &\textbf{65.45}&\textbf{49.14}&\textbf{65.02}&\textbf{52.25}&\textbf{67.85}&\textbf{54.12}&\textbf{69.20}&\textbf{45.11}&\textbf{61.38}&\textbf{38.08}&\textbf{55.54}&\textbf{51.01}&\textbf{66.93}&\textbf{42.08}&\textbf{59.33}\\
    \midrule
    Panorama-BEV~\cite{ye2024cross}   &\multirow{5}{*}{150}& 13.24 &17.55  & \underline{20.45}& 24.27& 19.13&22.90 & 17.58&21.69&11.01&13.02&11.47&16.28&15.53&22.69&14.34&21.39 \\
    Sample4Geo~\cite{Sample4geo} && 17.03 & 21.11 & 17.62 & 24.07 & \underline{27.44}&\underline{32.86} & \underline{20.33}&\underline{34.75 }&\underline{13.62}&\underline{26.00}&11.87&\underline{23.53}&\underline{20.82}&\underline{35.74}&\underline{18.39}&\underline{33.12} \\
    AuxGeo~\cite{AuxGeo}&& 14.72 &22.62 & 15.48 & 22.07 &10.80&14.84 & 14.79&18.11&12.28&24.38&10.14&22.96&15.84&18.46&10.45&14.10\\
    FRGeo~\cite{FRGeo}&& \underline{17.49} & \underline{23.38}& 18.53& \underline{25.11}& 15.77&22.94 & 18.58&20.94&9.93&17.89&\underline{13.01}&16.46&14.45&21.84&18.32&24.99 \\
    GeoBridge++(ours)& &\textbf{50.55}&\textbf{65.65}&\textbf{50.43}&\textbf{65.94}&\textbf{52.71}&\textbf{68.48}&\textbf{54.34}&\textbf{69.69}&\textbf{43.78}&\textbf{60.33}&\textbf{37.97}&\textbf{55.39}&\textbf{53.41}&\textbf{68.72}&\textbf{44.96}&\textbf{61.42}\\
    \midrule
    Panorama-BEV~\cite{ye2024cross}   &\multirow{5}{*}{180}& 14.82 &18.52  & 20.81 & 25.24& 19.38&23.42 & 18.66&23.37&12.22&13.20&12.15&16.59&16.03&24.48&15.66&23.99 \\
    Sample4Geo~\cite{Sample4geo} && 16.97 & 21.23 & 15.77 & 22.12 & \underline{28.48}&\underline{34.15} & \underline{21.14}&\underline{35.88}&\underline{13.07} &\underline{25.58}&12.27&\underline{25.39}&\underline{20.84}&\underline{36.51}&19.63&\underline{33.85} \\
    AuxGeo~\cite{AuxGeo}&& 16.20&21.89 & 15.73 & 23.67 &13.86&16.27& 15.01&19.53&11.79&13.21&\underline{12.80}&16.23&14.90&18.60&12.33&19.35\\
    FRGeo~\cite{FRGeo}&& \underline{19.09}& \underline{25.75}& \underline{20.91}& \underline{26.69}& 16.81&24.13 & 17.83&17.62&10.07&18.28&9.19&14.49&13.86&21.19&\underline{19.88}&27.23 \\
    GeoBridge++(ours)& &\textbf{49.16} &\textbf{64.95}&\textbf{48.66}&\textbf{64.47}&\textbf{50.49}&\textbf{66.34}&\textbf{52.12}&\textbf{67.31}&\textbf{44.16}&\textbf{60.08}&\textbf{39.00}&\textbf{56.73}&\textbf{51.57}&\textbf{67.36}&\textbf{46.14}&\textbf{62.69}\\
    \midrule
    Panorama-BEV~\cite{ye2024cross}   &\multirow{5}{*}{200}& 12.66 &16.19  & \underline{18.78} & 22.62& 18.76&22.39 & 16.72&21.80&10.97&12.33&10.57&15.96&15.94&23.05&15.64&23.86 \\
    Sample4Geo~\cite{Sample4geo} && 17.38 & 22.08& 16.78 & \underline{23.84} & \underline{29.79}&\underline{34.42} & \underline{21.13}&\underline{25.37}&\underline{14.79}&\underline{26.16}&\underline{13.47}&\underline{26.60}&\underline{21.19}&\underline{37.01}&\underline{19.71}& \underline{34.43}\\
    AuxGeo~\cite{AuxGeo}&& 14.07 &18.45 & 14.39 & 22.50 &13.47&16.35 & 14.35&18.41&14.03&19.66&11.5&20.6&14.94&18.25&11.04&17.74\\
    FRGeo~\cite{FRGeo}&& \underline{17.50} & \underline{23.57}& 17.89& 22.36& 15.55&23.02 & 17.35&18.11&10.76&19.42&8.08&15.12&13.10&21.89&17.04&25.58\\
    GeoBridge++(ours)& &\textbf{46.73}&\textbf{62.95}&\textbf{45.26}&\textbf{61.84}&\textbf{49.46}&\textbf{66.30}&\textbf{52.64}&\textbf{68.24}&\textbf{43.18}&\textbf{59.71}&\textbf{40.21}&\textbf{57.77}&\textbf{50.14}&\textbf{66.48}&\textbf{45.78}&\textbf{62.96}\\
    \bottomrule
  \end{tabular}}
  \vspace{-13pt}
\end{table*}

\begin{figure*}[t]
    \centering

    \subfloat[Drone to Map\label{sub1}]{%
        \includegraphics[width=0.211\textwidth]{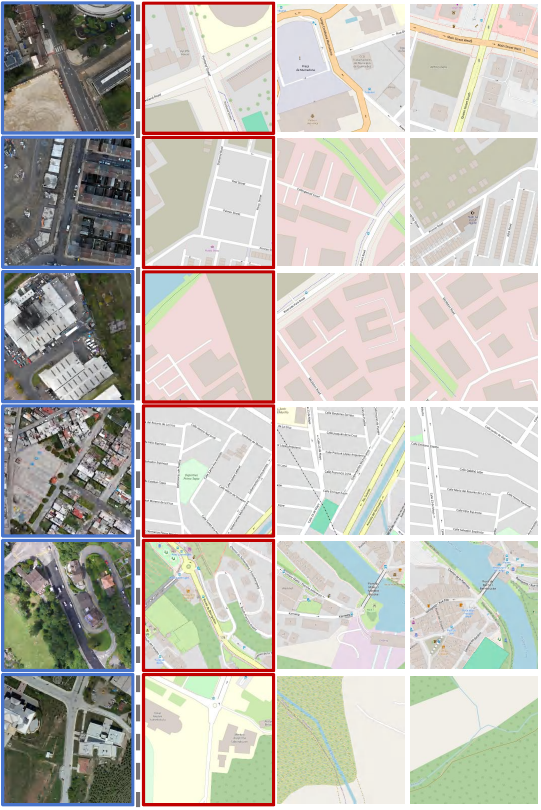}
    }
    \hspace{0.0\textwidth}
    \subfloat[Map to Satellite\label{sub2}]{%
        \includegraphics[width=0.211\textwidth]{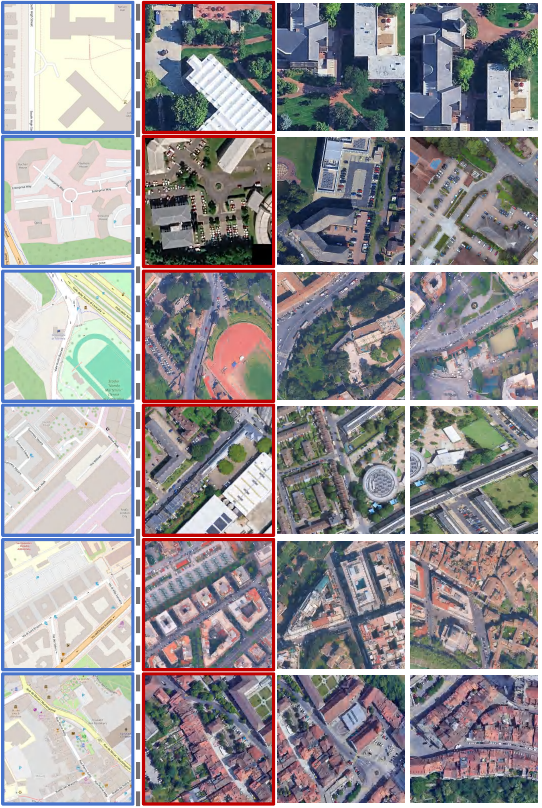}
    }
    \hspace{0.0\textwidth}
    \subfloat[Satellite to Street\label{sub3}]{%
        \includegraphics[width=0.27\textwidth]{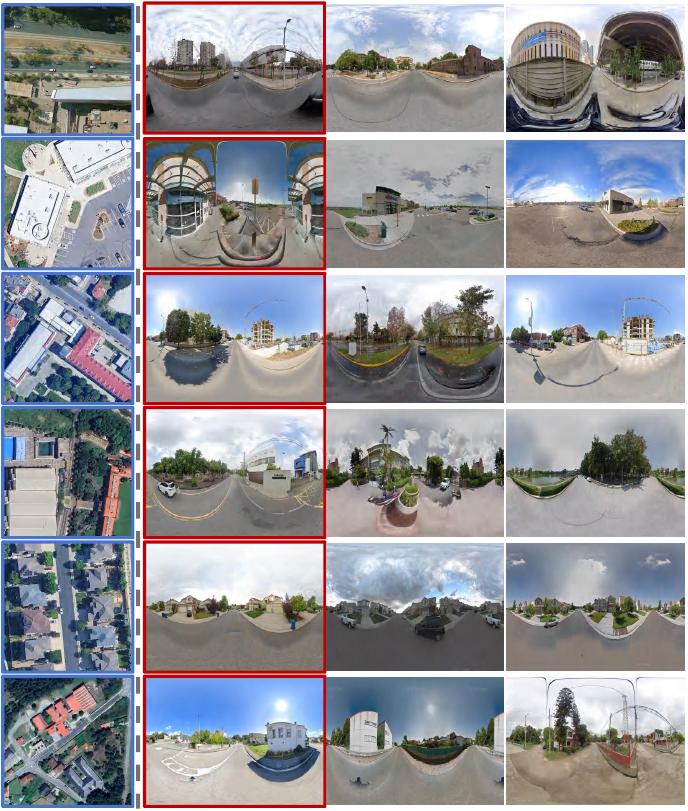}
    }
    \hspace{0.0\textwidth}
    \subfloat[Street to Drone\label{sub4}]{%
        \includegraphics[width=0.232\textwidth]{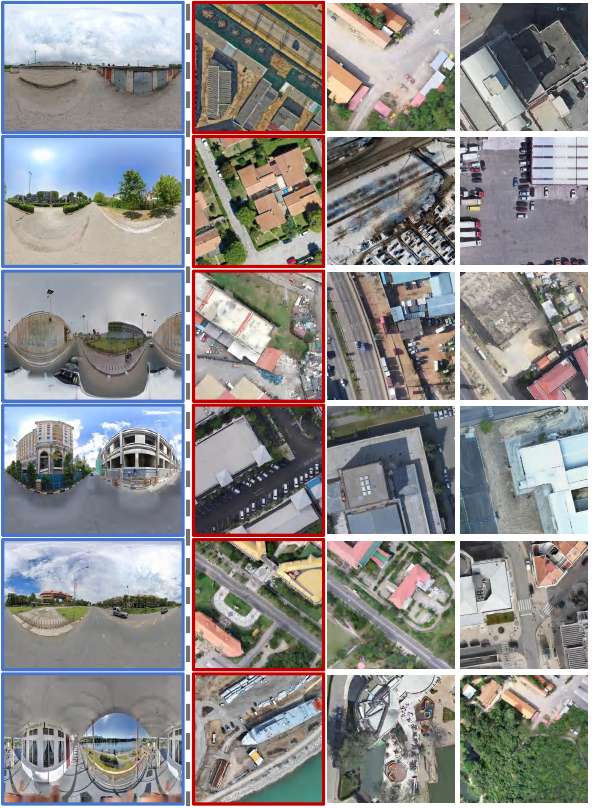}
    }

    \caption{Qualitative results of GeoBridge++ on GeoLoc-MM. Blue boxes denote queries, and red boxes indicate truth matches. The six rows from top to bottom correspond to spatial extents ranging from 80 to 200~\(m\).}
    \label{fig:quad}
    \vspace{-13pt}
\end{figure*}

\subsubsection{Street-Satellite Retrieval}

For street-to-satellite cross-view geo-localization, we evaluate GeoBridge and GeoBridge++ on CVUSA~\cite{CVUSA}, VIGOR~\cite{zhu2021vigor}, and CVGlobal~\cite{ye2024cross} datasets under the standard evaluation protocols of each dataset. The results are reported in Table~\ref{tab:street_satellite} and Table~\ref{CVG:street_satellite}.

On CVUSA and VIGOR, GeoBridge consistently performs favorably across different evaluation settings, validating the effectiveness of semantic anchors for establishing correspondence between street-level and overhead observations. The advantage is particularly evident under the VIGOR-Cross setting, where the geographic distributions of query and reference images differ, indicating improved generalization under cross-region distribution shifts. GeoBridge++ further strengthens the learned representation by incorporating geographically grounded and visually stable location facts, leading to consistent improvements over GeoBridge across the considered settings. These results suggest that fact guidance complements semantic-anchor alignment by providing additional location-level evidence for resolving ambiguities between heterogeneous views.

On CVGlobal, GeoBridge remains competitive on regional val and exhibits stronger performance under the more challenging regional test setting. GeoBridge++ further improves GeoBridge across all evaluation metrics, with more pronounced gains on regional test than on the relatively saturated regional val split. This trend suggests that fact guidance becomes beneficial under geographic distribution shifts, where appearance cues are less transferable across regions. By incorporating geographically grounded and visually stable location evidence, GeoBridge++ further reduces region-specific appearance dependence and strengthens cross-view location consistency. 

\begin{table}[t]
  \centering
  \caption{Comparison results on the RSIEval~\cite{HU2025272} dataset. The best results are shown in bold, the second-best results are underlined.}
  \label{tab:RSIEval}
  \begin{tabular}{ccccc}
    \toprule
    \textbf{Method}& &\textbf{R@1}& \textbf{R@5}&\textbf{R@10}\\
    \midrule
    ViLT~\cite{kim2021vilt}&& 2.00& 16.00&38.00\\
    BLIP-B~\cite{li2022blip}& &0.00 & 1.00 & 2.00\\
    EVA2-CLIP-B/16~\cite{fang2024eva}&&15.00 &38.00 &54.00\\
    EVA2-CLIP-L/14~\cite{fang2024eva} &&19.00&43.00&55.00\\
    CLIP-L/14~\cite{radford2021learning}& &25.00&60.00&74.00\\
    CLIP-B/16~\cite{radford2021learning}&&18.00&51.00&61.00\\
    CrossText2Loc\cite{ye2025cross}&&12.00&36.00&67.00\\
    
    \midrule
    GeoBridge (ours) &&\underline{29.00}& \underline{71.00}& \underline{88.00}\\
    GeoBridge++ (ours) && \textbf{32.00}&\textbf{73.00}&\textbf{89.00}\\
    \bottomrule
  \end{tabular}
  \vspace{-5pt}
\end{table}

\begin{figure}[t]
  \centering
   \includegraphics[width=1\linewidth]{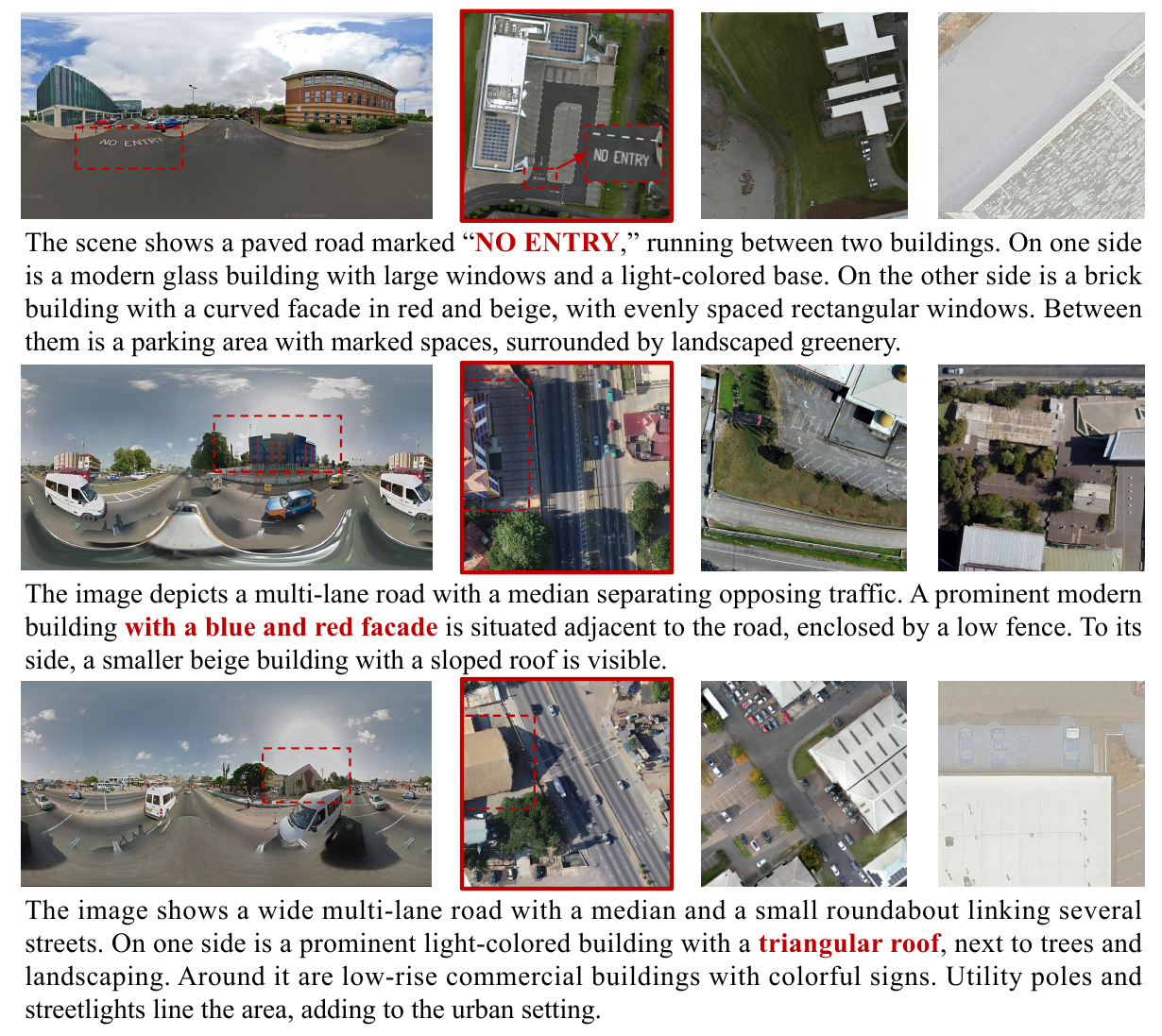}
   \caption{Qualitative cross-modal geo-localization results of GeoBridge on GeoLoc. Using street view descriptions to match drone perspectives, the top three results are reported; red boxes indicate correct matches.}
   \label{fig:language}
    \vspace{-13pt}
\end{figure}

\begin{figure}[t]
  \centering
   \includegraphics[width=1\linewidth]{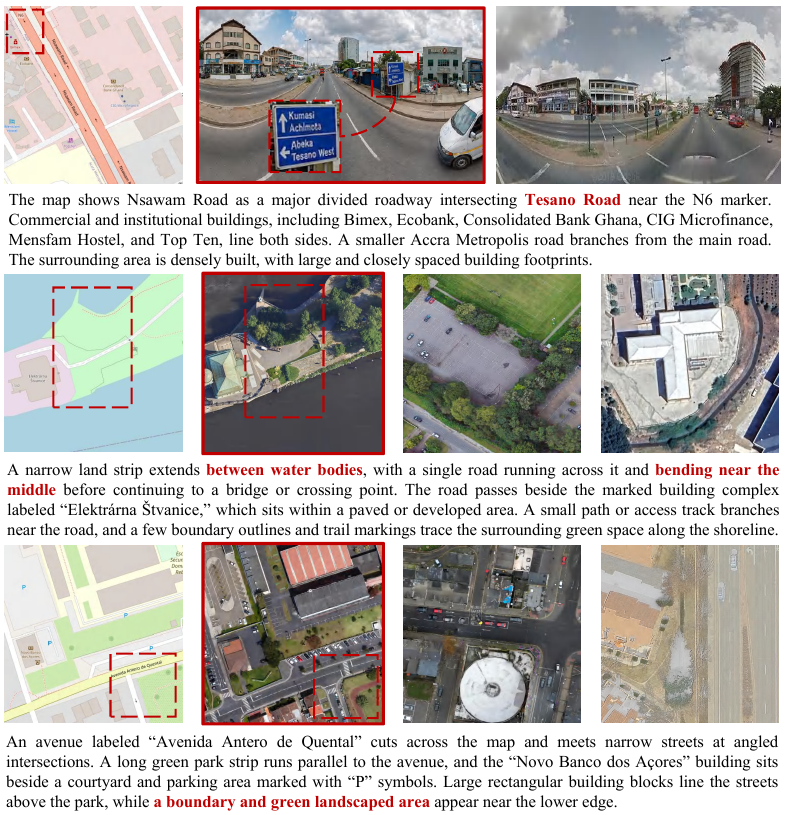}
   \caption{Qualitative cross-modal geo-localization results of GeoBridge++ on map-augmented GeoLoc. Descriptions generated from map observations are used to retrieve images from street-view, satellite, and drone perspectives. The top-ranked retrieved results are shown, with correct matches highlighted in red.}
   \label{fig:map}
    \vspace{-13pt}
\end{figure}

\begin{table*}[ht]
  \centering
  \caption{Comparison of cross-modal geo-localization methods on the GeoLoc dataset. Descriptions are generated from a single view; comparative results include two additional viewpoints. Best results are in bold; second-best are underlined.}
  \label{tab:streetlang-image}
  \resizebox{\textwidth}{!}{%
  \begin{tabular}{c c c c c  c c c c c c c c }
    \toprule
    \multirow{3}{*}{\textbf{Method}} & \multicolumn{4}{c}{\textbf{Street Description}}& \multicolumn{4}{c}{\textbf{Satellite Description}}& \multicolumn{4}{c}{\textbf{Drone Description}} \\
    \cmidrule(lr){2-5}\cmidrule(lr){6-9}\cmidrule(lr){10-13}
     &\multicolumn{2}{c}{\textbf{Satellite Image}}& \multicolumn{2}{c}{\textbf{Drone Image}}&\multicolumn{2}{c}{\textbf{Street Image}}& \multicolumn{2}{c}{\textbf{Drone Image}} &\multicolumn{2}{c}{\textbf{Street Image}}& \multicolumn{2}{c}{\textbf{Satellite Image}} \\
    \cmidrule(lr){2-3}\cmidrule(lr){4-5}\cmidrule(lr){6-7}\cmidrule(lr){8-9}\cmidrule(lr){10-11}\cmidrule(lr){12-13}

    & \textbf{R@1} & \textbf{L@50} & \textbf{R@1} & \textbf{L@50} & \textbf{R@1} & \textbf{L@50} & \textbf{R@1} & \textbf{L@50} & \textbf{R@1} & \textbf{L@50} & \textbf{R@1} & \textbf{L@50}\\
    \midrule
    ViLT~\cite{kim2021vilt}& 1.40& 6.41& 1.05&5.72&6.3&13.21&1.05&5.96  &6.20&13.49&1.16&5.76\\
    BLIP-B~\cite{li2022blip}& 0.84 & 4.6 & 0.78&  5.42 &0.93&5.06&0.73&5.59   &0.82&5.14&0.95&5.64\\
    EVA2-CLIP-B/16~\cite{fang2024eva}&1.01 &5.28 &0.92 &4.30 &1.21&5.74&0.90&4.26  &1.18&5.63&0.99&5.23\\
    EVA2-CLIP-L/14~\cite{fang2024eva} &1.16&4.97&1.83&6.95&1.23&5.57&1.23&5.42   &1.21&5.55&1.18&4.88\\
    CLIP-L/14~\cite{radford2021learning} &2.71&\underline{7.81}&2.02&7.96&12.20&20.71&1.89&7.21   &9.18&17.38&2.88&\underline{7.83}\\
    CLIP-B/16~\cite{radford2021learning}&1.91&7.49&1.79&8.04&6.91&14.93&1.85&7.36   &6.54&13.32&1.72&6.99\\
    CrossText2Loc~\cite{ye2025cross}&2.62&7.76&2.15&7.68&12.91&22.37&2.35&8.04  &9.25&17.83&2.11&7.4\\
    
    \midrule
    GeoBridge(ours) & \underline{5.68}& 7.61&\underline{6.10} &\underline{8.76} &\underline{21.58}&\underline{25.53}&\underline{2.99}&\underline{8.85}  &\underline{20.96}&\underline{24.73}&\underline{2.93}&7.66\\
    GeoBridge++(ours) &\textbf{7.10}&\textbf{10.29}&\textbf{11.91}&\textbf{15.34}&\textbf{23.16}&\textbf{27.88}&\textbf{7.48}&\textbf{14.30}&\textbf{23.69}&\textbf{27.45}&\textbf{6.54}&\textbf{11.40}\\
    \bottomrule
  \end{tabular}
  }
    \vspace{-5pt}
  \end{table*}

\begin{table}[ht]
  \centering
  \caption{Comparison of GeoBridge++ with cross-modal geo-localization methods on the map-augmented GeoLoc dataset. Desc. denotes descriptions generated from a single source view, with retrieval performed over images from other target views. The best results are shown in bold, and the second-best results are underlined.}
  \label{tab:mag-image}
  \resizebox{\columnwidth}{!}{%
  \begin{tabular}{ccc|ccccc}
    \toprule
    \textbf{\normalsize Desc.} & \textbf{\normalsize Target} & \textbf{\normalsize Metric}
    & \textbf{\makecell{\footnotesize ViLT\\~\cite{kim2021vilt}}}
    & \textbf{\makecell{\footnotesize CLIP\\-L/14~\cite{radford2021learning}}}
    & \textbf{\makecell{\footnotesize CLIP\\-B/16~\cite{radford2021learning}}}
    & \textbf{\makecell{\footnotesize CrossText\\2Loc~\cite{ye2025cross}}}
    & \textbf{\makecell{\footnotesize GeoBridge++ \\(Ours)}} \\
    \midrule

     \multirow{6}{*}{\textbf{\normalsize Map}}
    & \multirow{2}{*}{\textbf{\normalsize Street}}
    & \textbf{\normalsize R@1}
    & \normalsize 7.41 
    & \normalsize 12.37 & \normalsize 8.54 & \normalsize \underline{13.53} &  \textbf{\normalsize 23.48} \\
    &
    & \textbf{\normalsize L@50}
    & \normalsize 14.57 
    & \normalsize 22.84 & \normalsize 16.55 & \underline{\normalsize 23.24} & \textbf{\normalsize 27.49}\\

    \cmidrule(lr){2-8}
    &
    \multirow{2}{*}{\textbf{\normalsize Satellite}}
    & \textbf{\normalsize R@1}
    & \normalsize 1.73
    & \underline{\normalsize 2.65} & \normalsize 2.33 & \normalsize 2.60 & \textbf{\normalsize 8.99} \\
    &
    & \textbf{\normalsize L@50}
    & \normalsize 6.63  
    & \normalsize 7.30 & \normalsize 7.15 & \underline{\normalsize 7.55} & \textbf{\normalsize 10.60} \\

    \cmidrule(lr){2-8}
    &
    \multirow{2}{*}{\textbf{\normalsize Drone}}
    & \textbf{\normalsize R@1}
    & \normalsize 1.49 
    & \underline{\normalsize 2.57}& \normalsize 1.95 & \normalsize 2.44 & \textbf{\normalsize 9.20}\\
    &
    & \textbf{\normalsize L@50}
    & \normalsize 5.47
    & \normalsize 8.00 & \normalsize 7.65 & \underline{\normalsize 8.48} & \textbf{\normalsize 15.66} \\
    \midrule

    \multirow{2}{*}{\textbf{\normalsize Street}}
    & \multirow{2}{*}{\textbf{\normalsize Map}}
    & \textbf{\normalsize R@1}
    & \normalsize 1.35
    & \normalsize 2.04 & \normalsize 1.74 & \underline{\normalsize 2.29} & \textbf{\normalsize 6.35} \\
    &
    & \textbf{\normalsize L@50}
    & \normalsize 6.83  
    & \underline{\normalsize 8.74} & \normalsize 7.36 & \normalsize 8.56 & \textbf{\normalsize 9.27} \\

    \midrule

    \multirow{2}{*}{\textbf{\normalsize Satellite}}
    & \multirow{2}{*}{\textbf{\normalsize Map}}
    & \textbf{\normalsize R@1}
    & \normalsize 1.45 
    & \normalsize 2.01 & \normalsize 1.96 & \underline{\normalsize 2.14} & \textbf{\normalsize 6.42} \\
    &
    & \textbf{\normalsize L@50}
    & \normalsize 6.98 
    & \normalsize 8.14 & \normalsize 7.61 & \underline{\normalsize 8.84} & \textbf{\normalsize 9.70} \\

    \midrule

    \multirow{2}{*}{\textbf{\normalsize Drone}}
    & \multirow{2}{*}{\textbf{\normalsize Map}}
    &  \textbf{\normalsize R@1}
    & \normalsize 1.51
    & \normalsize 2.03 & \normalsize 1.72 & \underline{\normalsize 2.15
    }& \textbf{\normalsize 6.72} \\
    &
    & \textbf{\normalsize L@50}
    & \normalsize 6.53  
    & \underline{\normalsize 8.33} & \normalsize 7.66 & \normalsize 7.84& \textbf{\normalsize 9.12} \\
    \bottomrule
  \end{tabular}%
  }
  \vspace{-5pt}
\end{table}

\subsubsection{Street-Map Retrieval}

For street-to-map geo-localization, we evaluate GeoBridge++ on the CVGlobal~\cite{ye2024cross} dataset, with the results reported in Table~\ref{CVG:street_map}. Since CVGlobal only provides a street-to-map protocol for map-related evaluation and GeoBridge does not incorporate static-map observations, we report GeoBridge++ for comparison with existing methods.

GeoBridge++ consistently achieves the best performance across both Regional val and Regional test, with particularly clear advantages on the more challenging Regional test split. This indicates that the learned representations generalize well across geographically distinct regions rather than relying on region-specific visual patterns. More importantly, the consistent improvements across different recall levels demonstrate that GeoBridge++ effectively bridges the substantial representation gap between visual street-view imagery and symbolic maps, leading to more robust and transferable street-to-map geo-localization.

\subsubsection{Unified Cross-View Geo-Localization}

Beyond pairwise retrieval, we further evaluate GeoBridge and GeoBridge++ under a unified cross-view setting, where multiple view combinations are handled within a shared representation space. We first evaluate unified cross-view geo-localization on GeoLoc, where drone, satellite, and street-view observations are geographically aligned. As shown in Table~\ref{tab:GeoLoc}, GeoBridge already provides strong performance across different retrieval directions, demonstrating the effectiveness of learning a shared representation beyond conventional satellite-centric matching. GeoBridge++, trained with the map-augmented GeoLoc dataset while evaluated on the same GeoLoc D/P/S retrieval setting, further improves all evaluated directions by a clear margin. The gains are consistently observed for both aerial-to-aerial retrieval and the more challenging aerial-to-ground retrieval, indicating that fact-guided geo-semantic alignment improves the consistency and discriminability of heterogeneous view representations.

We further conduct a more comprehensive evaluation on GeoLoc-MM by introducing static maps and six spatial scales ranging from 80 to 200~\(m\). Table~\ref{tab:GeoLoc-MM-drone} evaluates methods originally designed for drone-to-satellite localization by extending them to other view combinations, while Table~\ref{tab:GeoLoc-MM-street} provides the analogous evaluation for methods developed for street-to-satellite localization. GeoBridge++ consistently outperforms the compared methods across all retrieval directions and spatial scales. In particular, the improvement is maintained not only for conventional drone--satellite and street--satellite retrieval, but also for retrieval involving static maps. This demonstrates that static maps can be treated as a peer geographic observation rather than merely an auxiliary reference, enabling localization through alternative view combinations when conventional visual observations are unavailable or less reliable. Across the six spatial scales, GeoBridge++ maintains consistently strong performance without exhibiting a systematic degradation as the geographic extent changes. This behavior is observed for both visual-view pairs and map-related pairs, despite substantial variations in visible content, spatial coverage, and representation form. The results therefore indicate that the learned embedding space is not tied to a particular observation scale or view pair. Instead, GeoBridge++ provides a unified representation that supports arbitrary retrieval among drone, satellite, street-view, and static-map observations while remaining robust to scale variations. 

\begin{table}[t]
  \centering
  \caption{Ablation \(R@1\) results of GeoBridge with different alignment strategies on the GeoLoc dataset. D, T, and S denote Drone, Street-View Panorama, and Satellite, respectively. The best results are shown in bold, the second-best results are underlined.}
  \label{tab:ablation}
  \resizebox{\columnwidth}{!}{
  \begin{tabular}{ccccccc}
    \toprule
    \textbf{Method}& \textbf{D2S} & \textbf{S2D} & \textbf{T2S} & \textbf{S2T} & \textbf{D2T} & \textbf{T2D} \\
    \midrule
    Image-only   & 38.20 &  34.63& 6.43 &6.95&7.16 &4.90 \\
    Text-only   & \underline{42.83}& \underline{42.83} &  \underline{35.40}&\underline{36.40}&\underline{39.00}&\underline{38.63} \\
    \midrule
    GeoBridge& \textbf{45.06}& \textbf{44.81}  &  \textbf{38.87}  & \textbf{39.21}& \textbf{41.23} & \textbf{41.15} \\
    \bottomrule
  \end{tabular}
    }
    \vspace{-5pt}
\end{table}

Qualitative retrieval results are shown in Figures.~\ref{fig:triple} and~\ref{fig:quad}. To avoid redundant visualization of reciprocal retrieval directions, we select representative closed-loop retrieval settings such that each view serves as both query and reference. On GeoLoc, GeoBridge retrieves the correct locations despite large differences in viewpoint and appearance. The successful matches are largely supported by consistent scene layout and spatial structure, rather than local appearance alone. On GeoLoc-MM, GeoBridge++ is further evaluated with static maps and multiple spatial extents. The results remain stable as the observation scale changes from 80 to 200~\(m\). More importantly, the map-related cases show that symbolic layouts can be matched with corresponding structures in real-world imagery, indicating that the learned representation preserves geographic consistency across both visual and symbolic observations. Additional visualizations are provided in the supplementary material.

\subsection{Cross-Modal Geo-localization Performance}

To further assess the effectiveness of GeoBridge for cross-modal retrieval, we evaluate it on the publicly available remote sensing annotation dataset RSIEval~\cite{HU2025272}, and results are shown in Table~\ref{tab:RSIEval}. This dataset contains 100 image-text pairs. Under identical evaluation settings, GeoBridge consistently outperforms the compared methods, while GeoBridge++ further achieves the best overall performance. 

\begin{table*}[t]
\centering
\caption{Ablation \(R@1\) results of GeoBridge++ with different strategies on the map-augmented GeoLoc dataset. Fact, Bridge, Pairwise denote location fact guidance, geo-semantic bridge alignment, and direct pairwise image alignment, respectively. D, T, S, and M denote Drone, Street-View Panorama, Satellite, and Static Map, respectively. The best results are shown in bold, the second-best results are underlined.}
\label{factgeo_ablation}
\resizebox{\textwidth}{!}{
\begin{tabular}{ccc|cccccccccccc}
\toprule
\textbf{Fact} & \textbf{Bridge} & \textbf{Pairwise} &
\textbf{D2S} &
\textbf{S2D} &
\textbf{T2S} &
\textbf{S2T} &
\textbf{D2T} &
\textbf{T2D} &
\textbf{D2M} &
\textbf{M2D} &
\textbf{S2M} &
\textbf{M2S} &
\textbf{T2M} &
\textbf{M2T} \\
\midrule
  \checkmark&  & &\underline{52.19} &   \underline{53.46} &  43.99 & \underline{41.31}& \underline{44.51}&\underline{44.96}& \underline{45.63} &\underline{43.00} &\underline{43.43}&\underline{37.71}&\underline{27.92}&\underline{28.09}\\
  &\checkmark  & & 25.83&  25.83 &  7.14  & 6.52& 6.69& 7.59& 13.87& 14.11&17.44&16.71&4.34&3.66 \\
  & &\checkmark   &21.94  & 20.31 &  1.18 &  0.64& 0.58& 0.64& 16.58& 16.91&14.46&16.03&0.60&15.61 \\
  \checkmark& &  \checkmark &42.53  & 44.33 & \textbf{57.72}  & 25.32& 25.49&  29.88&29.13&27.99&35.54&30.93&24.29&18.78 \\
  \midrule
  \textbf{\checkmark}  & \textbf{\checkmark}& &\textbf{58.01}& \textbf{57.88} & \underline{49.69}  & \textbf{45.84}&  \textbf{55.86}&  \textbf{59.24}& \textbf{49.09}& \textbf{44.22}&\textbf{46.10}&\textbf{45.45}&\textbf{39.58}&\textbf{36.72} \\
\bottomrule
\end{tabular}}
\end{table*}

\begin{table*}[t]
\centering
\caption{Ablation \(R@1\) results of GeoBridge++ with different types of location fact guidance on the map-augmented GeoLoc dataset. G, V, and SH denote geographic, visual, and shared geo-visual facts, respectively. D, T, S, and M denote Drone, Street-View Panorama, Satellite, and Static Map, respectively. The best results are shown in bold, the second-best results are underlined.}
\label{fact_ablation}
\resizebox{\textwidth}{!}{
\begin{tabular}{ccc|cccccccccccc}
\toprule
\textbf{G} & \textbf{V} & \textbf{SH} &
\textbf{D2S}&
\textbf{S2D} &
\textbf{T2S} &
\textbf{S2T} &
\textbf{D2T} &
\textbf{T2D} &
\textbf{D2M} &
\textbf{M2D} &
\textbf{S2M} &
\textbf{M2S} &
\textbf{T2M} &
\textbf{M2T} \\

\midrule
  \checkmark&  & &46.81 &   49.47 &  -- &  --&--& --& \underline{40.46} &38.91&35.86&33.90&--&--\\
  &\checkmark  & & \underline{52.19}&  51.15 &  \underline{39.49} & \underline{39.60}&  \underline{42.55}& \underline{42.18}&  --& --&--&--&--&-- \\
  & &\checkmark   &47.39 & 51.65 &  25.58  & 26.57&  27.92&  28.69&  37.08& \underline{39.92}&\underline{38.24}&\underline{36.95} &\underline{21.70}&\underline{22.93} \\
  \checkmark& \checkmark&   &51.99 & \underline{53.30}&  17.72 &  29.47&  21.53&  19.59& 31.17& 25.21&26.69&22.31&8.56&7.92 \\
  \midrule
  \checkmark  & \checkmark& \checkmark&\textbf{58.01}& \textbf{57.88}&  \textbf{49.69} & \textbf{45.84}&  \textbf{55.86}&  \textbf{59.24}&  \textbf{49.09}& \textbf{44.22} &\textbf{46.10} &\textbf{45.45} &\textbf{39.58} &\textbf{36.72} \\
\bottomrule
\end{tabular}}
\vspace{-5pt}
\end{table*}

We further evaluate cross-modal geo-localization on GeoLoc, where descriptions generated from a single source view are used to retrieve images from the remaining views. As shown in Table~\ref{tab:streetlang-image}, GeoBridge consistently outperforms the compared vision-language methods across most retrieval settings, demonstrating effective image--text alignment. GeoBridge++ further improves upon GeoBridge across all evaluated directions, indicating that fact-guided supervision provides more discriminative location-level semantics. To examine retrieval involving static maps, we augment GeoLoc with aligned map observations while retaining the original locations and visual observations. This setting avoids the additional multi-scale variations introduced in GeoLoc-MM and allows the effect of the map modality to be evaluated under otherwise consistent conditions. The results are reported in Table~\ref{tab:mag-image}, which extends the evaluation with map-related retrieval directions, including retrieval from map descriptions to images from visual views and from descriptions of visual views to map images. GeoBridge++ achieves the best performance across all settings, demonstrating its ability to associate symbolic geographic layouts with their corresponding visual observations.

Figure~\ref{fig:language} and Figure~\ref{fig:map} present qualitative results for cross-modal geo-localization using descriptions generated from single-view observations. In Figure~\ref{fig:language}, we select street-view-to-drone retrieval as a representative setting because of the substantial viewpoint gap between ground-level and aerial observations. GeoBridge effectively transfers location-relevant semantics from street-view descriptions to the drone perspective, enabling correct retrieval despite large viewpoint differences. Figure~\ref{fig:map} further evaluates GeoBridge++ using descriptions derived from map observations to retrieve street-view, satellite, and drone images, providing a more challenging setting that bridges symbolic geographic representations and visual observations. In this setting, geographic cues such as intersections, water boundaries, and surrounding structures provide effective correspondence across symbolic and visual representations. The correct matches are consistently ranked among the top retrieved candidates in both cases, illustrating that the learned representations preserve location-relevant semantics across heterogeneous views and modalities. Additional visualizations are provided in the supplementary material.

\subsection{Ablation Experiment}

For controlled component-level analysis, GeoBridge is pre-trained and evaluated on GeoLoc, whereas GeoBridge++ is pre-trained and evaluated on the map-augmented GeoLoc setting. This avoids the additional multi-scale expansion introduced in GeoLoc-MM and enables a focused evaluation of the proposed components.

\textbf{Semantic Anchor Alignment.} To assess the effectiveness of semantic anchors, we conduct an ablation study on alignment strategies. The \(R@1\) results are reported in Table~\ref{tab:ablation}. Image-only denotes image-image alignment, Text-only denotes text–image alignment, and GeoBridge fuses both while unifying multi-view representations in a shared semantic space via semantic anchors. Across all retrieval tasks, the fusion strategy attains the highest \(R@1\) results. With image-only alignment, robust correspondences between street panoramas and markedly different viewpoints (\eg, satellite and drone) are difficult to establish, yielding the weakest overall performance. In contrast, text-only alignment benefits from descriptions that capture key semantic cues across all three views, and thus ranks second overall. These results demonstrate that using location representations from multi-views as semantic anchors substantially narrows the semantic gap between ground and overhead views, improving the accuracy of cross-view retrieval. The AP results are reported in the supplementary material.

\textbf{Fact Guidance and Geo-Semantic Bridging.} We investigate different alignment strategies in Table~\ref{factgeo_ablation}. Fact, Bridge, and Pairwise denote location fact guidance, geo-semantic bridge alignment, and direct pairwise image alignment, respectively. Pairwise-only yields consistently poor performance, indicating that directly learning image--image correspondences is insufficient for aligning highly heterogeneous views. In contrast, Fact-only substantially improves all retrieval directions and clearly outperforms both Bridge-only and Pairwise-only, demonstrating the effectiveness of factual supervision in providing stable location-level guidance. Bridge-only also performs better than Pairwise-only, showing the benefit of coordinating multiple views through a shared representation. Combining Fact with Pairwise markedly improves over Pairwise-only, further confirming the value of fact guidance, but remains generally inferior to Fact-only except on T2S, suggesting possible conflicts introduced by direct pairwise constraints. In contrast, combining Fact with the geo-semantic bridge achieves the best performance on most directions, demonstrating that the two are complementary in providing stable supervision and effective cross-view coordination. The AP results are further reported in the supplementary material.

\textbf{Location Fact Types.} We further analyze the roles of geographic, visual, and shared geo-visual facts in Table~\ref{fact_ablation}. Geographic facts (\textit{G}) and visual facts (\textit{V}) provide modality-aware supervision over different view subsets, whereas shared geo-visual facts (\textit{SH}) supervise all views through a common semantic reference. Individually, \textit{G} and \textit{V} improve the views within their respective observable scopes, while \textit{SH} alone remains less discriminative than the full model. Notably, combining \textit{G} and \textit{V} without \textit{SH} degrades several retrieval directions, suggesting that separate modality-aware constraints are insufficient to maintain a globally consistent embedding space. Introducing \textit{SH} connects these view-specific supervision groups through a shared location-level reference, while \textit{G} and \textit{V} preserve complementary geographic and visual cues. Jointly using all three fact types achieves the best performance across all retrieval directions, demonstrating the importance of balancing modality-specific supervision with cross-view commonality. The AP results are further reported in the supplementary material.

\section{Conclusion}
In this work, we investigated unified cross-view geo-localization beyond the conventional satellite-centric and pair-specific paradigm. GeoBridge uses semantic anchors to align drone, satellite, and street-view observations in a shared space for cross-view image retrieval and language-to-image localization. GeoBridge++ further introduces location fact guidance and a geo-semantic bridge to reduce appearance ambiguity and extend unified localization to static maps. We also constructed GeoLoc and further extended it to GeoLoc-MM, providing geographically aligned multi-view observations with broader modality and scale coverage. Experiments on our datasets and public benchmarks show consistent improvements across retrieval directions, regions, scales, and cross-modal settings, while ablations verify the complementary roles of fact guidance and geo-semantic bridging. These results demonstrate the potential of moving cross-view geo-localization from pair-specific visual matching toward unified geographic representation learning across heterogeneous observations.
 
\bibliographystyle{IEEEtran}
\bibliography{paper}

@article{durgam2024cross,
  title={Cross-view geo-localization: a survey},
  author={Durgam, Abhilash and Paheding, Sidike and Dhiman, Vikas and Devabhaktuni, Vijay},
  journal={IEEE Access},
  year={2024},
  publisher={IEEE}
}

@article{wang2026geoeyes,
  title={GeoEyes: On-Demand Visual Focusing for Evidence-Grounded Understanding of Ultra-High-Resolution Remote Sensing Imagery},
  author={Wang, Fengxiang and Chen, Mingshuo and Li, Yueying and Yang, Yajie and Zhang, Yifan and Lan, Long and Yang, Xue and Sun, Hongda and Wang, Yulin and Wang, Di and others},
  journal={arXiv preprint arXiv:2602.14201},
  year={2026}
}

@article{fan2021design,
  title={Design and implementation of intelligent EOD system based on six-rotor UAV},
  author={Fan, Jiwei and Lu, Ruitao and Yang, Xiaogang and Gao, Fan and Li, Qingge and Zeng, Jun},
  journal={Drones},
  volume={5},
  number={4},
  pages={146},
  year={2021},
  publisher={MDPI}
}

@ARTICLE{hypersigma,
  author={Wang, Di and Hu, Meiqi and Jin, Yao and Miao, Yuchun and Yang, Jiaqi and Xu, Yichu and Qin, Xiaolei and Ma, Jiaqi and Sun, Lingyu and Li, Chenxing and Fu, Chuan and Chen, Hongruixuan and Han, Chengxi and Yokoya, Naoto and Zhang, Jing and Xu, Minqiang and Liu, Lin and Zhang, Lefei and Wu, Chen and Du, Bo and Tao, Dacheng and Zhang, Liangpei},
  journal={IEEE Transactions on Pattern Analysis and Machine Intelligence}, 
  title={HyperSIGMA: Hyperspectral Intelligence Comprehension Foundation Model}, 
  year={2025},
  volume={47},
  number={8},
  pages={6427-6444}
}

@inproceedings{sarlin2023orienternet,
  title={Orienternet: Visual localization in 2d public maps with neural matching},
  author={Sarlin, Paul-Edouard and DeTone, Daniel and Yang, Tsun-Yi and Avetisyan, Armen and Straub, Julian and Malisiewicz, Tomasz and Bulo, Samuel Rota and Newcombe, Richard and Kontschieder, Peter and Balntas, Vasileios},
  booktitle={Proceedings of the IEEE/CVF Conference on Computer Vision and Pattern Recognition},
  pages={21632--21642},
  year={2023}
}

@inproceedings{torii201524,
  title={24/7 place recognition by view synthesis},
  author={Torii, Akihiko and Arandjelovic, Relja and Sivic, Josef and Okutomi, Masatoshi and Pajdla, Tomas},
  booktitle={Proceedings of the IEEE conference on computer vision and pattern recognition},
  pages={1808--1817},
  year={2015}
}

@article{TiMo,
      title={TiMo: Spatiotemporal Foundation Model for Satellite Image Time Series}, 
      author={Xiaolei Qin and Di Wang and Jing Zhang and Fengxiang Wang and Xin Su and Bo Du and Liangpei Zhang},
      journal={arXiv preprint arXiv:2505.08723},
      year={2025}
}

@inproceedings{fervers2023uncertainty,
  title={Uncertainty-aware vision-based metric cross-view geolocalization},
  author={Fervers, Florian and Bullinger, Sebastian and Bodensteiner, Christoph and Arens, Michael and Stiefelhagen, Rainer},
  booktitle={Proceedings of the IEEE/CVF Conference on Computer Vision and Pattern Recognition},
  pages={21621--21631},
  year={2023}
}

@article{chen2023gnss,
  title={GNSS high-precision augmentation for autonomous vehicles: Requirements, solution, and technical challenges},
  author={Chen, Liang and Zheng, Fu and Gong, Xiaopeng and Jiang, Xinyuan},
  journal={Remote Sensing},
  volume={15},
  number={6},
  pages={1623},
  year={2023},
  publisher={MDPI}
}

@article{wu2025uav,
  title={UAV-GeoLoc: A Large-Vocabulary Dataset and Geometry-Transformed Method for UAV Geo-Localization},
  author={Wu, Rouwan and Deng, Jiacheng and Mou, Mingyu and He, Xingyi and Zhang, Maojun and Liu, Yu and Yan, Shen},
  journal={IEEE Robotics and Automation Letters},
  year={2025},
  publisher={IEEE}
}

@article{li2025cross,
  title={Cross-view geolocalization and disaster mapping with street-view and VHR satellite imagery: A case study of Hurricane IAN},
  author={Li, Hao and Deuser, Fabian and Yin, Wenping and Luo, Xuanshu and Walther, Paul and Mai, Gengchen and Huang, Wei and Werner, Martin},
  journal={ISPRS Journal of Photogrammetry and Remote Sensing},
  volume={220},
  pages={841--854},
  year={2025},
  publisher={Elsevier}
}

@article{wang2021each,
  title={Each part matters: Local patterns facilitate cross-view geo-localization},
  author={Wang, Tingyu and Zheng, Zhedong and Yan, Chenggang and Zhang, Jiyong and Sun, Yaoqi and Zheng, Bolun and Yang, Yi},
  journal={IEEE Transactions on Circuits and Systems for Video Technology},
  volume={32},
  number={2},
  pages={867--879},
  year={2021},
  publisher={IEEE}
}

@inproceedings{zheng2020university,
  title={University-1652: A multi-view multi-source benchmark for drone-based geo-localization},
  author={Zheng, Zhedong and Wei, Yunchao and Yang, Yi},
  booktitle={Proceedings of the 28th ACM international conference on Multimedia},
  pages={1395--1403},
  year={2020}
}

@inproceedings{song2026geobridge,
  title={Geobridge: A semantic-anchored multi-view foundation model bridging images and text for geo-localization},
  author={Song, Zixuan and Zhang, Jing and Wang, Di and Zhou, Zidie and Liu, Wenbin and Guo, Haonan and Wang, En and Du, Bo},
  booktitle={Proceedings of the IEEE/CVF Conference on Computer Vision and Pattern Recognition},
  pages={27793--27803},
  year={2026}
}

@inproceedings{ye2024cross,
  title={Cross-view image geo-localization with Panorama-BEV Co-Retrieval Network},
  author={Ye, Junyan and Lv, Zhutao and Li, Weijia and Yu, Jinhua and Yang, Haote and Zhong, Huaping and He, Conghui},
  booktitle={European Conference on Computer Vision},
  pages={74--90},
  year={2024},
  organization={Springer}
}

@article{dai2021transformer,
  title={A transformer-based feature segmentation and region alignment method for UAV-view geo-localization},
  author={Dai, Ming and Hu, Jianhong and Zhuang, Jiedong and Zheng, Enhui},
  journal={IEEE Transactions on Circuits and Systems for Video Technology},
  volume={32},
  number={7},
  pages={4376--4389},
  year={2021},
  publisher={IEEE}
}

@article{li2024geoformer,
  title={GeoFormer: An effective transformer-based Siamese network for UAV geolocalization},
  author={Li, Qingge and Yang, Xiaogang and Fan, Jiwei and Lu, Ruitao and Tang, Bin and Wang, Siyu and Su, Shuang},
  journal={IEEE Journal of Selected Topics in Applied Earth Observations and Remote Sensing},
  volume={17},
  pages={9470--9491},
  year={2024},
  publisher={IEEE}
}

@article{zhu2023sues,
  title={SUES-200: A multi-height multi-scene cross-view image benchmark across drone and satellite},
  author={Zhu, Runzhe and Yin, Ling and Yang, Mingze and Wu, Fei and Yang, Yuncheng and Hu, Wenbo},
  journal={IEEE Transactions on Circuits and Systems for Video Technology},
  volume={33},
  number={9},
  pages={4825--4839},
  year={2023},
  publisher={IEEE}
}

@article{dai2023vision,
  title={Vision-based UAV self-positioning in low-altitude urban environments},
  author={Dai, Ming and Zheng, Enhui and Feng, Zhenhua and Qi, Lei and Zhuang, Jiedong and Yang, Wankou},
  journal={IEEE Transactions on Image Processing},
  volume={33},
  pages={493--508},
  year={2023},
  publisher={IEEE}
}

@article{lv2024direction,
  title={Direction-guided multiscale feature fusion network for geo-localization},
  author={Lv, Hongxiang and Zhu, Hai and Zhu, Runzhe and Wu, Fei and Wang, Chunyuan and Cai, Meiyu and Zhang, Kaiyu},
  journal={IEEE Transactions on Geoscience and Remote Sensing},
  volume={62},
  pages={1--13},
  year={2024},
  publisher={IEEE}
}

@article{mai2022review,
  title={A review of location encoding for GeoAI: methods and applications},
  author={Mai, Gengchen and Janowicz, Krzysztof and Hu, Yingjie and Gao, Song and Yan, Bo and Zhu, Rui and Cai, Ling and Lao, Ni},
  journal={International Journal of Geographical Information Science},
  volume={36},
  number={4},
  pages={639--673},
  year={2022},
  publisher={Taylor \& Francis}
}

@inproceedings{vo2016localizing,
  title={Localizing and orienting street views using overhead imagery},
  author={Vo, Nam N and Hays, James},
  booktitle={European conference on computer vision},
  pages={494--509},
  year={2016},
  organization={Springer}
}

@inproceedings{regmi2019bridging,
  title={Bridging the domain gap for ground-to-aerial image matching},
  author={Regmi, Krishna and Shah, Mubarak},
  booktitle={Proceedings of the IEEE/CVF International Conference on Computer Vision},
  pages={470--479},
  year={2019}
}

@inproceedings{toker2021coming,
  title={Coming down to earth: Satellite-to-street view synthesis for geo-localization},
  author={Toker, Aysim and Zhou, Qunjie and Maximov, Maxim and Leal-Taix{\'e}, Laura},
  booktitle={Proceedings of the IEEE/CVF Conference on Computer Vision and Pattern Recognition},
  pages={6488--6497},
  year={2021}
}

@inproceedings{zhou2021efficient,
  title={Efficient localisation using images and OpenStreetMaps},
  author={Zhou, Mengjie and Chen, Xieyuanli and Samano, Noe and Stachniss, Cyrill and Calway, Andrew},
  booktitle={2021 IEEE/RSJ International Conference on Intelligent Robots and Systems (IROS)},
  pages={5507--5513},
  year={2021},
  organization={IEEE}
}

@inproceedings{jia2021scaling,
  title={Scaling up visual and vision-language representation learning with noisy text supervision},
  author={Jia, Chao and Yang, Yinfei and Xia, Ye and Chen, Yi-Ting and Parekh, Zarana and Pham, Hieu and Le, Quoc and Sung, Yun-Hsuan and Li, Zhen and Duerig, Tom},
  booktitle={International conference on machine learning},
  pages={4904--4916},
  year={2021},
  organization={PMLR}
}

@article{wu2026vision,
  title={Vision-Language Reasoning for Geolocalization: A Reinforcement Learning Approach},
  author={Wu, Biao and Fang, Meng and Chen, Ling and Xu, Ke and Cheng, Tao and Wang, Jun},
  journal={arXiv preprint arXiv:2601.00388},
  year={2026}
}

@inproceedings{ye2025cross,
  title={Where am i? cross-view geo-localization with natural language descriptions},
  author={Ye, Junyan and Lin, Honglin and Ou, Leyan and Chen, Dairong and Wang, Zihao and Zhu, Qi and He, Conghui and Li, Weijia},
  booktitle={Proceedings of the IEEE/CVF International Conference on Computer Vision},
  pages={5890--5900},
  year={2025}
}

@inproceedings{CVUSA,
  title={Wide-area image geolocalization with aerial reference imagery},
  author={Workman, Scott and Souvenir, Richard and Jacobs, Nathan},
  booktitle={Proceedings of the IEEE International Conference on Computer Vision},
  pages={3961--3969},
  year={2015}
}

@inproceedings{CVACT,
  title={Lending orientation to neural networks for cross-view geo-localization},
  author={Liu, Liu and Li, Hongdong},
  booktitle={Proceedings of the IEEE/CVF conference on computer vision and pattern recognition},
  pages={5624--5633},
  year={2019}
}

@inproceedings{zhu2021vigor,
  title={Vigor: Cross-view image geo-localization beyond one-to-one retrieval},
  author={Zhu, Sijie and Yang, Taojiannan and Chen, Chen},
  booktitle={Proceedings of the IEEE/CVF Conference on Computer Vision and Pattern Recognition},
  pages={3640--3649},
  year={2021}
}

@inproceedings{tian2017cross,
  title={Cross-view image matching for geo-localization in urban environments},
  author={Tian, Yicong and Chen, Chen and Shah, Mubarak},
  booktitle={Proceedings of the IEEE Conference on Computer Vision and Pattern Recognition},
  pages={3608--3616},
  year={2017}
}

@article{xu2024uav,
  title={Uav-visloc: A large-scale dataset for uav visual localization},
  author={Xu, Wenjia and Yao, Yaxuan and Cao, Jiaqi and Wei, Zhiwei and Liu, Chunbo and Wang, Jiuniu and Peng, Mugen},
  journal={arXiv preprint arXiv:2405.11936},
  year={2024}
}

@inproceedings{xia2024adapting,
  title={Adapting fine-grained cross-view localization to areas without fine ground truth},
  author={Xia, Zimin and Shi, Yujiao and Li, Hongdong and FP Kooij, Julian},
  booktitle={European Conference on Computer Vision},
  pages={397--415},
  year={2024},
  organization={Springer}
}

@inproceedings{roller2012supervised,
  title={Supervised text-based geolocation using language models on an adaptive grid},
  author={Roller, Stephen and Speriosu, Michael and Rallapalli, Sarat and Wing, Benjamin and Baldridge, Jason},
  booktitle={Proceedings of the 2012 joint conference on empirical methods in natural language processing and computational natural language learning},
  pages={1500--1510},
  year={2012}
}

@inproceedings{li2024georeasoner,
  title={Georeasoner: Geo-localization with reasoning in street views using a large vision-language model},
  author={Li, Ling and Ye, Yu and Jiang, Bingchuan and Zeng, Wei},
  booktitle={Forty-first International Conference on Machine Learning},
  year={2024}
}

@article{wang2026text,
  title={Text Before Vision: Staged Knowledge Injection Matters for Agentic RLVR in Ultra-High-Resolution Remote Sensing Understanding},
  author={Wang, Fengxiang and Chen, Mingshuo and Li, Yueying and Yang, Yajie and Zhou, Yuhao and Wang, Di and Zhang, Yifan and Wang, Haoyu and Zhao, Haiyan and Sun, Hongda and others},
  journal={arXiv preprint arXiv:2602.14225},
  year={2026}
}

@article{wang2025geozero,
  title   = {GeoZero: Incentivizing Reasoning from Scratch on Geospatial Scenes},
  author  = {Wang, Di and Liu, Shunyu and Jiang, Wentao and Wang, Fengxiang and Liu, Yi and Qin, Xiaolei and Luo, Zhiming and Zhou, Chaoyang and Guo, Haonan and Zhang, Jing and Du, Bo and Tao, Dacheng and Zhang, Liangpei},
  journal = {arXiv preprint arXiv:2511.22645},
  year    = {2025}
}

@inproceedings{chu2024towards,
  title={Towards natural language-guided drones: GeoText-1652 benchmark with spatial relation matching},
  author={Chu, Meng and Zheng, Zhedong and Ji, Wei and Wang, Tingyu and Chua, Tat-Seng},
  booktitle={European Conference on Computer Vision},
  pages={213--231},
  year={2024},
  organization={Springer}
}

@article{yuan2021florence,
  title={Florence: A new foundation model for computer vision. arXiv 2021},
  author={Yuan, Lu and Chen, Dongdong and Chen, Yi-Ling and Codella, Noel and Dai, Xiyang and Gao, Jianfeng and Hu, Houdong and Huang, Xuedong and Li, Boxin and Li, Chunyuan and others},
  journal={arXiv preprint arXiv:2111.11432},
  year={2021}
}

@inproceedings{li2022blip,
  title={Blip: Bootstrapping language-image pre-training for unified vision-language understanding and generation},
  author={Li, Junnan and Li, Dongxu and Xiong, Caiming and Hoi, Steven},
  booktitle={International conference on machine learning},
  pages={12888--12900},
  year={2022},
  organization={PMLR}
}

@inproceedings{li2023blip,
  title={Blip-2: Bootstrapping language-image pre-training with frozen image encoders and large language models},
  author={Li, Junnan and Li, Dongxu and Savarese, Silvio and Hoi, Steven},
  booktitle={International conference on machine learning},
  pages={19730--19742},
  year={2023},
  organization={PMLR}
}

@article{liu2026radar,
  title={Seeing Clearly without Training: Mitigating Hallucinations in Multimodal LLMs for Remote Sensing},
  author={Liu, Yi and Zhang, Jing and Wang, Di and Tian, Xiaoyu and Guo, Haonan and Du, Bo},
  journal={arXiv preprint arXiv:2603.02754},
  year={2026},
  doi={10.48550/arXiv.2603.02754}}

@article{hurst2024gpt,
  title={Gpt-4o system card},
  author={Hurst, Aaron and Lerer, Adam and Goucher, Adam P and Perelman, Adam and Ramesh, Aditya and Clark, Aidan and Ostrow, AJ and Welihinda, Akila and Hayes, Alan and Radford, Alec and others},
  journal={arXiv preprint arXiv:2410.21276},
  year={2024}
}

@inproceedings{radford2021learning,
  title={Learning transferable visual models from natural language supervision},
  author={Radford, Alec and Kim, Jong Wook and Hallacy, Chris and Ramesh, Aditya and Goh, Gabriel and Agarwal, Sandhini and Sastry, Girish and Askell, Amanda and Mishkin, Pamela and Clark, Jack and others},
  booktitle={International conference on machine learning},
  pages={8748--8763},
  year={2021},
  organization={PmLR}
}

@article{HU2025272,
title = {RSGPT: A remote sensing vision language model and benchmark},
journal = {ISPRS Journal of Photogrammetry and Remote Sensing},
volume = {224},
pages = {272-286},
year = {2025},
issn = {0924-2716},
doi = {https://doi.org/10.1016/j.isprsjprs.2025.03.028},
author = {Yuan Hu and Jianlong Yuan and Congcong Wen and Xiaonan Lu and Yu Liu and Xiang Li},
}

@inproceedings{kim2021vilt,
  title={Vilt: Vision-and-language transformer without convolution or region supervision},
  author={Kim, Wonjae and Son, Bokyung and Kim, Ildoo},
  booktitle={International conference on machine learning},
  pages={5583--5594},
  year={2021},
  organization={PMLR}
}

@article{fang2024eva,
  title={Eva-02: A visual representation for neon genesis},
  author={Fang, Yuxin and Sun, Quan and Wang, Xinggang and Huang, Tiejun and Wang, Xinlong and Cao, Yue},
  journal={Image and Vision Computing},
  volume={149},
  pages={105171},
  year={2024},
  publisher={Elsevier}
}

@article{SAIG-D,
  title={Simple, effective and general: A new backbone for cross-view image geo-localization},
  author={Zhu, Yingying and Yang, Hongji and Lu, Yuxin and Huang, Qiang},
  journal={arXiv preprint arXiv:2302.01572},
  year={2023}
}

@article{DWDR,
  title={Learning cross-view geo-localization embeddings via dynamic weighted decorrelation regularization},
  author={Wang, Tingyu and Zheng, Zhedong and Zhu, Zunjie and Sun, Yaoqi and Yan, Chenggang and Yang, Yi},
  journal={IEEE Transactions on Geoscience and Remote Sensing},
  year={2024},
  publisher={IEEE}
}

@article{MBF,
  title={UAV’s status is worth considering: A fusion representations matching method for geo-localization},
  author={Zhu, Runzhe and Yang, Mingze and Yin, Ling and Wu, Fei and Yang, Yuncheng},
  journal={Sensors},
  volume={23},
  number={2},
  pages={720},
  year={2023},
  publisher={MDPI}
}

@ARTICLE{MCCG,
  author={Shen, Tianrui and Wei, Yingmei and Kang, Lai and Wan, Shanshan and Yang, Yee-Hong},
  journal={IEEE Transactions on Circuits and Systems for Video Technology}, 
  title={MCCG: A ConvNeXt-Based Multiple-Classifier Method for Cross-View Geo-Localization}, 
  year={2024},
  volume={34},
  number={3},
  pages={1456-1468},
  doi={10.1109/TCSVT.2023.3296074}}

@article{SeGCN,
  title={Segcn: A semantic-aware graph convolutional network for uav geo-localization},
  author={Liu, Xiangzeng and Wang, Ziyao and Wu, Yue and Miao, Qiguang},
  journal={IEEE Journal of Selected Topics in Applied Earth Observations and Remote Sensing},
  volume={17},
  pages={6055--6066},
  year={2024},
  publisher={IEEE}
}

@ARTICLE{CCR,
  author={Du, Haolin and He, Jingfei and Zhao, Yuanqing},
  journal={IEEE Transactions on Circuits and Systems for Video Technology}, 
  title={CCR: A Counterfactual Causal Reasoning-Based Method for Cross-View Geo-Localization}, 
  year={2024},
  volume={34},
  number={11},
  pages={11630-11643},
  doi={10.1109/TCSVT.2024.3425509}}

@ARTICLE{MEAN,
  author={Chen, Zhongwei and Yang, Zhao-Xu and Rong, Hai-Jun},
  journal={IEEE Transactions on Geoscience and Remote Sensing}, 
  title={Multilevel Embedding and Alignment Network With Consistency and Invariance Learning for Cross-View Geo-Localization}, 
  year={2025},
  volume={63},
  number={},
  pages={1-15},
  doi={10.1109/TGRS.2025.3572775}}

@ARTICLE{DAC,
  author={Xia, Panwang and Wan, Yi and Zheng, Zhi and Zhang, Yongjun and Deng, Jiwei},
  journal={IEEE Transactions on Circuits and Systems for Video Technology}, 
  title={Enhancing Cross-View Geo-Localization With Domain Alignment and Scene Consistency}, 
  year={2024},
  volume={34},
  number={12},
  pages={13271-13281},
  doi={10.1109/TCSVT.2024.3443510}}

@inproceedings{Sample4geo,
  title={Sample4geo: Hard negative sampling for cross-view geo-localisation},
  author={Deuser, Fabian and Habel, Konrad and Oswald, Norbert},
  booktitle={Proceedings of the IEEE/CVF International Conference on Computer Vision},
  pages={16847--16856},
  year={2023}
}

@ARTICLE{CGSI,
  author={Sun, Jian and Huang, Junlang and Jiang, Xinyu and Zhou, Yimin and VONG, Chi-Man},
  journal={IEEE Transactions on Circuits and Systems for Video Technology}, 
  title={CGSI: Context-Guided and UAV's Status Informed Multimodal Framework for Generalizable Cross-View Geo-Localization}, 
  year={2025},
  volume={},
  number={},
  pages={1-1},
  doi={10.1109/TCSVT.2025.3604002}
  }

@inproceedings{Transgeo,
  title={Transgeo: Transformer is all you need for cross-view image geo-localization},
  author={Zhu, Sijie and Shah, Mubarak and Chen, Chen},
  booktitle={Proceedings of the IEEE/CVF Conference on Computer Vision and Pattern Recognition},
  pages={1162--1171},
  year={2022}
}

@inproceedings{FRGeo,
  title={Aligning geometric spatial layout in cross-view geo-localization via feature recombination},
  author={Zhang, Qingwang and Zhu, Yingying},
  booktitle={Proceedings of the AAAI Conference on Artificial Intelligence},
  volume={38},
  number={7},
  pages={7251--7259},
  year={2024}
}

@inproceedings{VimGeo,
  title     = {VimGeo: Efficient Cross-View Geo-Localization with Vision Mamba Architecture},
  author    = {Huang, Jinglin and Wu, Maoqiang and Li, Peichun and Wu, Wen and Yu, Rong},
  booktitle = {Proceedings of the Thirty-Fourth International Joint Conference on
               Artificial Intelligence, {IJCAI-25}},
  publisher = {International Joint Conferences on Artificial Intelligence Organization},
  editor    = {James Kwok},
  pages     = {1188--1196},
  year      = {2025},
  month     = {8},
  note      = {Main Track},
  doi       = {10.24963/ijcai.2025/133},
}

@article{AuxGeo,
title = {Cross-view geo-localization with panoramic street-view and VHR satellite imagery in decentrality settings},
journal = {ISPRS Journal of Photogrammetry and Remote Sensing},
volume = {227},
pages = {1-11},
year = {2025},
issn = {0924-2716},
doi = {https://doi.org/10.1016/j.isprsjprs.2025.04.033},
author = {Panwang Xia and Lei Yu and Yi Wan and Qiong Wu and Peiqi Chen and Liheng Zhong and Yongxiang Yao and Dong Wei and Xinyi Liu and Lixiang Ru and Yingying Zhang and Jiangwei Lao and Jingdong Chen and Ming Yang and Yongjun Zhang},
}

@article{HC-Net,
  title={Fine-grained cross-view geo-localization using a correlation-aware homography estimator},
  author={Wang, Xiaolong and Xu, Runsen and Cui, Zhuofan and Wan, Zeyu and Zhang, Yu},
  journal={Advances in Neural Information Processing Systems},
  volume={36},
  pages={5301--5319},
  year={2023}
}

@article{CAMP,
  title={Camp: A cross-view geo-localization method using contrastive attributes mining and position-aware partitioning},
  author={Wu, Qiong and Wan, Yi and Zheng, Zhi and Zhang, Yongjun and Wang, Guangshuai and Zhao, Zhenyang},
  journal={IEEE Transactions on Geoscience and Remote Sensing},
  year={2024},
  publisher={IEEE}
}

@inproceedings{zhang2023cross,
  title={Cross-view geo-localization via learning disentangled geometric layout correspondence},
  author={Zhang, Xiaohan and Li, Xingyu and Sultani, Waqas and Zhou, Yi and Wshah, Safwan},
  booktitle={Proceedings of the AAAI conference on artificial intelligence},
  volume={37},
  number={3},
  pages={3480--3488},
  year={2023}
}

@article{qwen3,
  title={Qwen3 technical report},
  author={Yang, An and Li, Anfeng and Yang, Baosong and Zhang, Beichen and Hui, Binyuan and Zheng, Bo and Yu, Bowen and Gao, Chang and Huang, Chengen and Lv, Chenxu and others},
  journal={arXiv preprint arXiv:2505.09388},
  year={2025}
}

@article{team2023gemini,
  title={Gemini: a family of highly capable multimodal models},
  author={Team, Gemini and Anil, Rohan and Borgeaud, Sebastian and Alayrac, Jean-Baptiste and Yu, Jiahui and Soricut, Radu and Schalkwyk, Johan and Dai, Andrew M and Hauth, Anja and Millican, Katie and others},
  journal={arXiv preprint arXiv:2312.11805},
  year={2023}
}

@article{gebru2021datasheets,
  title={Datasheets for datasets},
  author={Gebru, Timnit and Morgenstern, Jamie and Vecchione, Briana and Vaughan, Jennifer Wortman and Wallach, Hanna and Iii, Hal Daum{\'e} and Crawford, Kate},
  journal={Communications of the ACM},
  volume={64},
  number={12},
  pages={86--92},
  year={2021},
  publisher={ACM New York, NY, USA}
}

\vspace{-5ex}
\begin{IEEEbiography}[{\includegraphics[width=1in,height=1.25in,clip,keepaspectratio]{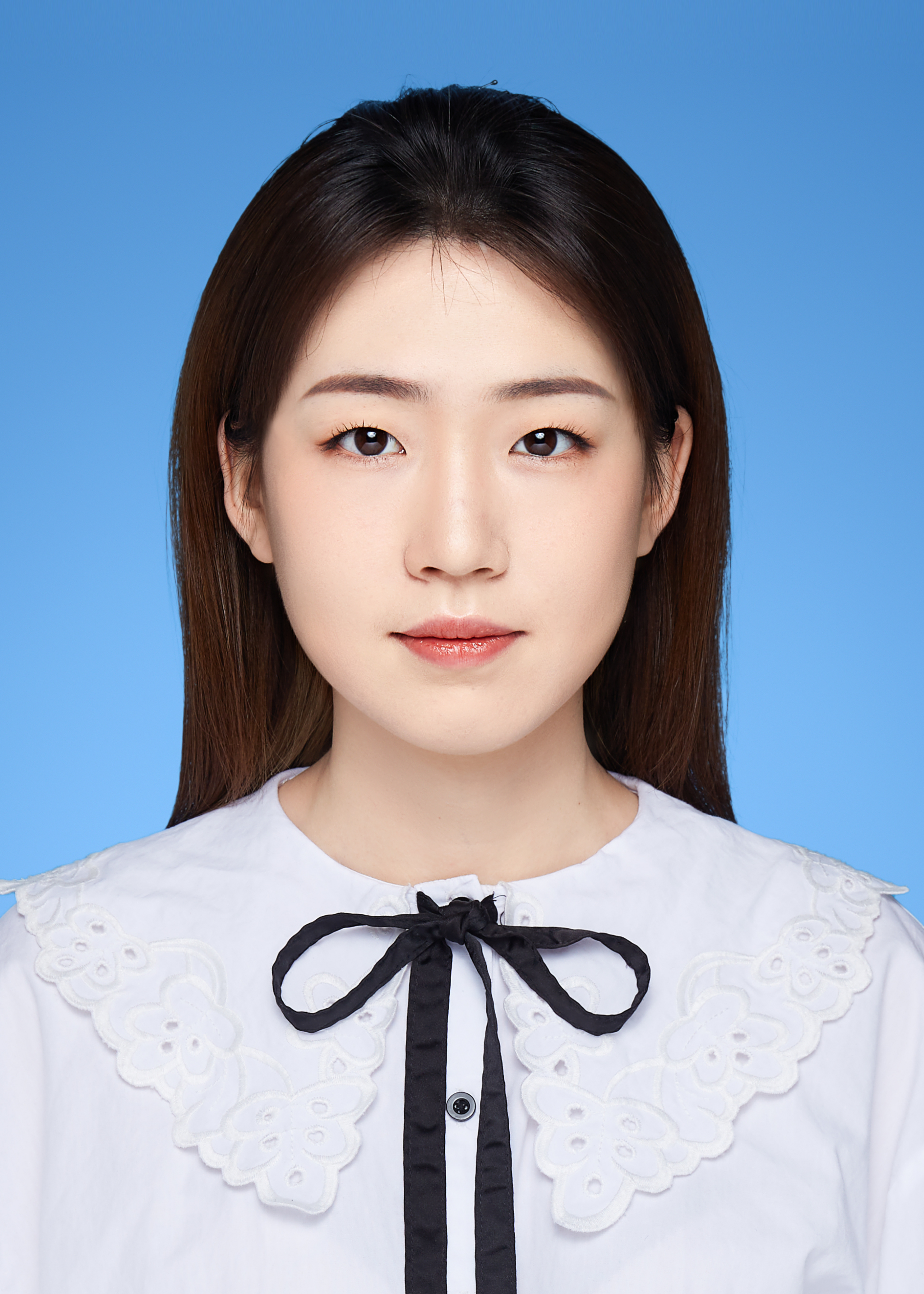}}]{Zixuan~Song} 
     received her M.E. degree in Computer Technology from Jilin University, Changchun, China, in 2023. She is currently pursuing the Ph.D. degree in Computer System Architecture with the College of Computer Science and Technology, Jilin University, Changchun, China. In 2024, she joined Zhongguancun Academy, Beijing, China. Her research interests include sparse Mobile CrowdSensing, Neural Networks, spatiotemporal data mining, and multi-view geo-localization.
\end{IEEEbiography}

\vspace{-5ex}
\begin{IEEEbiography}[{\includegraphics[width=1in,height=1.25in,clip,keepaspectratio]{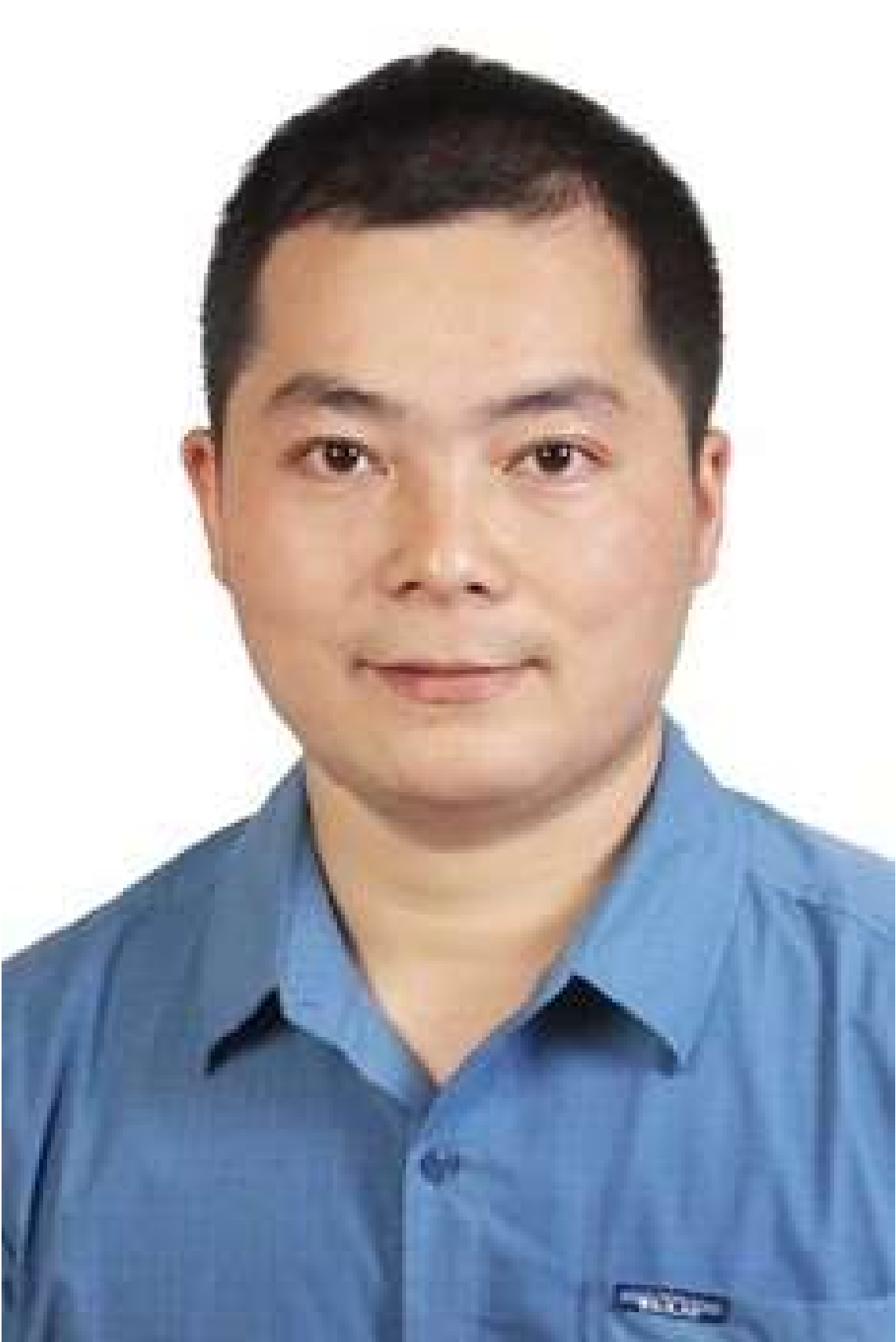}}]{Jing~Zhang}(Senior Member, IEEE) is currently a Professor with the School of Computer Science, Wuhan University, Wuhan, China. Previously, he was a Research Fellow with the School of Computer Science, The University of Sydney. He has published over 100 papers in leading venues, including CVPR, NeurIPS, IEEE Transactions on Pattern Analysis and Machine Intelligence, and IJCV. His research focuses on computer vision and deep learning, and his work has received over 24,000 citations. He serves as an Area Chair for NeurIPS, ICLR, ICML and AAAI. He also regularly serves as a reviewer for top-tier journals and conferences.
\end{IEEEbiography}
\vspace{-5ex}

\begin{IEEEbiography}[{\includegraphics[width=1in,height=1.25in,clip,keepaspectratio]{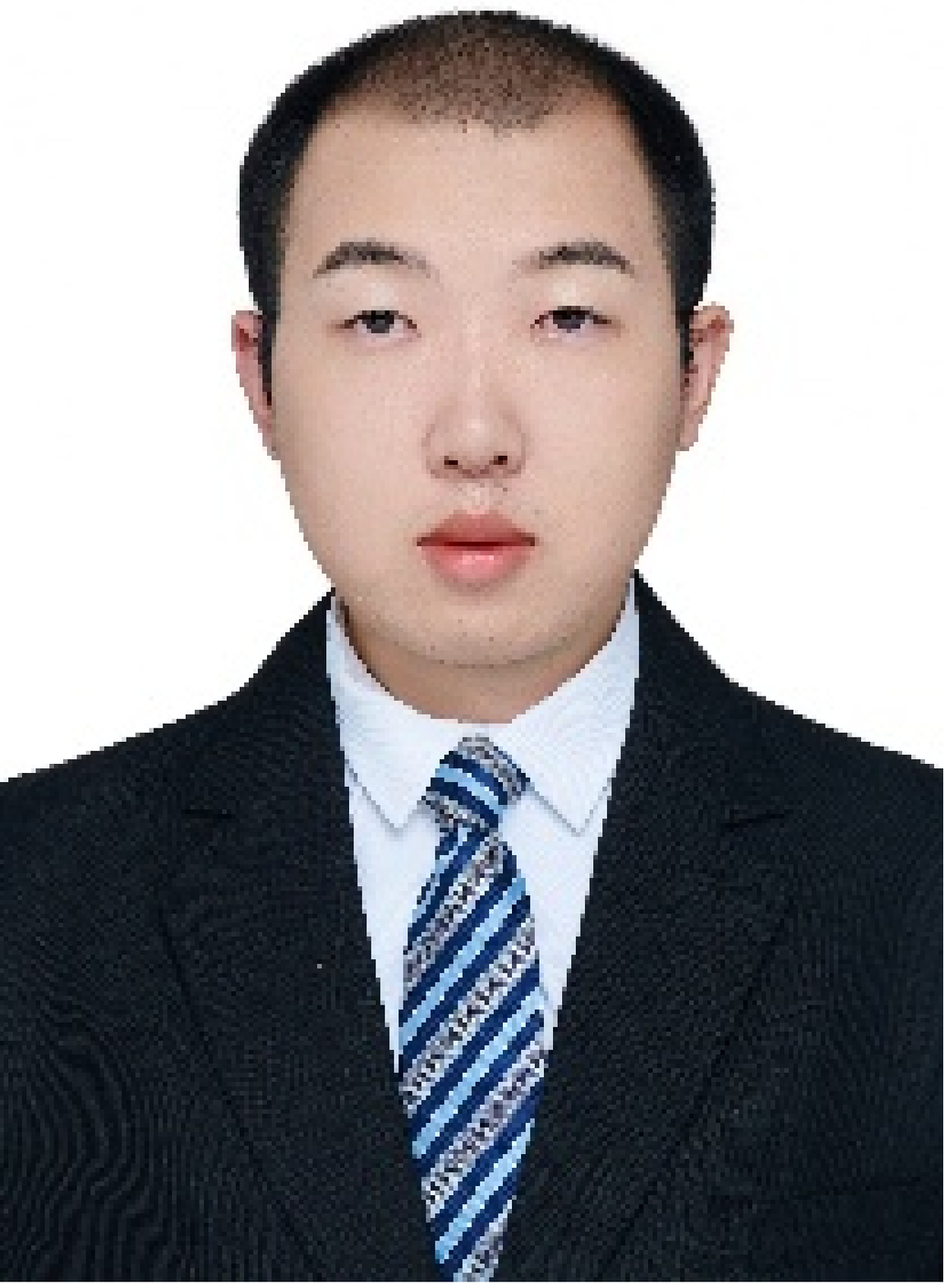}}]{Di~Wang} (Member, IEEE) received the Ph.D. degree from the School of Computer Science, Wuhan University, Wuhan, China, in 2026, where he is currently a Postdoctoral Researcher. His research interests include deep learning, computer vision, and remote sensing.
\end{IEEEbiography}
\vspace{-5ex}

\begin{IEEEbiography}[{\includegraphics[width=1in,height=1.25in,clip,keepaspectratio]{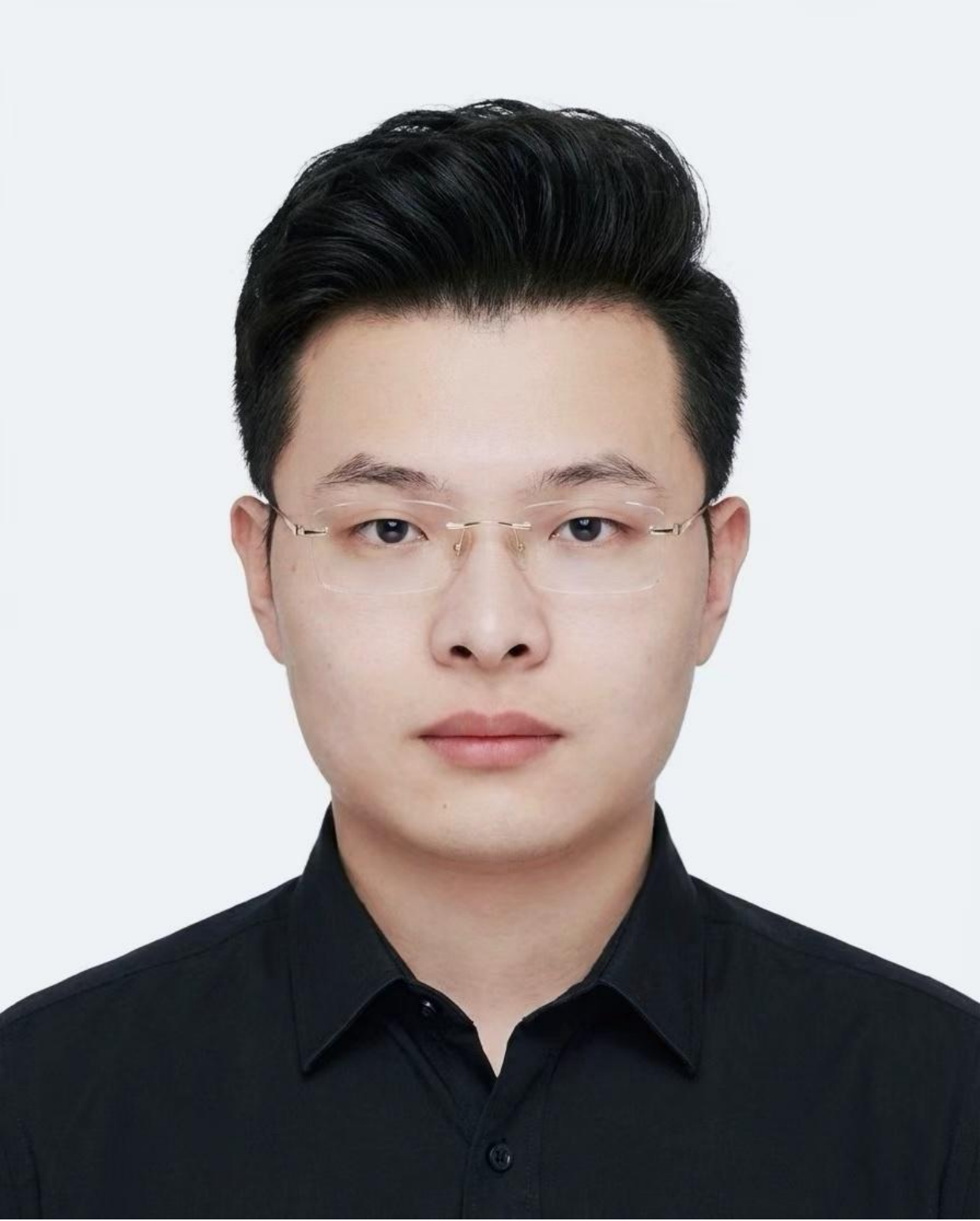}}]{Zhiming~Luo} is currently pursuing his Ph.D. degree at the School of Computer Science, Wuhan University, Wuhan, China. His work has been published and accepted in premier journals, including IEEE Transactions on Geoscience and Remote Sensing and Pattern Recognition. His research interests include image processing, pattern recognition, and multimodal large language models.
\end{IEEEbiography}

\vspace{-5ex}
\begin{IEEEbiography}[{\includegraphics[width=1in,height=1.25in,clip,keepaspectratio]{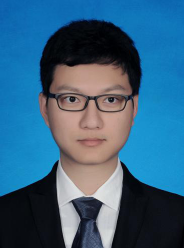}}]{Wenbin Liu}
	(Member, IEEE) received the B.S. degree in physics and the Ph.D. degree in computer science and technology from Jilin University, China, in 2012 and 2020, where he is currently an associate professor with the College of Computer Science and Technology. He was also a visiting PhD student with the Wireless Networks and Multimedia Services Department, Telecom SudParis/Institut Mines-Telecom, France. His research interests include mobile crowdsensing, spatiotemporal crowdsourcing, and ubiquitous computing.
\end{IEEEbiography}

\vspace{-5ex} 
\begin{IEEEbiography}[{\includegraphics[width=1in,height=1.25in,clip,keepaspectratio]{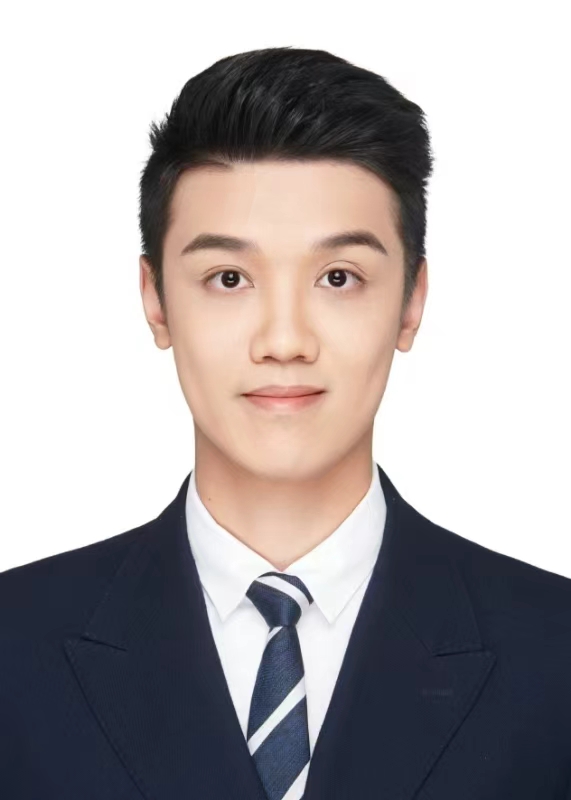}}]{Haonan Guo}
Haonan Guo received the  Ph.D. degree from Wuhan University, Wuhan, China, in 2025 and bachelor's degree from Sun Yat-sen University, Guangzhou, China, in 2020. He is currently a postdoctoral researcher at the State Key Laboratory of Information Engineering in Surveying, Mapping and Remote Sensing, Wuhan University. His research interests include foundation model, building footprint extraction,  and multi-sensor image processing.
\end{IEEEbiography}
\vspace{-5ex} 
\begin{IEEEbiography}[{\includegraphics[width=1in,height=1.25in,clip,keepaspectratio]{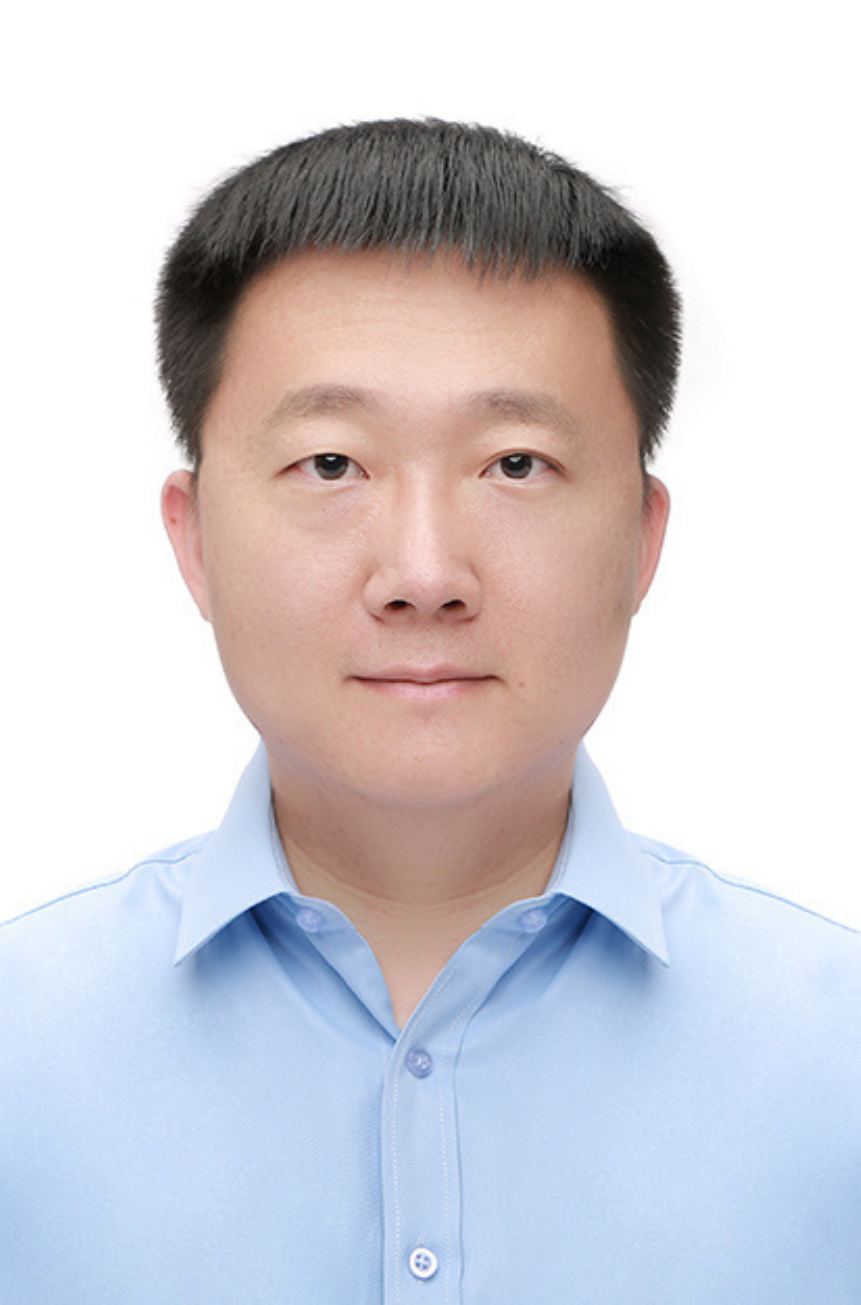}}]{En Wang}
	(Member, IEEE) received his B.E. degree in software engineering from Jilin University, Changchun in 2011, and his M.E. degree and Ph.D. in computer science and technology from Jilin University, Changchun in 2013 and 2016. He was also a joint Ph.D. student with the department of Computer and Information Science, Temple University. He is currently a professor in the Department of Computer Science and Technology at Jilin University, Changchun. His current research focuses on Mobile Computing, Crowd Intelligence and Data Mining.
\end{IEEEbiography}

\vspace{-5ex}

\begin{IEEEbiography}[{\includegraphics[width=1in,height=1.25in,clip,keepaspectratio]{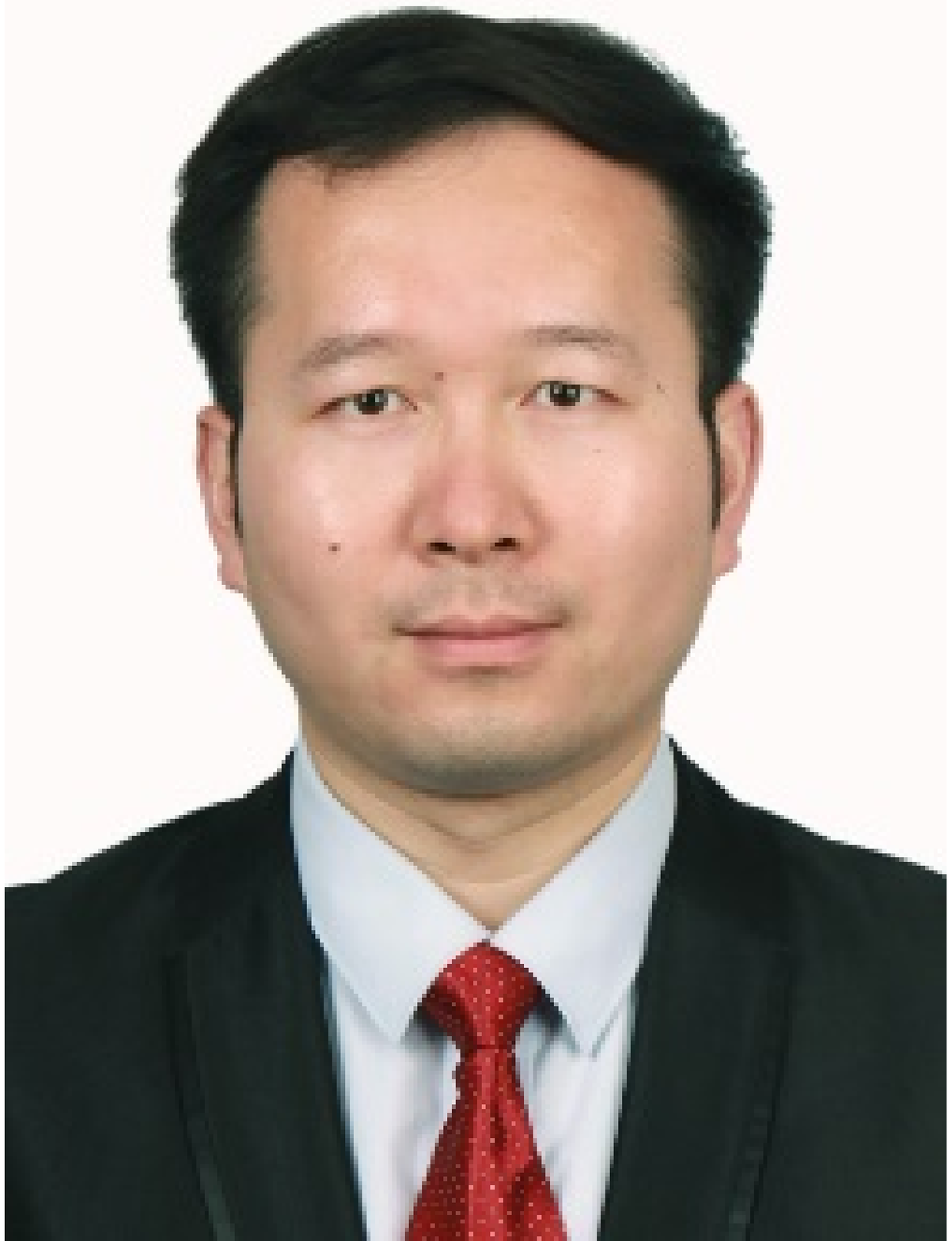}}]{Bo~Du} (Senior Member, IEEE) is currently a Professor with the School of Computer Science and the Institute of Artificial Intelligence, Wuhan University, where he is also the Director with the National Engineering Research Center for Multimedia Software. He has more than 80 research articles published in the IEEE TPAMI, TIP, TCYB, TGRS, JSTARS, and GRSL, and 13 of them are ESI hot papers or highly cited papers. His main research interests include pattern recognition, hyperspectral image processing, and signal processing. He serves as a Senior PC Member for IJCAI and AAAI and the Area Chair for ICPR. He received the Clarivate Highly Cited Researcher (2019–2022), the 2018 IJCAI Distinguished Paper Prize, the 2018 IEEE Data Fusion Contest Champion, the 2018 IEEE WHISPERS Best Paper Award, and the First Place in NLP Competitions GLUE and SuperGLUE in 2022. He serves as an Associate Editor for Neural Networks, Pattern Recognition, and Neurocomputing.
\end{IEEEbiography}

\vspace{-5ex}

\begin{IEEEbiography}[{\includegraphics[width=1in,height=1.25in,clip,keepaspectratio]{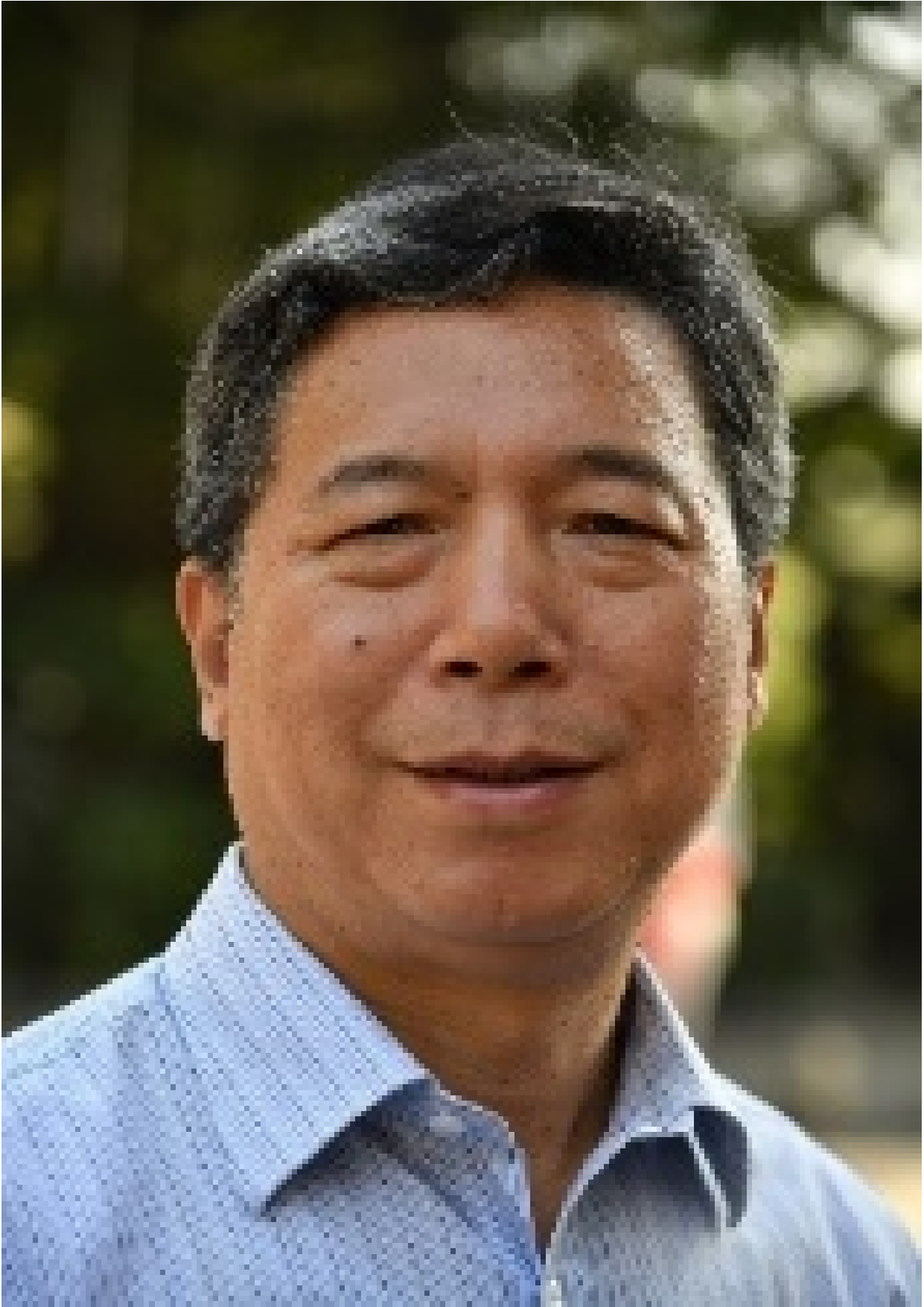}}]{Liangpei~Zhang} (Fellow, IEEE) is currently a “Chang-Jiang Scholar” Chair Professor appointed by the Ministry of Education of China, State Key Laboratory of Information Engineering in Surveying, Mapping, and Remote Sensing (LIESMARS), Wuhan University. He was a Principal Scientist of the China State Key Basic Research Project appointed by the Ministry of National Science and Technology of China to lead the remote sensing program from 2011 to 2016. He has published more than 700 research papers and five books. He is the Institute for Scientific Information (ISI) highly cited author. He is the holder of 30 patents. His research interests include hyperspectral remote sensing, high-resolution remote sensing, image processing, and artificial intelligence. He is also an associate editor for IEEE TGRS. He is a fellow of the Institution of Engineering and Technology (IET).
\end{IEEEbiography}
\vfill
\clearpage

\twocolumn
\newpage

\makeatletter

\newcommand{\repeatmaketitle}{%
  \if@twocolumn
    \twocolumn[{
      \begingroup
      \centering
      \vspace*{3.25\baselineskip}
      {\@IEEEcompsoc@title@font \@title \par}
      \vspace{11pt}
      {\@IEEEcompsoc@author@font
       \def\IEEEauthorhalign##1{\centering##1}
       \@author \par}
      \vspace{2\baselineskip}
      \endgroup
    }]%
  \else
    \thispagestyle{empty}
    \begingroup
    \centering
    \vspace*{3.25\baselineskip}
    {\@IEEEcompsoc@title@font \@title \par}
    \vspace{11pt}
    {\@IEEEcompsoc@author@font \@author \par}
    \vspace{2\baselineskip}
    \endgroup
  \fi
  \captionsetup{
    font=normalfont,       
    textfont=normalfont,   
    size=normalsize,       
}
}

\newcommand{\@IEEEcompsoc@title@font}{%
  \sffamily\Huge\selectfont
  \ifCLASSOPTIONconference\large\fi
}

\newcommand{\@IEEEcompsoc@author@font}{%
  \sffamily\sublargesize\selectfont
  \ifCLASSOPTIONconference\sublargesize\fi
}

\def\IEEEauthorblockN#1{%
  \vskip0.5ex
  {\bfseries #1}\par
}

\def\IEEEauthorblockA#1{%
  \vskip0.25ex
  {\itshape #1}\par
}

\makeatother

\title{GeoBridge++: Fact-Guided Geo-Semantic \mbox{Bridging for Unified Cross-View Geo-Localization}\\
---Supplementary Material---}
\author{Zixuan~Song, Jing~Zhang\textsuperscript{\dag},~\IEEEmembership{Senior Member,~IEEE}, Di~Wang\textsuperscript{\dag}, Zhiming~Luo, Wenbin~Liu, Haonan~Guo, En~Wang\textsuperscript{\dag}, Bo~Du\textsuperscript{\dag},~\IEEEmembership{Senior Member,~IEEE}, Liangpei~Zhang,~\IEEEmembership{Fellow,~IEEE}

\thanks{\textsuperscript{\dag}: Corresponding authors.}
}

\repeatmaketitle

\appendices

\section{Overview}
\label{sec:Overview}
This supplementary material provides supplementary details on GeoBridge, GeoBridge++, GeoLoc, and GeoLoc-MM that are omitted from the main paper due to space constraints. The remainder of the supplementary material is organized as follows:
\begin{itemize}
\item Section~\ref{sec:experiments}. Supplementary experiments for further evaluation of the proposed methods.
\item Section~\ref{sec:dataset details}. Additional details on the construction and preprocessing of GeoLoc and GeoLoc-MM.
\item Section~\ref{sec:instruction datails}. Implementation details of the instruction design and description construction.
\item Section~\ref{sec:examples}. Additional qualitative examples of retrieval and inference results.
\item Section~\ref{sec:datasheets}. Datasheets for GeoLoc and GeoLoc-MM.
\item Section~\ref{sec:app-limitation}. Discussion on limitations and societal impact.
\end{itemize}

\section{Supplementary Experiment}
\label{sec:experiments}
\subsection{Text Description Validation}

\begin{table*}[t]
\caption{Further validation of the text description for GeoBridge on the GeoLoc dataset. D, T, and S denote Drone, Street-View Panorama, and Satellite, respectively. The best results are shown in bold, the second-best results are underlined.}
\centering
\setlength{\tabcolsep}{6pt}
  \footnotesize
    \resizebox{1.0\linewidth}{!}{\begin{tabular}{c*{6}{cc}}
   \hline
    \multirow{2}{*}{\textbf{Method}}
      & \multicolumn{2}{c}{\textbf{D2S}} 
      & \multicolumn{2}{c}{\textbf{S2D}} 
      & \multicolumn{2}{c}{\textbf{T2S}} 
      & \multicolumn{2}{c}{\textbf{S2T}}  & \multicolumn{2}{c}{\textbf{D2T}}  & \multicolumn{2}{c}{\textbf{T2D}} \\
      \cmidrule(lr){2-3}\cmidrule(lr){4-5}\cmidrule(lr){6-7}\cmidrule(lr){8-9}\cmidrule(lr){10-11}\cmidrule(lr){12-13}
     & \textbf{R@1} & \textbf{AP} & \textbf{R@1} & \textbf{AP} & \textbf{R@1} & \textbf{AP} & \textbf{R@1} & \textbf{AP}& \textbf{R@1} & \textbf{AP}& \textbf{R@1} & \textbf{AP} \\
    \hline
    No semantic anchor   & 38.20 & 43.76 & 34.63 & 47.82 & 6.43 & 14.12 & 6.95 & 15.98 &7.16 & 16.26& 4.90 & 12.25\\
    Qwen3~\cite{qwen3}  & \textbf{46.07} & 47.26 & 44.72 & 45.49 & 30.56 & 35.16 & 38.97 & 43.69 & 30.24 & 36.59 & \underline{32.05} & 32.25\\
    Gemini3~\cite{team2023gemini}  &  38.39 & 47.85 & 37.47 & \underline{46.38} & 24.58 & 37.64 & 28.12 & \underline{43.71} & 25.51 & 29.89 & 24.95 & 36.33\\
    Gemini3~\cite{team2023gemini} (Qwen3\cite{qwen3} + GPT-4o~\cite{hurst2024gpt})  &   42.91 & \textbf{53.42} & \underline{41.68} & 41.89 & \underline{31.81} & \textbf{47.33} & 32.41 & \textbf{45.52} & \underline{31.81} & \underline{38.71} & 31.79 & \underline{43.34} \\
     GeoBridge & \underline{45.05} & \underline{49.05} & \textbf{44.81} & \textbf{48.76} & \textbf{38.87} & \underline{42.10} & \textbf{39.20} & 41.96 & \textbf{41.22} & \textbf{43.54} & \textbf{41.15} & \textbf{43.41}\\
   \bottomrule
    \end{tabular}}
    \label{textcaption1}
\end{table*}

Our method does not rely on GPT-4o~\cite{hurst2024gpt}, it uses GPT-4o only as a tool to generate cross-view textual descriptions. To validate this, we also employ text generated by the open-source models Qwen3~\cite{qwen3} and Gemini3~\cite{team2023gemini}, as well as ensemble text produced by Gemini3 combining outputs from Qwen3 and GPT-4o, to serve as semantic anchors in GeoBridge. As shown in Table~\ref{textcaption1}, using text from different models leads to consistent improvements. 

\subsection{Public Subset Construction}

To further verify that our method does not depend on any proprietary data source, we constructed a three-view subset from the publicly available CVUSA~\cite{CVACT} street–satellite geo-localization dataset and generated the corresponding semantic anchors. Specifically, we re-cropped the drone images to 750 × 750 pixels based on the coordinate points and applied the same filtering procedure, yielding 481 image sets, of which 81 were reserved for testing. We obtained the corresponding text descriptions in the same way and conducted experiments with GeoBridge. The results are presented in Table~\ref{table-ablation1}. Owing to the limited size of the dataset, only the projection layer was fine-tuned. The retraining results on this subset consistently show that our method outperforms the baseline.

\begin{table*}[t]
\caption{Comparison on the CVUSA subdataset. D, T, and S denote Drone, Street-View Panorama, and Satellite, respectively. The best results are shown in bold, the second-best results are underlined.}
\centering
\setlength{\tabcolsep}{6pt}
  \footnotesize
   \resizebox{1.0\linewidth}{!}{\begin{tabular}{c*{6}{cc}}
   \hline
    \multirow{2}{*}{\textbf{Method}}
      & \multicolumn{2}{c}{\textbf{D2S}} 
      & \multicolumn{2}{c}{\textbf{S2D}} 
      & \multicolumn{2}{c}{\textbf{T2S}} 
      & \multicolumn{2}{c}{\textbf{S2T}}  & \multicolumn{2}{c}{\textbf{D2T}}  & \multicolumn{2}{c}{\textbf{T2D}} \\
      \cmidrule(lr){2-3}\cmidrule(lr){4-5}\cmidrule(lr){6-7}\cmidrule(lr){8-9}\cmidrule(lr){10-11}\cmidrule(lr){12-13}
     & \textbf{R@1} & \textbf{AP} & \textbf{R@1} & \textbf{AP} & \textbf{R@1} & \textbf{AP} & \textbf{R@1} & \textbf{AP}& \textbf{R@1} & \textbf{AP}& \textbf{R@1} & \textbf{AP} \\
    \hline
    Sample4Geo~\cite{Sample4geo}  & \underline{13.70} & \underline{22.22}  & 7.40 & 16.34 & \underline{7.41} & 14.81  & 8.64 & 12.35 & 13.58 & \underline{17.68}&\underline{12.35} & \underline{17.37}\\
    MEAN~\cite{MEAN}  & 12.35 & 17.37 & 13.58 & \underline{19.60}& -&-&-&- & 4.41 & 7.27 & 5.23 & 8.94 \\
    MCCG~\cite{MCCG}   & 10.27 & 14.22 &\underline{14.17} & 17.89 & -&-&-&-  & 2.63 & 6.31& 5.58 & 7.16\\
    panorama-BEV~\cite{ye2024cross}  &  -&-&-&- & 4.94 & 8.64 & \underline{11.11} & 17.28 &6.17 & 14.81& 9.88 & 15.60 \\
    AuxGeo~\cite{AuxGeo}  &  -&-&-&-& 3.07 & \underline{18.93} & 7.48 & \underline{21.80} &\underline{13.74} & 14.95 & 10.00 & 12.49 \\
     GeoBridge & \textbf{49.38} & \textbf{62.46} & \textbf{49.39} & \textbf{63.17} & \textbf{28.40} & \textbf{43.68} & \textbf{28.39} & \textbf{42.42} & \textbf{20.99} & \textbf{31.52} & \textbf{16.05} & \textbf{31.03}\\
   \bottomrule
    \end{tabular}}
    \label{table-ablation1}
\end{table*}

\subsection{Additional Ablation Results}

In the main paper, we evaluate different alignment strategies of GeoBridge using \(R@1\). To provide a complementary evaluation of retrieval performance, we further report the corresponding \(AP\) results under the same experimental setting. As shown in Table~\ref{tab:ablation_GeoBridge}, Image-only denotes image--image alignment, Text-only denotes text--image alignment using the semantic anchor, and GeoBridge jointly performs both types of alignment. These results are consistent with the \(R@1\) observations in the main paper and further demonstrate that semantic-anchor alignment complements visual correspondence learning for cross-view geo-localization.

We further complement the ablation studies of GeoBridge++ with the corresponding \(AP\) results. Table~\ref{tab:ablation_FactBridge_AP} compares location fact guidance, geo-semantic bridge alignment, and direct pairwise image alignment under the same setting as in the main paper. Fact guidance provides more effective supervision than direct pairwise alignment, while the geo-semantic bridge further facilitates the coordination of heterogeneous views. Their combination achieves the strongest overall performance, consistent with the \(R@1\) results reported in the main paper.

Table~\ref{tab:ablation_fact_AP} reports the \(AP\) results for different combinations of geographic (G), visual (V), and shared geo-visual (SH) facts. Geographic and visual facts provide complementary modality-aware supervision, while shared geo-visual facts establish a common reference across all views. Jointly incorporating all three types of facts yields the best overall performance, further supporting the complementary roles of modality-specific and shared factual supervision.

\begin{table}[t]
  \centering
  \caption{Ablation \(AP\) results of GeoBridge with different alignment strategies on GeoLoc dataset. D, T, and S denote Drone, Street-View Panorama, and Satellite, respectively. The best results are shown in bold, the second-best results are underlined.}
  \label{tab:ablation_GeoBridge}
  \resizebox{\columnwidth}{!}{
  \begin{tabular}{ccccccc}
    \toprule
    \textbf{Method}& \textbf{D2S} & \textbf{S2D} & \textbf{T2S} & \textbf{S2T} & \textbf{D2T} & \textbf{T2D} \\
    \midrule
    Image-only    & 43.76 &\underline{47.83}& 14.12 &15.98&16.26 &12.25 \\
    Text-only    &\underline{46.49}  &46.49 &  \underline{38.70}&\underline{39.50}&\underline{41.21}&\underline{41.05} \\
    \midrule
    GeoBridge&  \textbf{49.05}& \textbf{48.76} & \textbf{42.10} &  \textbf{41.96}&  \textbf{43.54} & \textbf{43.41} \\
    \bottomrule
  \end{tabular}
    }
\end{table}

\begin{table*}[t]
\centering
\caption{Ablation \(AP\) results of GeoBridge++ with different strategies on the map-augmented GeoLoc dataset. Fact, Bridge, Pairwise denote location fact guidance, geo-semantic bridge alignment, and direct pairwise image alignment, respectively. D, T, S, and M denote Drone, Street-View Panorama, Satellite, and Static Map, respectively. The best results are shown in bold, the second-best results are underlined.}
\label{tab:ablation_FactBridge_AP}
\resizebox{\textwidth}{!}{
\begin{tabular}{ccc|cccccccccccccccccc}
\toprule
\textbf{Fact} & \textbf{Bridge} & \textbf{Pairwise} &
\textbf{D2S} &
\textbf{S2D} &
\textbf{T2S} &
\textbf{S2T} &
\textbf{D2T} &
\textbf{T2D} &
\textbf{D2M} &
\textbf{M2D} &
\textbf{S2M} &
\textbf{M2S} &
\textbf{T2M} &
\textbf{M2T} \\
\midrule
  \checkmark&  &  &  \underline{62.42} &  \underline{62.92} &  \underline{59.05}&  \underline{57.79}& \underline{59.78}& \underline{60.41}& \underline{61.17}&\underline{58.00} & \underline{59.35}&\underline{53.46}&\underline{41.87}&\underline{41.44}\\
  &\checkmark  & &  39.18&  39.75 &  16.44 & 14.79&  14.69&  16.47&  26.38&26.41&31.18&29.81&11.20&9.84 \\
  & &\checkmark   &35.69 & 33.93 &  3.38 &  2.52&  2.47&  2.69&  29.48&30.26&26.99&28.84&2.53&23.35 \\
  \checkmark& &  \checkmark &57.72 &  59.59 &  42.46 &  40.88&  41.61&  46.03& 45.83&44.45&26.15&48.06&39.56&34.46 \\
  \midrule
  \checkmark  & \checkmark& &\textbf{71.01}&  \textbf{70.66} &  \textbf{64.69} &  \textbf{61.35}&  \textbf{69.29}&  \textbf{71.64}&  \textbf{63.21}&\textbf{58.31}&\textbf{60.67}&\textbf{59.85}&\textbf{55.23}&\textbf{52.62} \\
\bottomrule
\end{tabular}}
\end{table*}

\begin{table*}[t]
\centering
\caption{Ablation \(AP\) results of GeoBridge++ with different types of location fact guidance on the map-augmented GeoLoc dataset. G, V, and SH denote geographic, visual, and shared geo-visual facts, respectively. D, T, S, and M denote Drone, Street-View Panorama, Satellite, and Static Map, respectively. The best results are shown in bold, the second-best results are underlined.}
\label{tab:ablation_fact_AP}
\resizebox{\textwidth}{!}{
\begin{tabular}{ccc|cccccccccccccccccc}
\toprule
\textbf{G} & \textbf{V} & \textbf{SH} &
\textbf{D2S} &
\textbf{S2D} &
\textbf{T2S} &
\textbf{S2T} &
\textbf{D2T} &
\textbf{T2D} &
\textbf{D2M} &
\textbf{M2D} &
\textbf{S2M} &
\textbf{M2S} &
\textbf{T2M}&
\textbf{M2T} \\

\midrule
  \checkmark&  &  &  60.88  & 63.17 &  --&  --&  --& --& \underline{55.45}&54.62&51.47&49.11&--&--\\
  &\checkmark  & &  64.71&  63.47 &  54.72 &  \underline{55.52}&  \underline{57.29}&  \underline{57.79}&  --&--&--&--&--&-- \\
  & &\checkmark   &61.54 &  65.22 &  \underline{39.85} &  41.69&  42.44&  43.25&  51.83&\underline{55.15}&\underline{53.25}&\underline{52.02}&\underline{35.05}&\underline{37.40} \\
  \checkmark& \checkmark&   &\underline{65.18} &  \underline{66.53} & 30.49 &  44.79&  35.96& 34.00& 46.61&40.17&41.90&36.35&18.85&18.77 \\
  \midrule
  \checkmark  & \checkmark& \checkmark&\textbf{71.01}&  \textbf{70.66} & \textbf{ 64.69} & \textbf{ 61.35}&  \textbf{69.29}&  \textbf{71.64}&  \textbf{63.21}&\textbf{58.31}&\textbf{60.67}&\textbf{59.85}&\textbf{55.23}&\textbf{52.62} \\
\bottomrule
\end{tabular}}
\end{table*}

\section{Dataset Construction and Processing}
\label{sec:dataset details}

\subsection{GeoLoc Dataset}

After acquiring the GeoLoc data, we performed systematic, multi-stage cleaning and quality control. Compared with the description in the main paper, this section provides more fine-grained operational details. It is worth noting that each step of the pipeline is manually monitored, and the overall process requires approximately 150 hours of human effort.

\subsubsection{Drone Data Acquisition and Cropping}

\begin{figure}[t]
  \centering

  \subfloat{\label{fig:sub1}}{%
    \includegraphics[width=0.155\columnwidth]{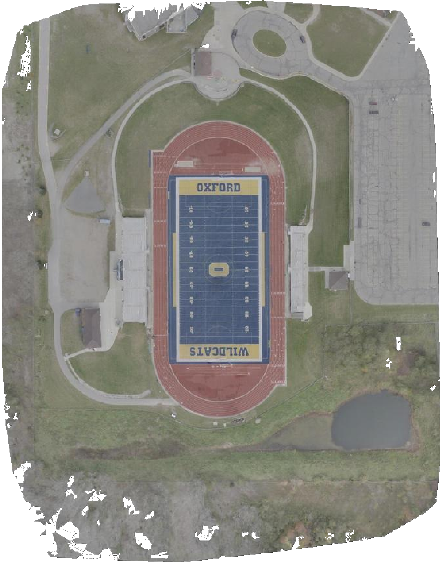}%
  }\hfill
  \subfloat{\label{fig:sub2}}{%
    \includegraphics[width=0.155\columnwidth]{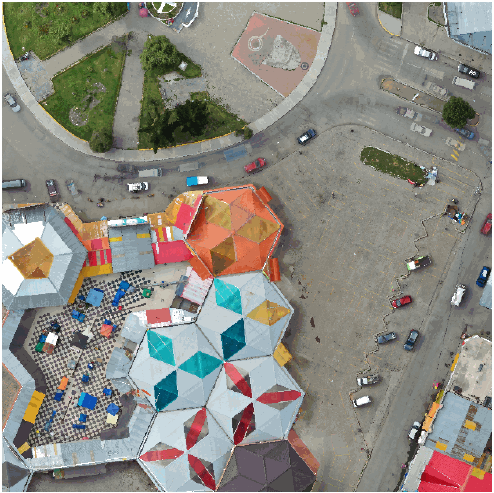}%
  }\hfill
  \subfloat{\label{fig:sub3}}{%
    \includegraphics[width=0.155\columnwidth]{c.pdf}%
  }\hfill
  \subfloat{\label{fig:sub4}}{%
    \includegraphics[width=0.155\columnwidth]{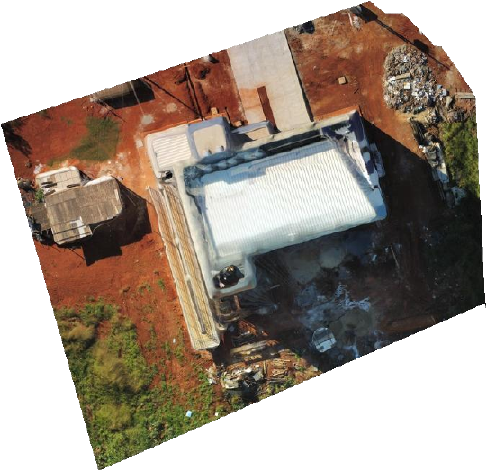}%
  }\hfill
  \subfloat{\label{fig:sub5}}{%
    \includegraphics[width=0.155\columnwidth]{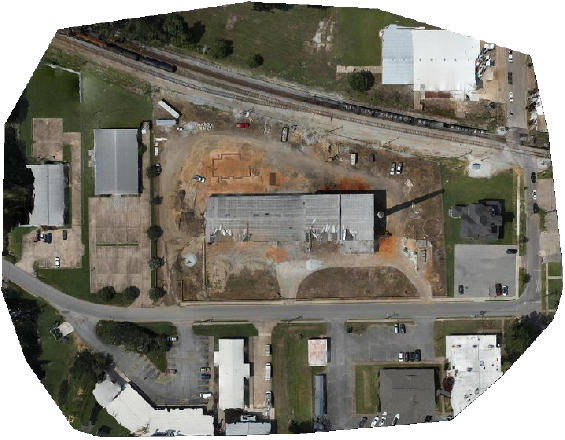}%
  }\hfill
  \subfloat{\label{fig:sub6}}{%
    \includegraphics[width=0.155\columnwidth]{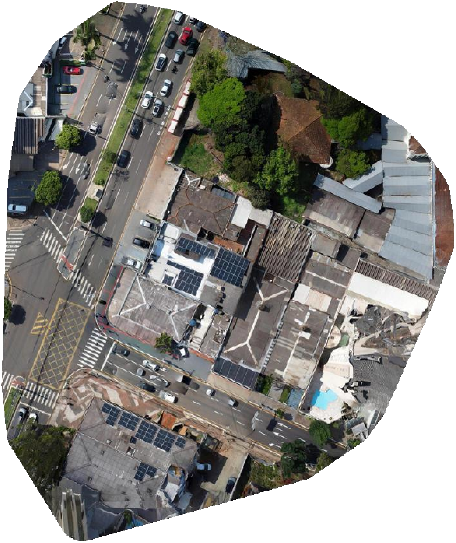}%
  }

  \subfloat{\label{fig:sub7}}{%
    \includegraphics[width=0.155\columnwidth]{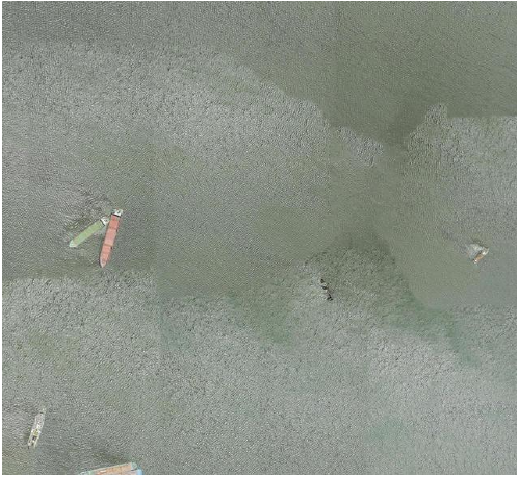}%
  }\hfill
  \subfloat{\label{fig:sub8}}{%
    \includegraphics[width=0.155\columnwidth]{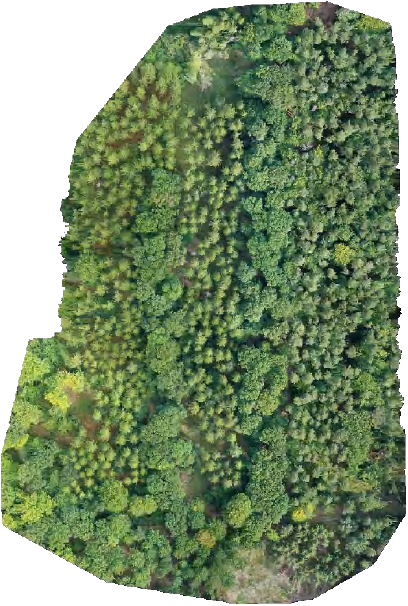}%
  }\hfill
  \subfloat{\label{fig:sub9}}{%
    \includegraphics[width=0.155\columnwidth]{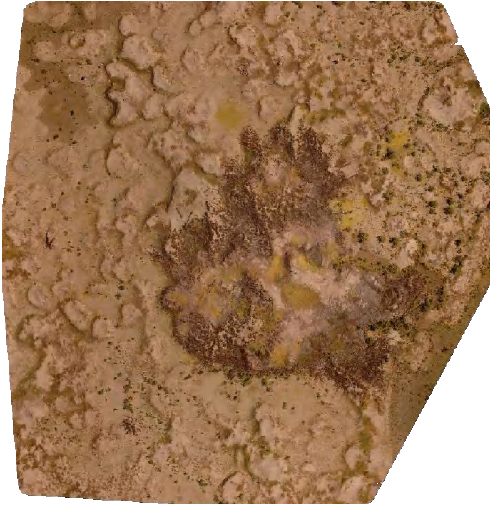}%
  }\hfill
  \subfloat{\label{fig:sub10}}{%
    \includegraphics[width=0.155\columnwidth]{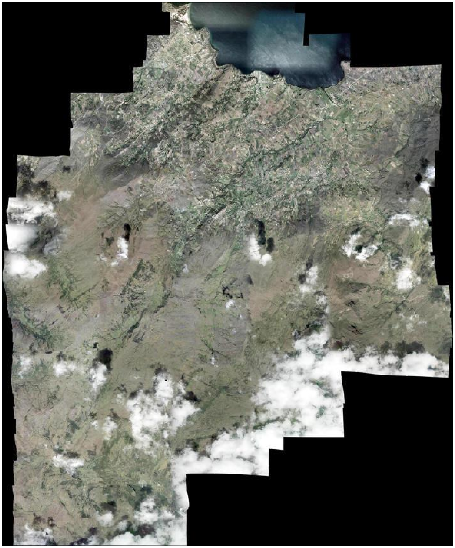}%
  }\hfill
  \subfloat{\label{fig:sub11}}{%
    \includegraphics[width=0.155\columnwidth]{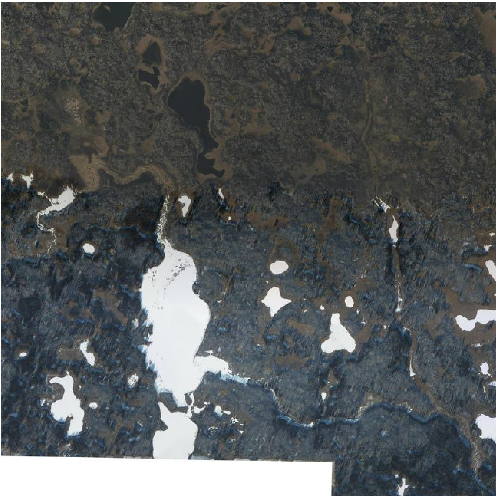}%
  }\hfill
  \subfloat{\label{fig:sub12}}{%
    \includegraphics[width=0.155\columnwidth]{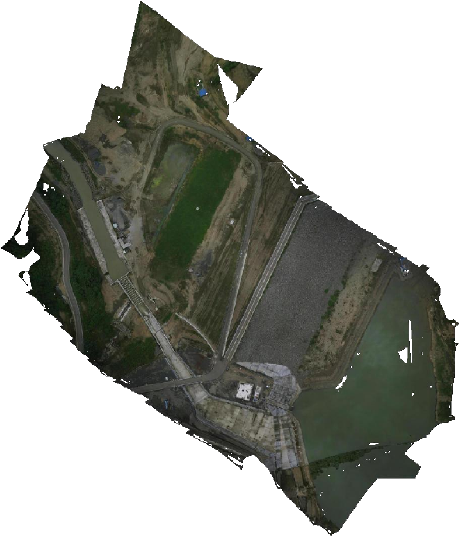}%
  }

  \caption{Examples of original drone images.}
  \label{fig:original-example}
    \vspace{-13pt}
\end{figure}

The original drone imagery used in GeoLoc is illustrated in Fig.~\ref{fig:original-example}. It covers a wide range of scene types, including regions with prominent structural features (Fig.~\ref{fig:sub1},~\ref{fig:sub2}), locations with highly similar ground patterns (Fig.~\ref{fig:sub3},~\ref{fig:sub4}), and samples captured from different flight altitudes, resulting in varying perspectives (Fig.~\ref{fig:sub5},~\ref{fig:sub6}). It also contains large-scale natural environments with limited discriminative structure, such as extensive water bodies, forests, grasslands, and deserts (Fig.~\ref{fig:sub7},~\ref{fig:sub8},~\ref{fig:sub9},~\ref{fig:sub10},~\ref{fig:sub11}). In addition, due to diverse acquisition conditions, some images exhibit degraded quality issues such as low illumination (Fig.~\ref{fig:sub12}).

In the initial stage of data construction, we manually inspected all original drone images and removed those with poor visual quality or dominated by large natural areas lacking clearly identifiable ground features (\eg, Fig.~\ref{fig:sub7},~\ref{fig:sub8},~\ref{fig:sub9},~\ref{fig:sub10}~\ref{fig:sub11}, and~\ref{fig:sub12}). This manual screening process required approximately 20 hours in total.

To obtain drone-view images with precise geometric structure and strict correspondence to street-view imagery, we applied a two-stage cropping and alignment procedure to the original drone data.

First, since the availability of street-view imagery at specific geographic coordinates was unknown a priori, we applied a sliding-window cropping operation over the original drone imagery using a preset \(80 \times 80\) pixel window. For each cropped region, we queried the Google Street View Static to obtain candidate street-view coordinates. For locations where a valid street-view image was found, the associated coordinates were then used as the center for a reverse cropping operation on the original drone image. In this way, the street-view viewpoint is approximately centered within the corresponding drone sub-image, yielding more precise spatial alignment between the two views.

Furthermore, to enhance scene diversity and scale robustness, we did not crop patches based on a fixed image resolution. Instead, we defined cropping windows according to different ground coverage areas. This design enables us to capture multi-scale perspectives at varying resolutions and flight altitudes while maintaining comparability in geographic scale, which is beneficial for learning more generalizable cross-view representations. We obtain drone subimages covering ground areas of approximately \(80 \times 80\), \(100 \times 100\), \(120 \times 120\), \(150 \times 150\), and \(180 \times 180\) \((m^2)\).

\subsubsection{Basic validity screening}

After obtaining the initial cropped drone sub-images, we first perform deduplication based on spatial coverage. For any pair of sub-images whose ground coverage overlaps by more than \(50\%\), we retain only one representative sample and discard the others as duplicates. 

\begin{figure}[t]
  \centering
   \includegraphics[width=1\linewidth]{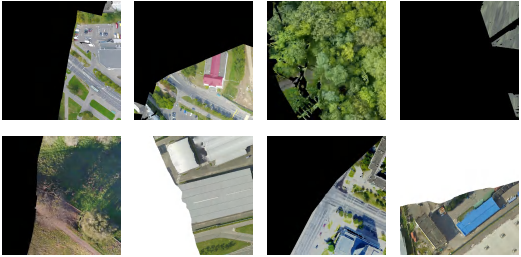}
   \caption{Examples of basic validity screening.}
   \label{fig:black}
     \vspace{-13pt}
\end{figure}

We then conduct basic quality checks to ensure that each image contains a sufficient number of valid pixels and adequate spatial resolution. Subimages that are too small to clearly depict meaningful ground structures are removed. In addition, some images may contain large pure-black or pure-white regions due to sensor failures, rendering errors, or cropping at the image borders. For each image, we compute the proportion of pure-black and pure-white pixels. Images in which the proportion of pure-black or pure-white pixels exceeds \(1\%\) are considered invalid and discarded. Representative examples of such removed samples are shown in Fig.~\ref{fig:black}.

\subsubsection{Multi-Metric Image Quality Filtering}
We applied a three-stage quality filtering process to further refine the drone imagery.

\begin{figure}[t]
  \centering
   \includegraphics[width=1\linewidth]{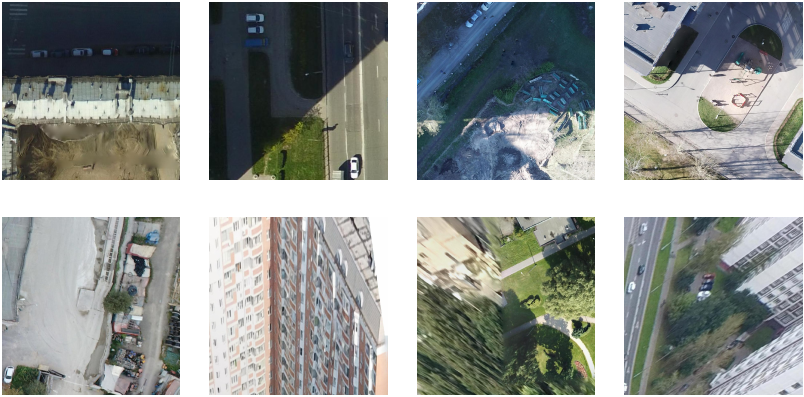}
   \caption{Examples of blurry drone subimages.}
   \label{fig:BH}
     \vspace{-13pt}
\end{figure}

\textbf{BH-Gate: Filtering Blur and Low-Texture Images.} This module primarily targets blurry images (\eg, motion blur, defocus), low-texture scenes caused by haze, cloud cover, or uneven illumination, as well as images with severe compression artifacts that lead to substantial detail loss. BH-Gate combines global pixel variance with an image sharpness measure to detect the absence of meaningful spatial detail. As illustrated in Fig.\ref{fig:BH}, when an image exhibits extremely low texture variation, it is deemed to contain insufficient visual information and is directly discarded.

\begin{figure}[t]
  \centering
   \includegraphics[width=1\linewidth]{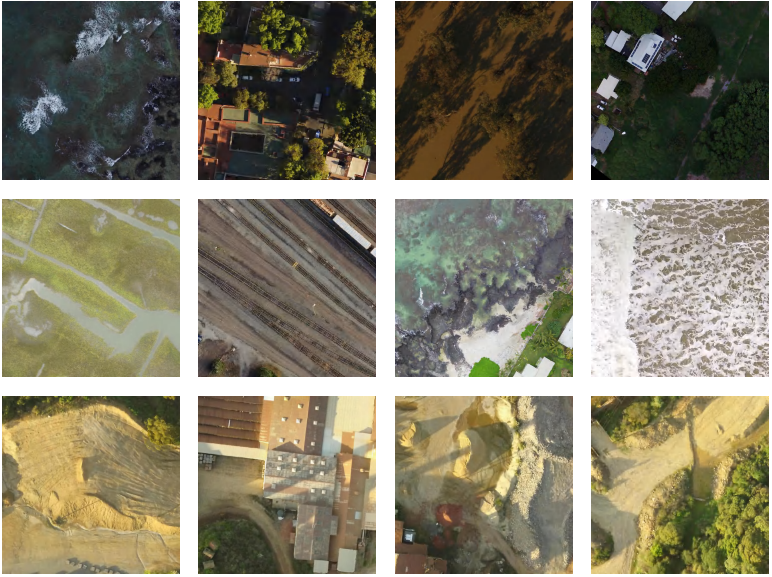}
   \caption{Examples of low global-contrast drone subimages}
   \label{fig:global-contrast}
       \vspace{-13pt}
\end{figure}

\textbf{C-Gate: Filtering Images with Insufficient Global Contrast.} Although some images pass the first-stage screening, they can still exhibit slight blurring or brightness drift. C-Gate targets such cases by identifying images with an insufficient grayscale dynamic range. These are typically overexposed or underexposed images whose grayscale values are highly concentrated, as well as mildly blurred images that retain only weak structural cues. To this end, we analyze the dispersion of the grayscale distribution (\ie, global contrast). As shown in Fig.~\ref{fig:global-contrast}, such images usually fail to provide informative ground textures and contribute little to cross-view and cross-modal learning, and are further filtered out.

\begin{figure}[t]
  \centering
   \includegraphics[width=1\linewidth]{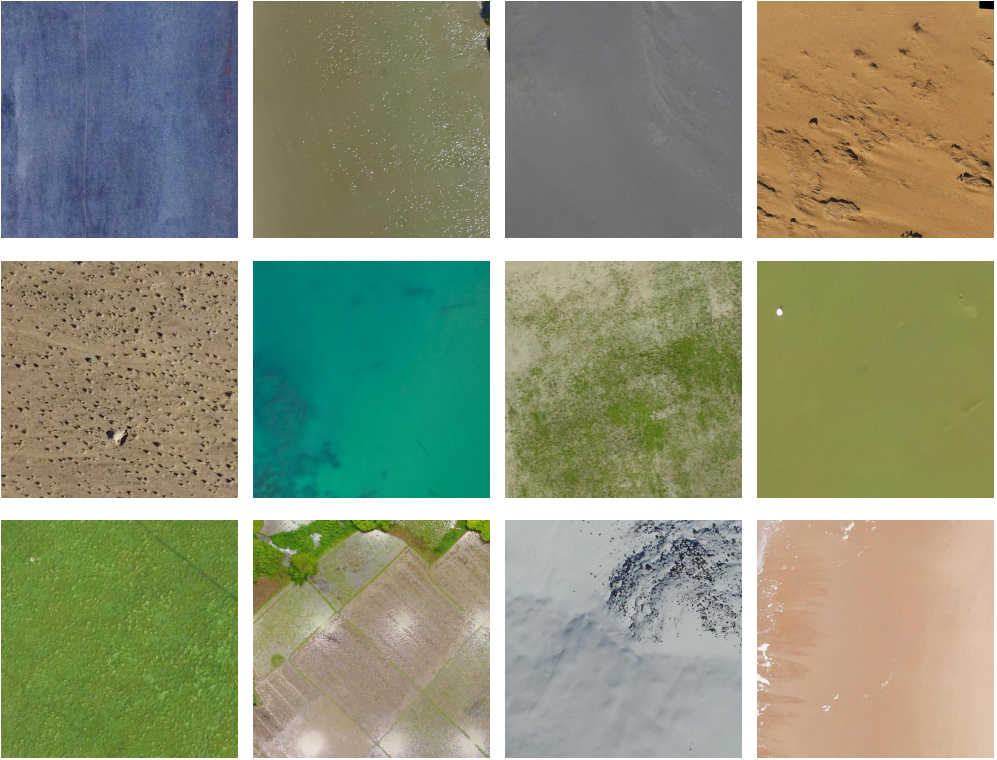}
   \caption{Examples of uniform-texture and noisy pseudo-texture drone subimages}
   \label{fig:UN-Gate}
    \vspace{-13pt}
\end{figure}

\textbf{UN-Gate: Filtering Uniform-Texture and Noisy Pseudo-Texture Images.} After the first two stages of filtering, there may still remain visually near-uniform yet structurally uninformative scenes (\eg, large expanses of water, sky, grassland, desert, or snow), as well as sensor-induced pseudo-textures or noise patterns (including white noise, mosaic artifacts, stripe noise, and random texture blocks). To handle these challenging cases, UN-Gate further combines information entropy, variance range, and the proportion of saturated pixels to assess the information content of each image, distinguish pseudo-texture noise, and detect images that are close to white-noise patterns. Based on these metrics, we determine whether an image falls into the categories of no effective semantics or noisy pseudo-texture. As shown in Fig.\ref{fig:UN-Gate}, images that exhibit highly uniform textures or textures dominated by noise are discarded at this stage.

\subsubsection{Tri-view Alignment}
After completing the filtering of all drone sub-images, we download the corresponding street-view panoramic images and satellite images using the candidate coordinates associated with the retained drone samples.

\begin{figure}[t]
  \centering
   \includegraphics[width=1\linewidth]{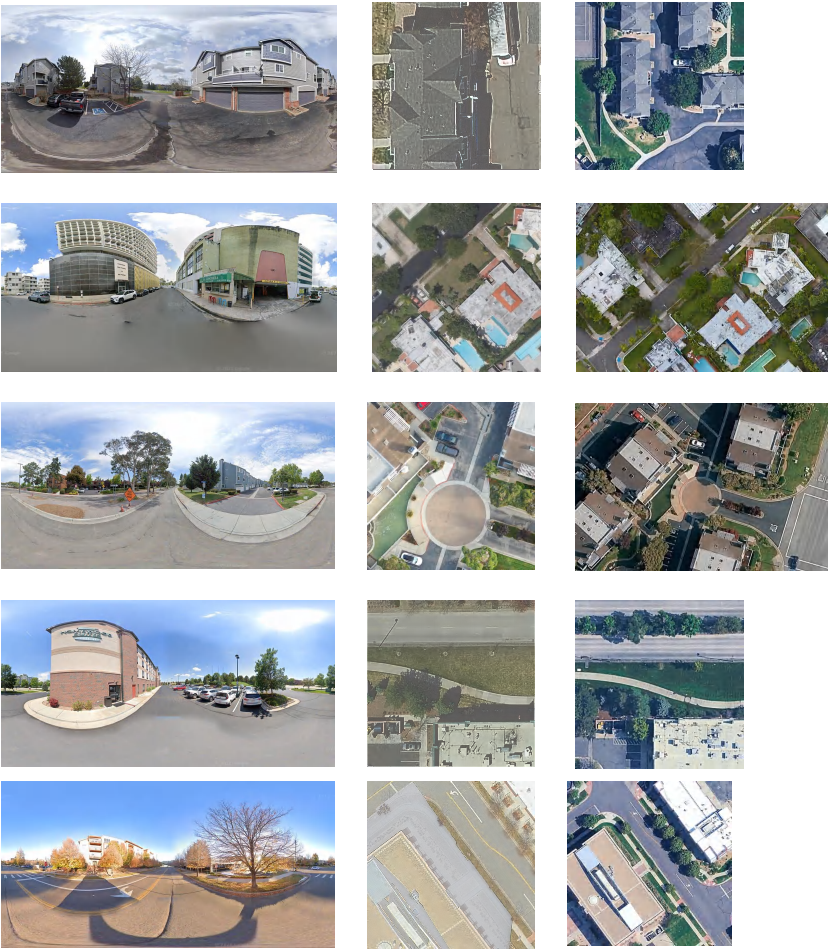}
   \caption{Examples of aligned tri-view images.}
   \label{fig:all}
     \vspace{-13pt}
\end{figure}

When downloading street-view images, we query the Google Street View Static API at the specified panorama location, with the camera heading fixed to face north, the pitch set to eye level, and the field of view set to 120°. Under this configuration, the API returns a street-view image rendered from the high-resolution panorama at that location, facing a consistent direction. For satellite imagery, we retrieve Google Maps satellite images covering the geographic extent corresponding to the latitude–longitude bounds of each retained drone subimage. Examples of the resulting tri-view correspondence (drone, street-view, and satellite) are shown in Fig.~\ref{fig:all}.

\subsection{GeoLoc-MM Dataset}

In GeoLoc-MM, each geographic location is organized around a single location identity shared by all observations. The street-view panorama provides the corresponding ground-level observation, while the drone, satellite, and static-map views characterize the same location at multiple spatial extents. All observations associated with the same location therefore consistently maintain geographic correspondence, although their visible content and representation forms may vary substantially across views and scales.

\begin{figure*}[t]
  \centering
   \includegraphics[width=0.9\linewidth]{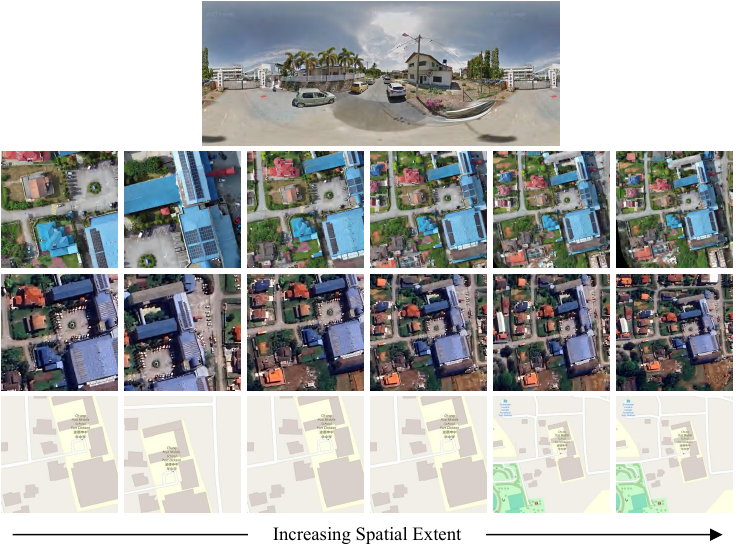}
   \caption{Example of multi-view and multi-scale observations in GeoLoc-MM.}
   \label{fig:geoloc_mm_examples}
    \vspace{-13pt}
\end{figure*}

Figure~\ref{fig:geoloc_mm_examples} presents representative multi-scale groups at six spatial extents, \ie, 80, 100, 120, 150, 180, and 200~\(m\). As the spatial extent increases, the overhead and map observations incorporate progressively broader surrounding context. This organization separates scale-induced variation from differences caused by viewpoint or representation form, providing a controlled setting for evaluating cross-view correspondence under heterogeneous spatial coverage.

\section{Instruction Design and Fact Construction}\label{sec:instruction datails}

\subsection{Instruction Details for the GeoLoc Dataset}
\begin{figure*}[t]
  \centering
   \includegraphics[width=1\linewidth]{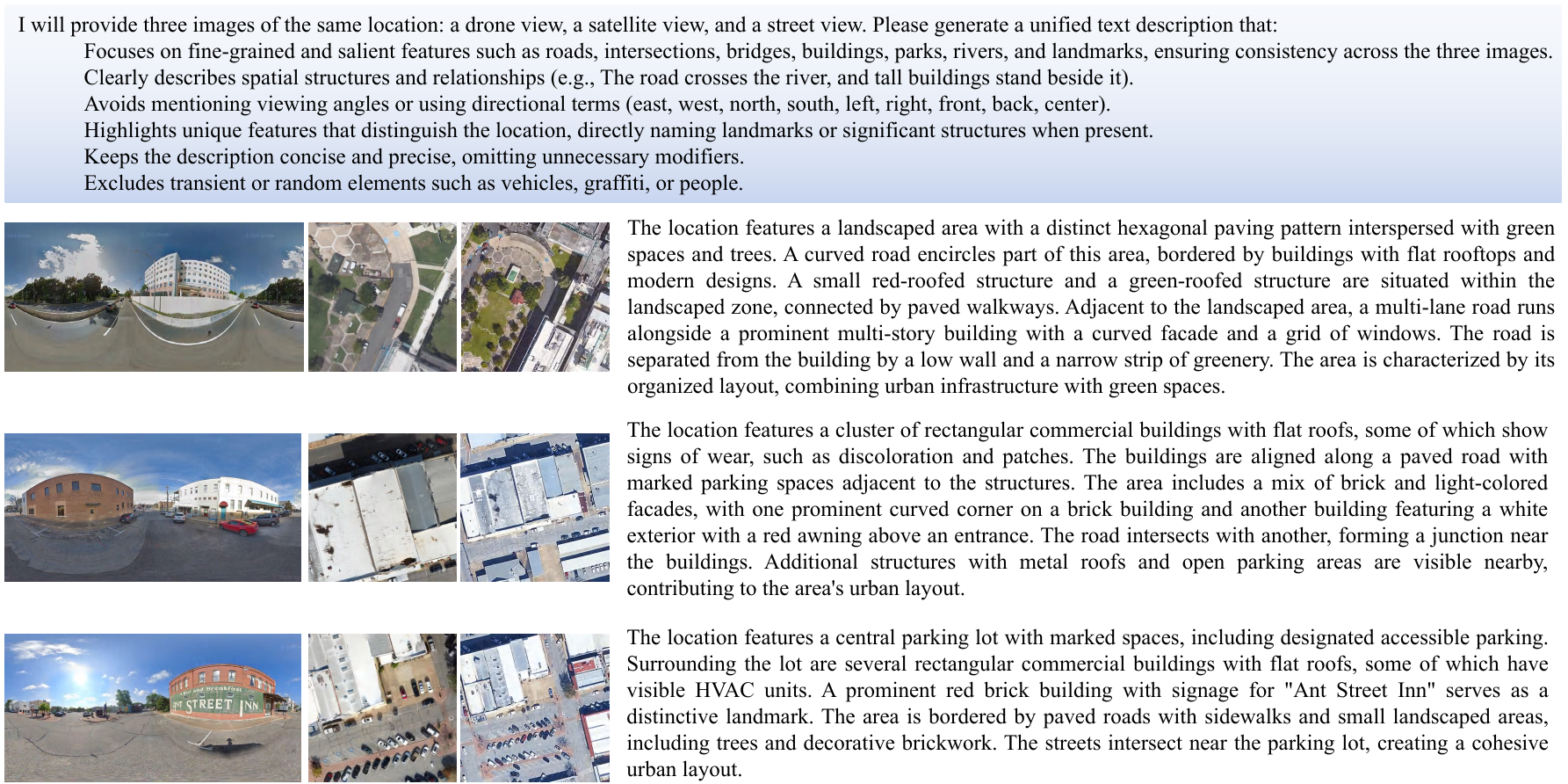}
   \caption{Tri-View instruction protocol for generating unified semantic descriptions. The blue text box denotes the instruction prompt. The first column presents the street-view panorama, the second column shows the drone-view image, the third column displays the satellite image, and the fourth column contains the generated textual description.}
   \label{fig:tri-view-des}
\end{figure*}

\begin{figure*}[t]
  \centering
   \includegraphics[width=1\linewidth]{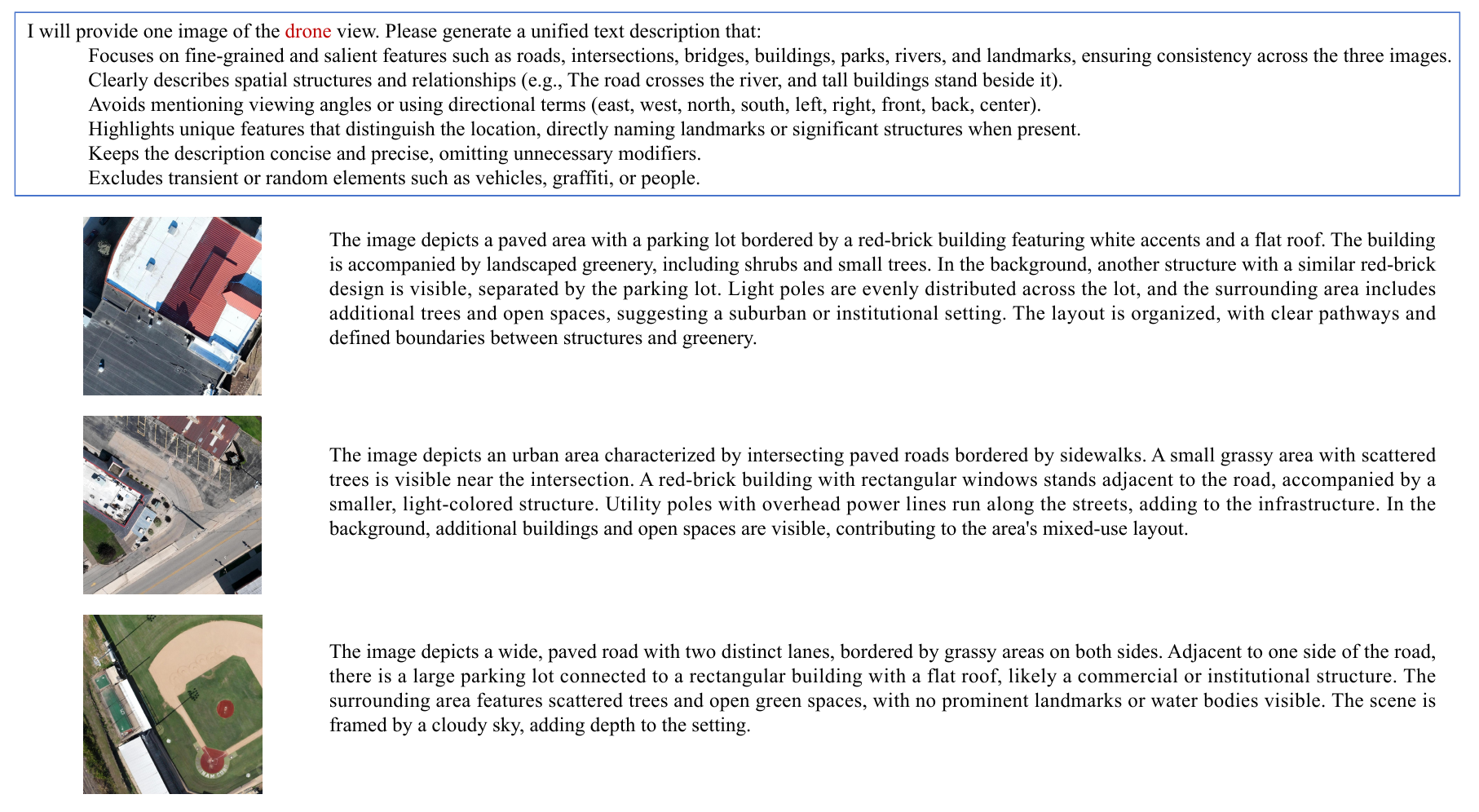}
   \caption{Cross-modal geo-location drone image description instructions, with blue text boxes indicating the prompts.}
   \label{fig:drone-des}
    \vspace{-13pt}
\end{figure*}

\begin{figure*}[h]
  \centering
   \includegraphics[width=1\linewidth]{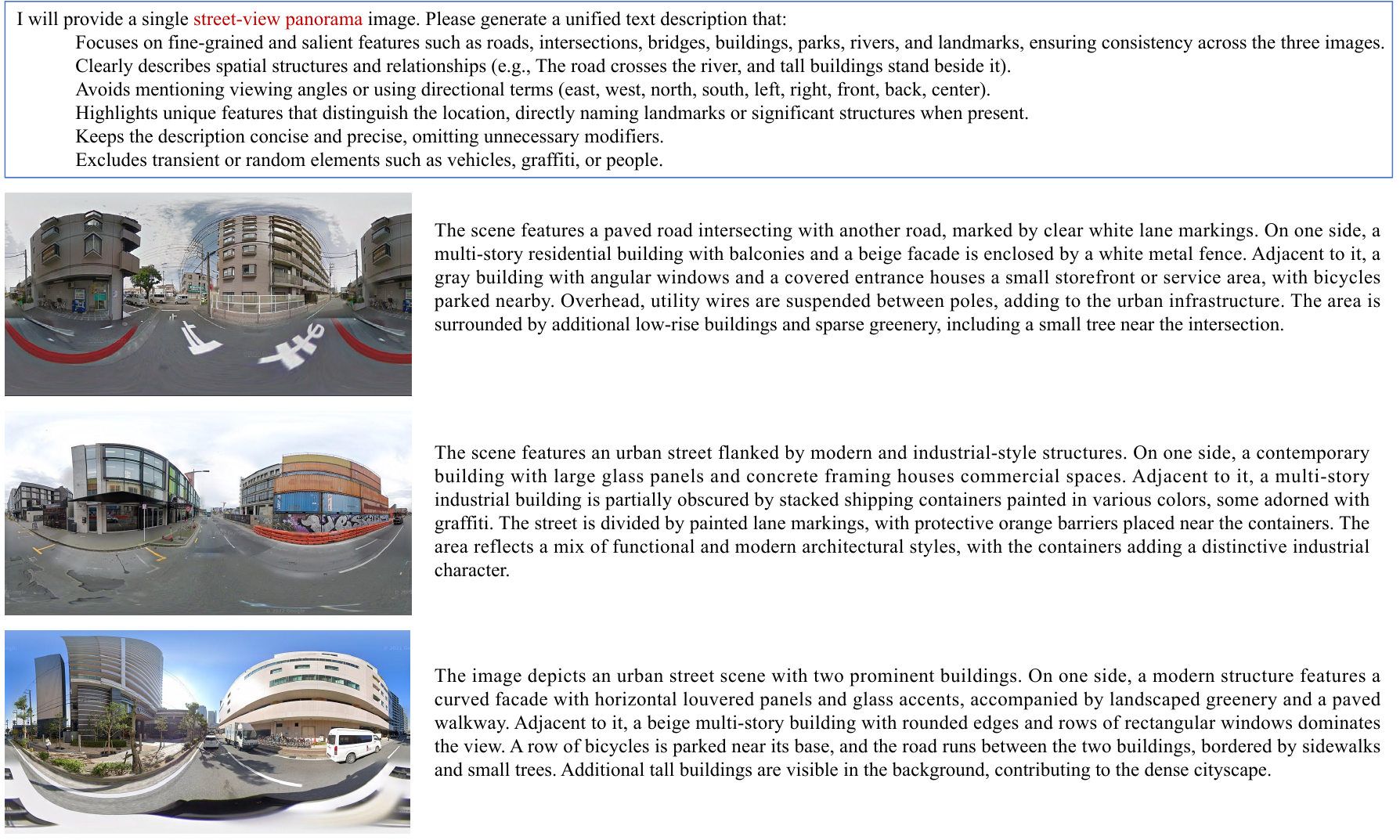}
   \caption{Cross-modal geo-location street-panorama image description instructions, with blue text boxes indicating the prompts.}
   \label{fig:street-des}
    \vspace{-13pt}
\end{figure*}

\begin{figure*}[h]
  \centering
   \includegraphics[width=1\linewidth]{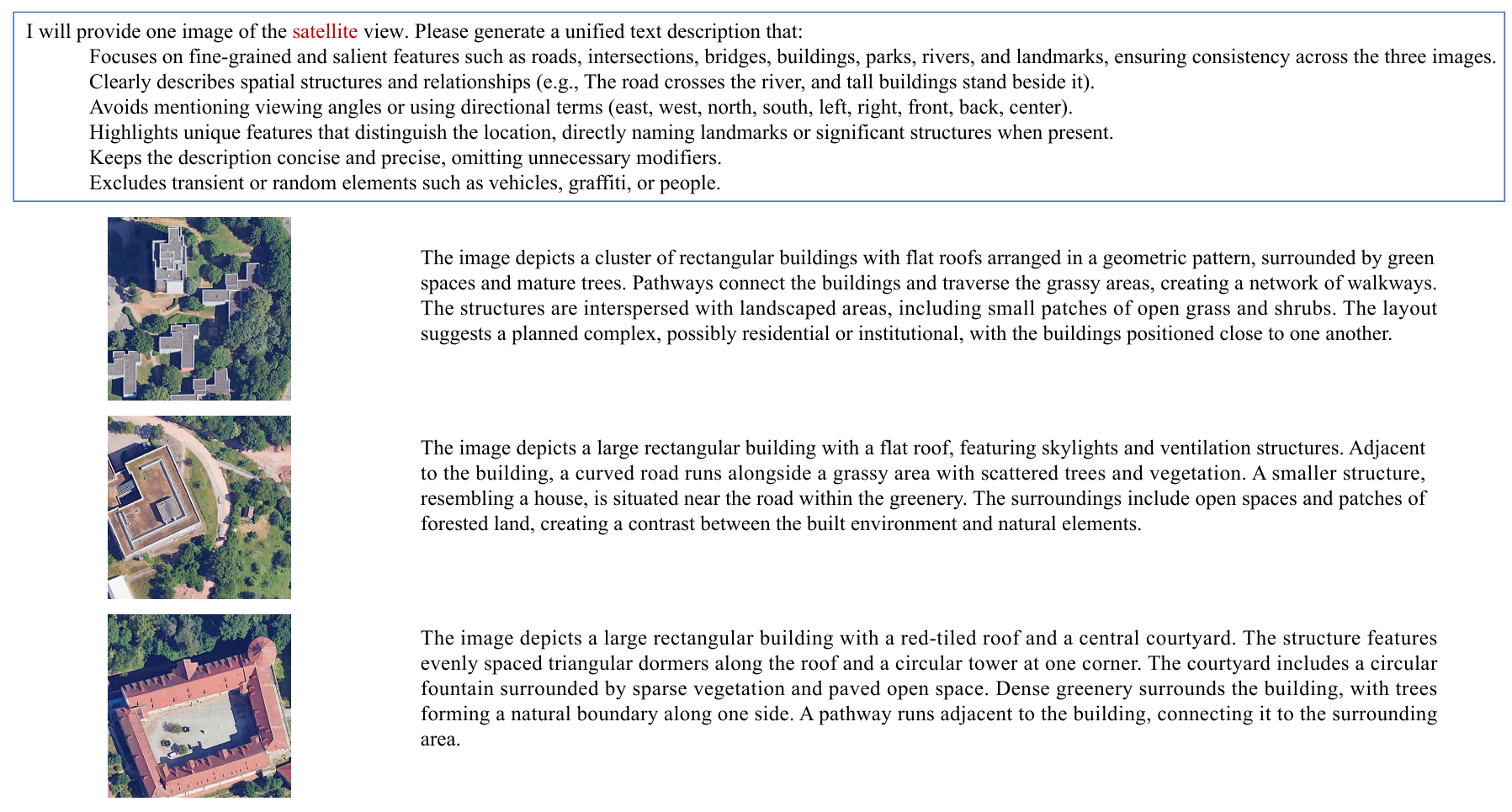}
   \caption{Cross-modal geo-location satellite image description instructions, with blue text boxes indicating the prompts.}
   \label{fig:sate-des}
    \vspace{-13pt}
\end{figure*}

To construct high-quality cross-view semantic descriptions for GeoLoc, we design a unified instruction protocol that guides a large language model to generate consistent, viewpoint-agnostic textual annotations for each tri-view set (drone, satellite, and street-view images). The goal of these instructions is to ensure that the resulting descriptions serve as reliable semantic anchors, capturing stable structural cues shared across views while avoiding viewpoint-specific bias. The specific details are shown in Figs.~\ref{fig:tri-view-des},~\ref{fig:drone-des},~\ref{fig:street-des}, and~\ref{fig:sate-des}.

The instruction set is crafted to ensure that the generated descriptions meet the following requirements:
\begin{itemize}
    \item Cross-view semantic consistency: the text must focus on structural elements that remain stable across drone, satellite, and street-view perspectives. These include roads, intersections, bridges, buildings, parks, rivers, land parcels, plazas, and distinctive landmarks. The model is instructed to ignore features that vary across viewpoints.
    \item Explicit modeling of spatial structures and relationships: the description must convey meaningful spatial relations (\eg, a bridge spans the river, the road curves around the building) while avoiding explicit directional terminology. Instead of using directional words such as north, south, east, west, left, right, front, or back, the text should describe structural relationships directly and unambiguously.
    \item Fine-grained but concise representation: each description is restricted to a single, compact paragraph. The instruction emphasizes the inclusion of salient structural cues without unnecessary adjectives, narrative elements, or scene-level speculation.
    \item Landmark naming when applicable: when a location contains identifiable landmarks such as notable buildings, bridges, squares, or monuments, the model is instructed to explicitly name them. This enhances the distinctiveness of the descriptions and improves performance in cross-view retrieval and localization.
    \item Exclusion of transient or unstable elements: the instruction explicitly prohibits mentioning transient objects (\eg, people, cars, bicycles, animals), weather, temporary decorations, graffiti, or other viewpoint-dependent artifacts. This ensures that descriptions remain stable over time and across viewpoints.
\end{itemize}


The above instructions also apply when generating descriptions for cross-modal geo-location.

\subsection{Fact Construction Details}

To complement the description in the main paper, we provide additional implementation details of the location fact construction used in GeoBridge++. Different from directly prompting an LLM to summarize multiple observations into a free-form caption, we first extract structured or modality-relevant attributes from each information source. These attributes are then filtered and organized according to their semantic scope before being verbalized into geographic, visual, and shared geo-visual factual descriptions. This procedure constrains the generated text to observable evidence and reduces the introduction of unsupported information.
\subsubsection{Geographic Fact Construction Details}

To construct geographically grounded factual descriptions for GeoBridge++, we first convert raw OSM records into compact and location-relevant geographic attributes. This procedure normalizes and summarizes the geographic entities within each region and then uses the resulting facts as constrained evidence for textual verbalization. The geographic fact construction follows the following principles:

\begin{itemize}
\item Normalization of geographic entities:
raw OSM annotations are mapped into a reduced set of semantically meaningful categories. For road-related elements, fine-grained highway tags are grouped into functional categories such as major roads, main roads, local roads, service roads, footpaths, and pedestrian ways. Building annotations are similarly normalized into coarse categories, including residential, apartment, commercial, industrial, retail, school, and hospital buildings. 

\item Extraction of spatial characteristics:
geometric information is used to characterize the spatial organization of geographic entities. For each road segment, its geometry is used to estimate a coarse orientation, which is categorized as east--west, north--south, diagonal, or unknown. The dominant road orientation within the region is then retained as a location-level attribute, providing complementary spatial information.

\item Compact representation of building distribution:
the number and types of buildings within each region are summarized to characterize the local built environment. Building density is discretized into four levels: none, sparse, moderate, and dense. Specifically, regions containing one or two buildings are considered sparse, those containing three to ten buildings are considered moderate, and those containing more than ten buildings are considered dense.

\item Preservation of functional geographic cues:
available amenity annotations are retained as complementary information about the functional characteristics of the location. These cues supplement the spatial and compositional information captured by roads and buildings.

\item Evidence-constrained verbalization: the extracted attributes are organized into geographic facts describing the scene, road characteristics, building distribution, and available functional facilities. The LLM is then instructed to produce a single concise English paragraph using only the provided evidence.
\end{itemize}

\begin{figure*}[t]
  \centering
   \includegraphics[width=1\linewidth]{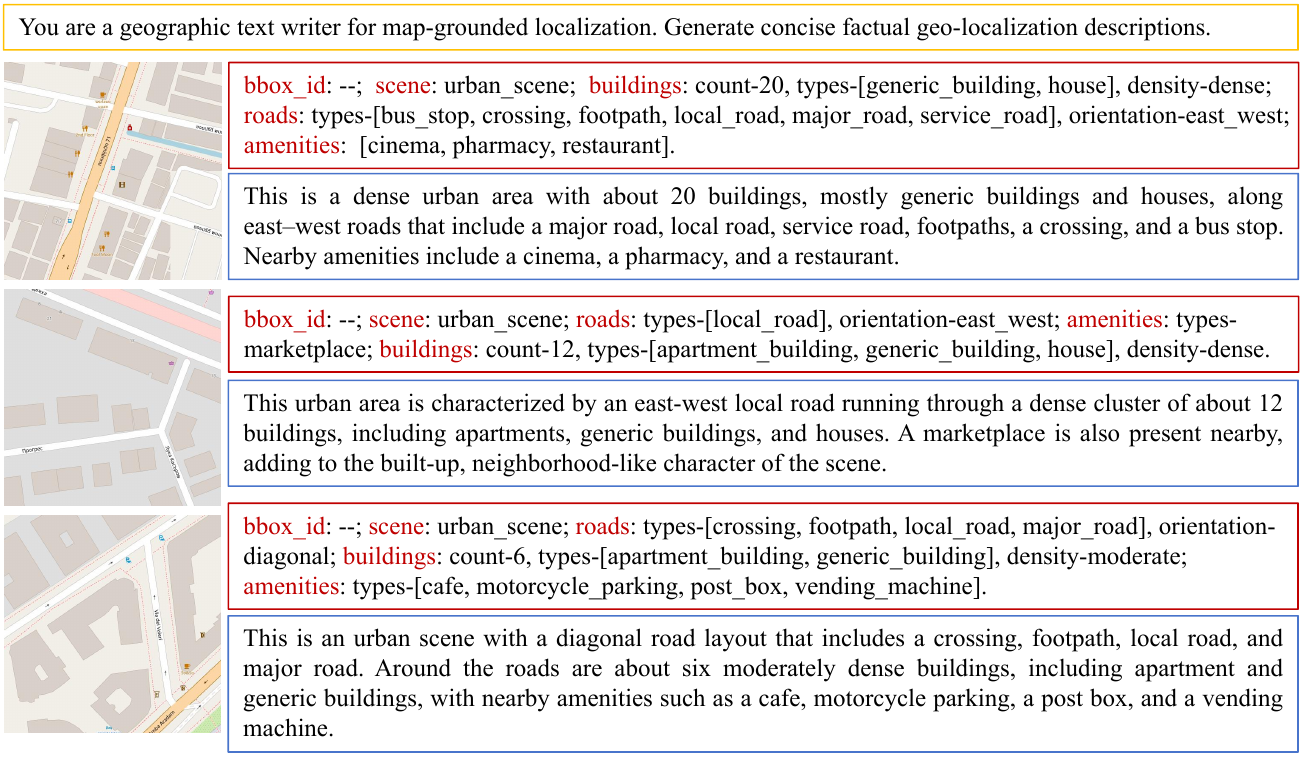}
   \caption{Examples of geographic fact construction from OSM data. Yellow boxes indicate the LLM instruction, red boxes show the extracted geographic facts, and blue boxes present the corresponding geographic factual descriptions.}
   \label{fig:geo_fact_example}
    \vspace{-13pt}
\end{figure*}

Through this process, geographic attribute extraction is explicitly separated from natural-language generation. The LLM therefore serves only to verbalize the normalized geographic evidence, while the factual content is determined by the preceding rule-based extraction and aggregation. The specific details are shown in Figure~\ref{fig:geo_fact_example}. 

\subsubsection{Visual Fact Construction Details}

Visual facts are constructed through four steps: view-specific visual attribute extraction, cross-view consensus selection, discriminative fact identification, and visual factual description composition. Rather than directly generating a description from multiple images, we first extract comparable visual attributes from each view and construct the final factual description from the resulting evidence.

\begin{itemize}

\item Independent view-specific visual attribute extraction: for each location, the street-view, drone-view, and satellite-view observations are independently provided to the LLM using the same instruction and attribute schema. The extracted attributes describe stable appearance properties, including building height and density, vegetation distribution, spatial openness, layout regularity, roof appearance, facade continuity, and surface characteristics. Temporary objects, environmental conditions, and geographic information already represented by geographic facts are explicitly excluded.

\item Cross-view consensus selection:
the attributes from the three views are compared independently for each field. Empty or unknown values are discarded, and the value supported by the largest number of views is retained as the cross-view consensus. This process preserves appearance properties that are consistently observed across different viewpoints.

\item Discriminative visual fact identification:
in addition to the consensus attributes, an LLM is used to identify a discriminative visual fact. The LLM receives the three view-specific attributes, rather than the original images, and selects a distinctive yet stable visual pattern.

\item Visual fact composition: the consensus attributes and discriminative visual fact are then provided to the LLM to generate a concise visual factual description.

\end{itemize}

This procedure separates visual evidence extraction from language composition. The details are shown in Figure~\ref{fig:visual_fact_example}.

\begin{figure*}[t]
  \centering
   \includegraphics[width=1\linewidth]{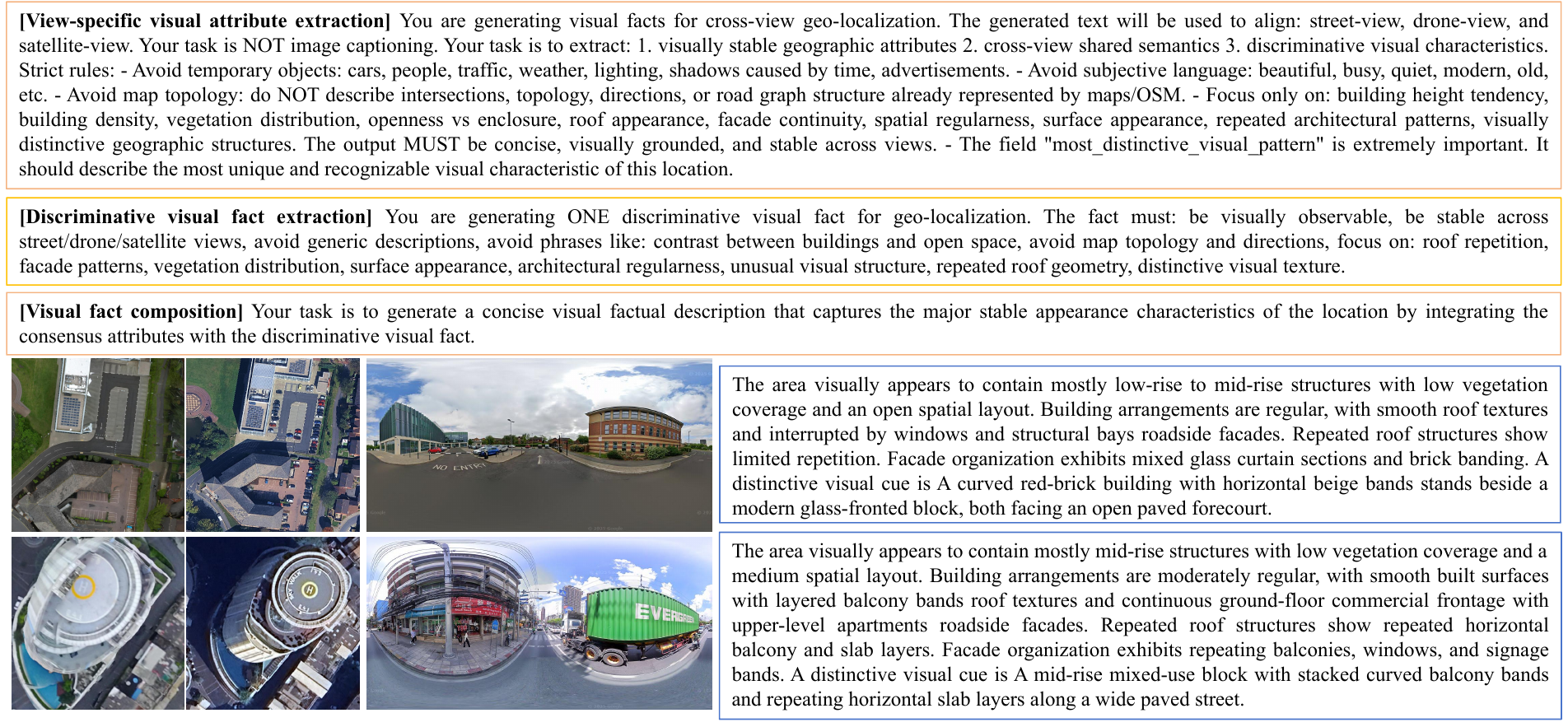}
   \caption{Examples of visual fact construction from cross-views. Yellow boxes indicate the LLM instructions, and blue boxes present the corresponding visual factual descriptions. The first column presents the drone-view image, the second column shows the satellite image, and the third column displays the street-view panorama.}
   \label{fig:visual_fact_example}
    \vspace{-13pt}
\end{figure*}

\subsubsection{Shared Geo-Visual Fact Construction Details}

Shared geo-visual facts are constructed by identifying attributes jointly supported by geographic and visual facts, followed by constrained language generation. Rather than directly prompting an LLM to merge the two factual descriptions, we first determine their common evidence at the attribute level and use the shared attributes to construct the final description. The detailed procedure is as follows:

\begin{itemize}

\item Shared attribute identification:
for each location, the geographic facts are compared with the visual facts. We focus on properties that can be supported by both sources, including scene type, building density, spatial openness, layout regularity, vegetation or open-space presence, surface characteristics, and stable structural patterns. An attribute is retained only when corresponding geographic evidence is available and is compatible with the visual observation.

\item Cross-source consistency filtering:
different attributes are examined according to their evidence. Building density is retained when the density estimated from geographic and visual facts is consistent or falls within a compatible range. Spatial openness is compared with the openness implied by the geographic building distribution. Visual layout regularity is preserved only when buildings or roads are present in the geographic facts, while vegetation, open-space, and surface characteristics require land-use, natural, amenity, or open-space evidence. Attributes supported by only one source are excluded from the shared facts.

\item Conservative cue selection:
appearance-specific details, such as roof repetition, roof texture, facade patterns, and the discriminative visual fact, are not directly treated as shared information. They are retained only when the geographic facts provide sufficient evidence of built structures and are further converted into more general structural characteristics. This prevents view-specific appearance details from being incorrectly introduced as shared properties.

\item Shared fact verbalization:
the retained attributes are provided to the LLM to generate a concise shared geo-visual factual description. The LLM is instructed to use only these filtered attributes and is not allowed to introduce additional information from the original geographic or visual descriptions. The text therefore summarizes the common properties of the location while suppressing modality-specific details, unsupported visual cues, and geographic information.

\end{itemize}

This procedure explicitly separates shared evidence identification from natural-language generation. The corresponding details are shown in Figure~\ref{fig:visual_fact_example}.

\begin{figure*}[t]
  \centering
   \includegraphics[width=1\linewidth]{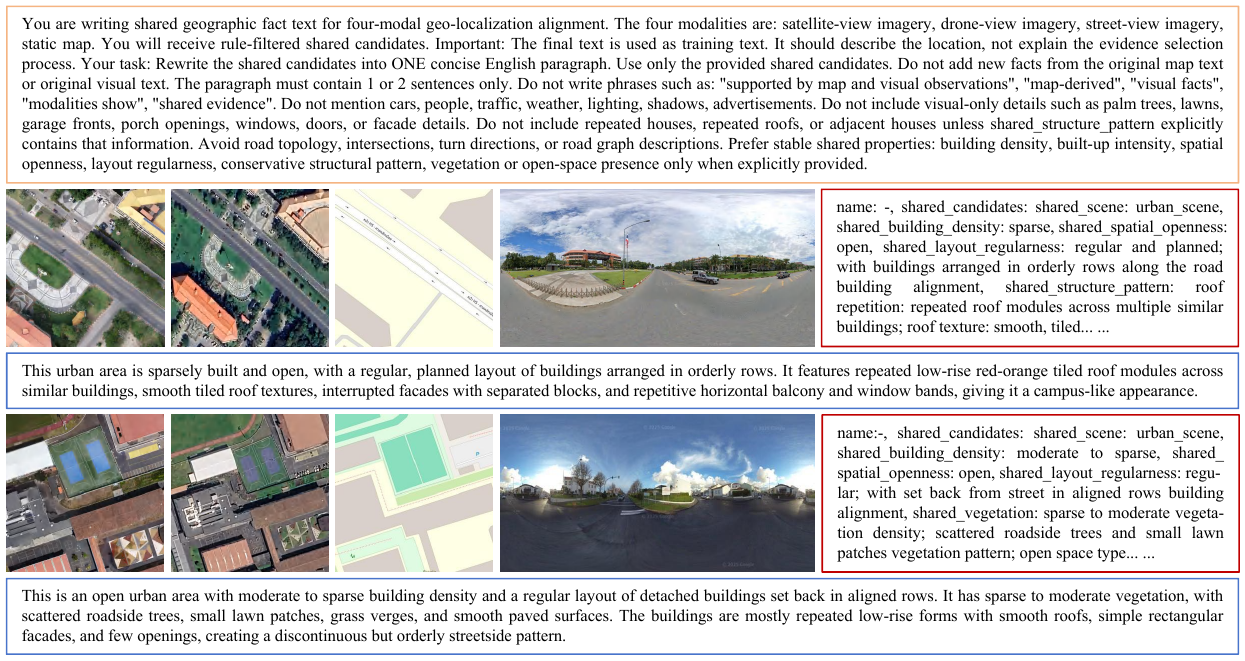}
   \caption{Examples of shared geo-visual fact construction. Yellow boxes indicate the LLM instruction, red boxes show the extracted shared geo-visual facts, and blue boxes present the corresponding shared geo-visual factual descriptions. The four images correspond to the drone, satellite, static-map, and street-view panorama views, respectively.}
   \label{fig:visual_fact_example}
    \vspace{-13pt}
\end{figure*}

\section{Additional Qualitative Results}\label{sec:examples}

\begin{figure*}[t]
  \centering
   \includegraphics[width=1\linewidth]{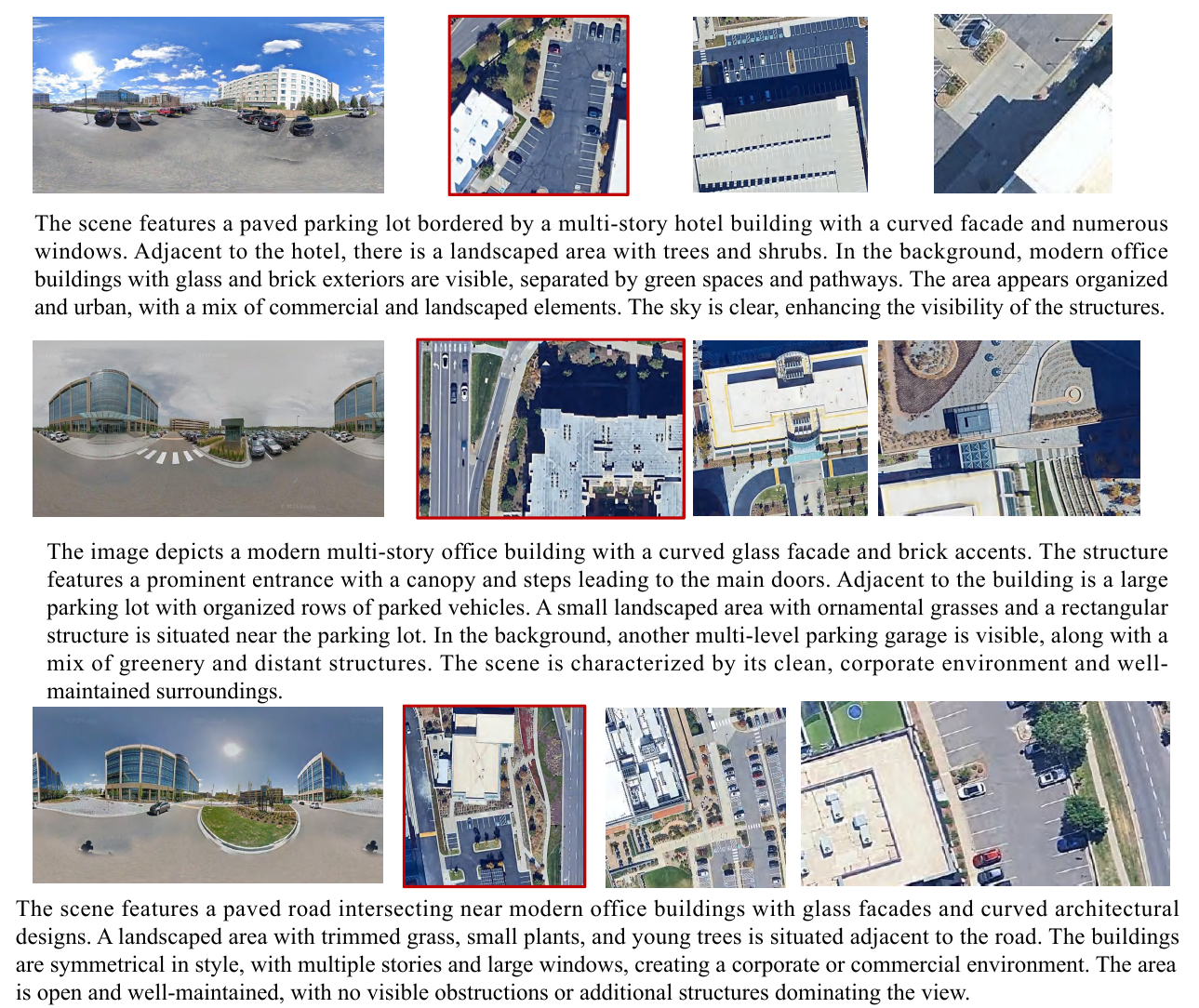}
   \caption{Qualitative results  of GeoBridge on the GeoLoc dataset for cross-modal geo-location. Using street view descriptions to match satellite perspectives, the top three results are reported; red boxes indicate correct matches.}
   \label{fig:t2s}
          \vspace{-13pt}
\end{figure*}

\begin{figure*}[t]
  \centering
   \includegraphics[width=1\linewidth]{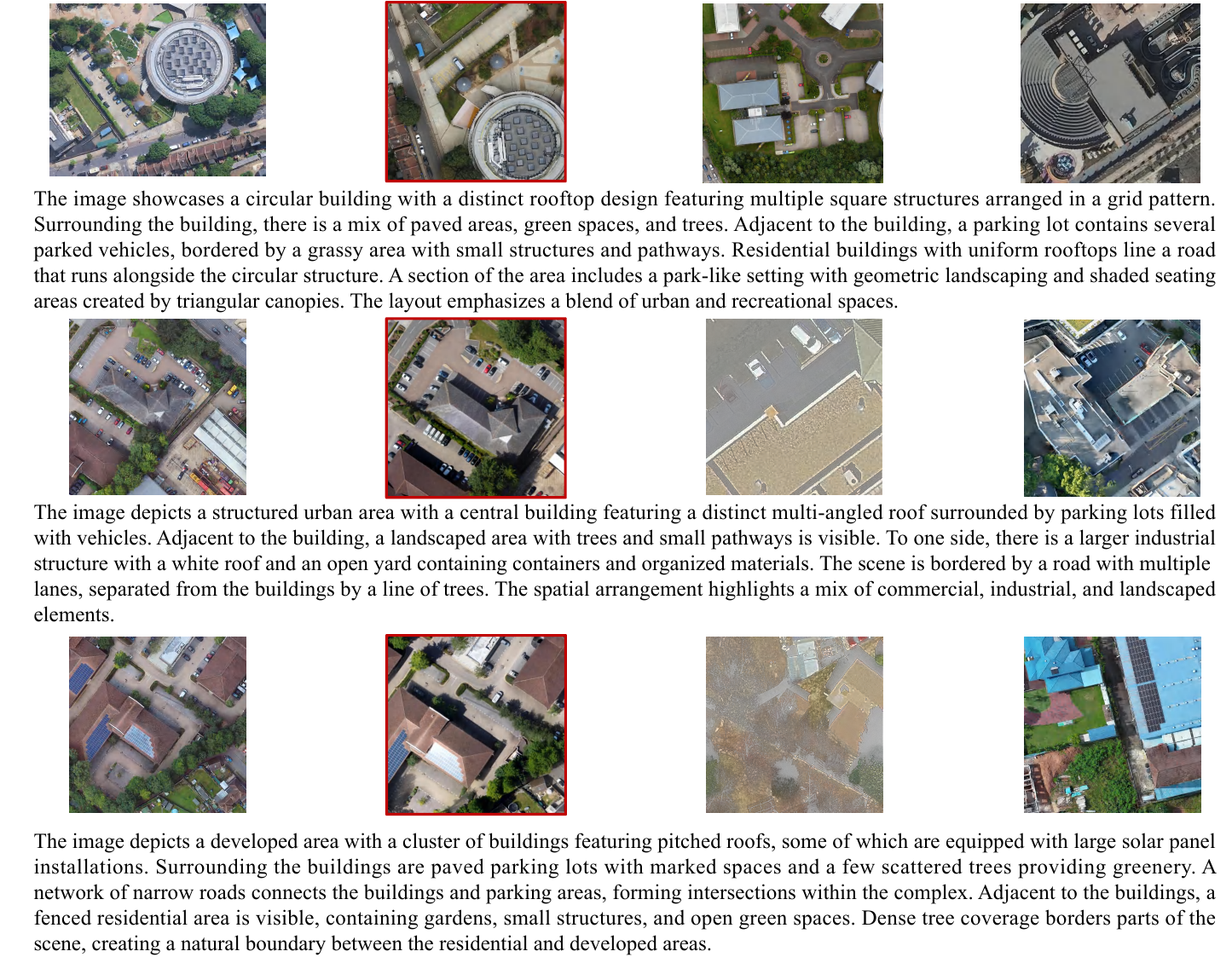}
   \caption{Qualitative results  of GeoBridge on the GeoLoc dataset for cross-modal geo-location. Using satellite view descriptions to match drone perspectives, the top three results are reported; red boxes indicate correct matches.}
   \label{fig:s2d}
    \vspace{-13pt}
\end{figure*}

\begin{figure*}[h]
  \centering
   \includegraphics[width=1\linewidth]{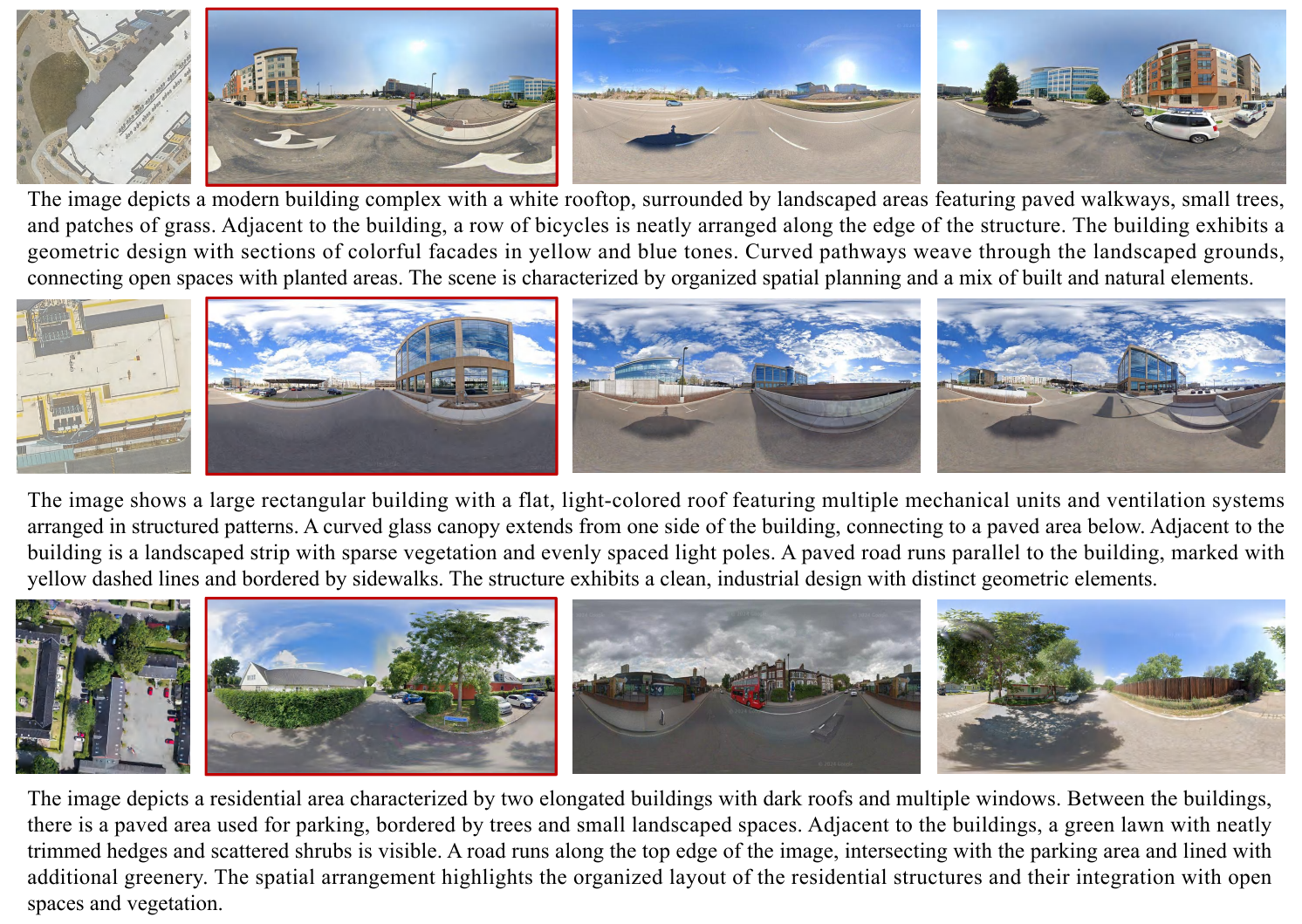}
   \caption{Qualitative results  of GeoBridge on the GeoLoc dataset for cross-modal geo-location. Using drone view descriptions to match street perspectives, the top three results are reported; red boxes indicate correct matches.}
   \label{fig:d2t}
          \vspace{-13pt}
\end{figure*}

We visualize the cross-modal geo-localization results of GeoBridge on the GeoLoc dataset. A textual description is generated from a single view and used to retrieve images from the other views. Figure~\ref{fig:t2s}, ~\ref{fig:s2d}, and ~\ref{fig:d2t} show the retrieval results for satellite, drone, and street-view images using descriptions derived from street-view, satellite, and drone images, respectively. The results indicate that GeoBridge can preserve discriminative semantic cues across viewpoints and use them to identify the corresponding locations, leading to higher rankings for the correct matches.


\begin{figure*}[t]
    \centering
    \subfloat[Drone to Satellite\label{sub1}]{%
        \includegraphics[width=0.4\textwidth]{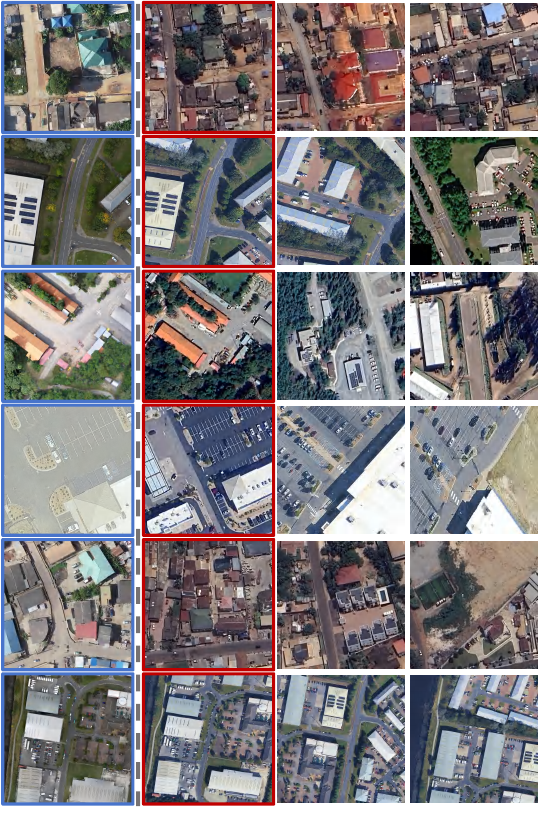}
    }
    \hspace{0.0\textwidth}
    \subfloat[Drone to Street\label{sub2}]{%
        \includegraphics[width=0.51\textwidth]{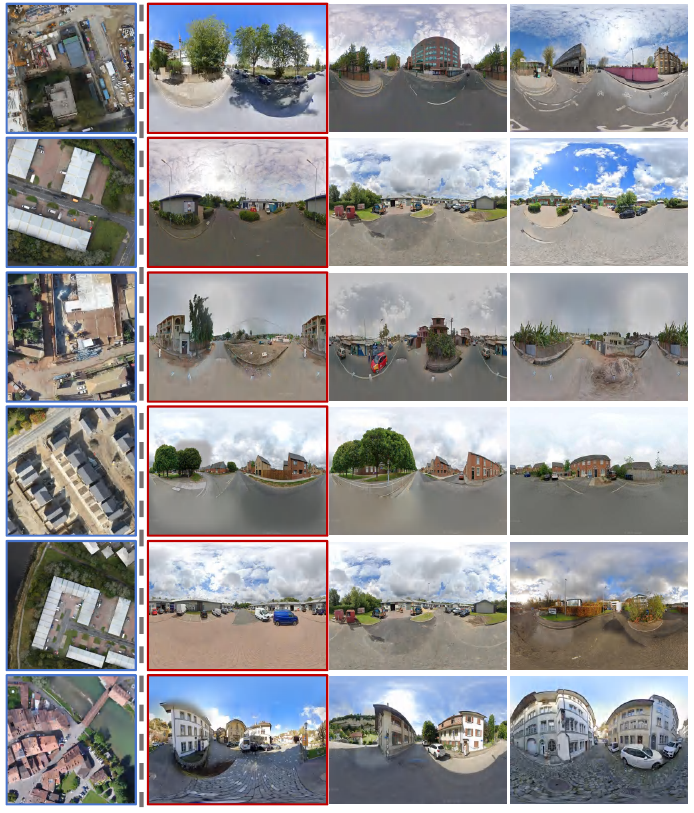}
    }
    \caption{Qualitative results of GeoBridge++ on GeoLoc-MM. Blue boxes denote queries, and red boxes indicate truth matches. The six rows from top to bottom correspond to spatial extents ranging from 80 to 200~\(m\).}
    \label{MM-d}
    \vspace{-13pt}
\end{figure*}

\begin{figure*}[t]
    \centering
    \subfloat[Satellite to Drone\label{sub1}]{%
        \includegraphics[width=0.4\textwidth]{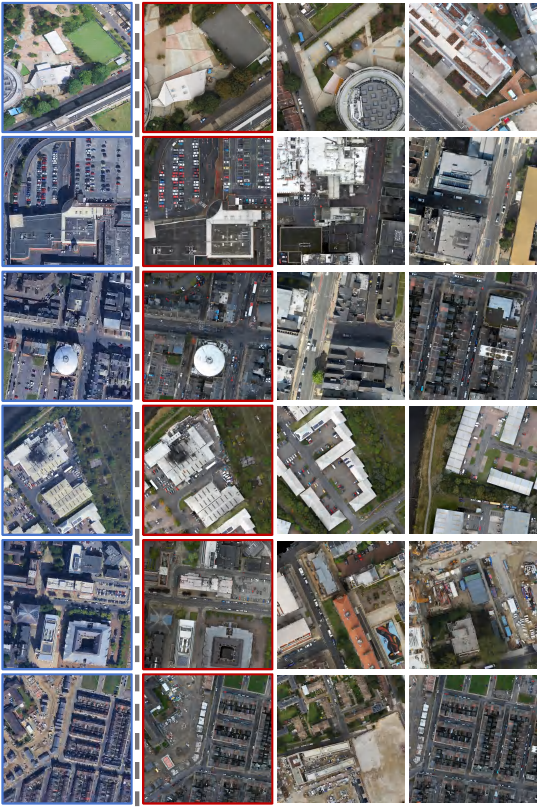}
    }
    \hspace{0.0\textwidth}
    \subfloat[Satellite to Map\label{sub2}]{%
        \includegraphics[width=0.4\textwidth]{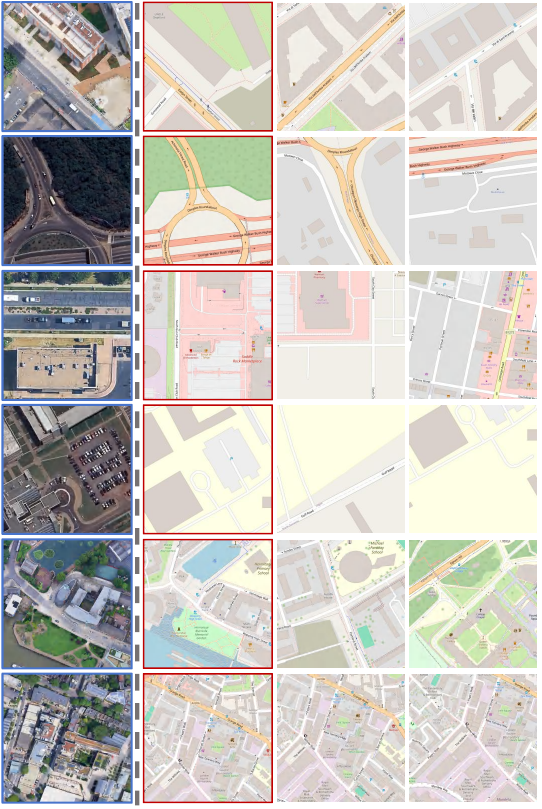}
    }
    \caption{Qualitative results of GeoBridge++ on GeoLoc-MM. Blue boxes denote queries, and red boxes indicate truth matches. The six rows from top to bottom correspond to spatial extents ranging from 80 to 200~\(m\).}
    \label{MM-s}
    \vspace{-13pt}
\end{figure*}

\begin{figure*}[t]
    \centering
    \subfloat[Street to Map\label{sub1}]{%
        \includegraphics[width=0.4\textwidth]{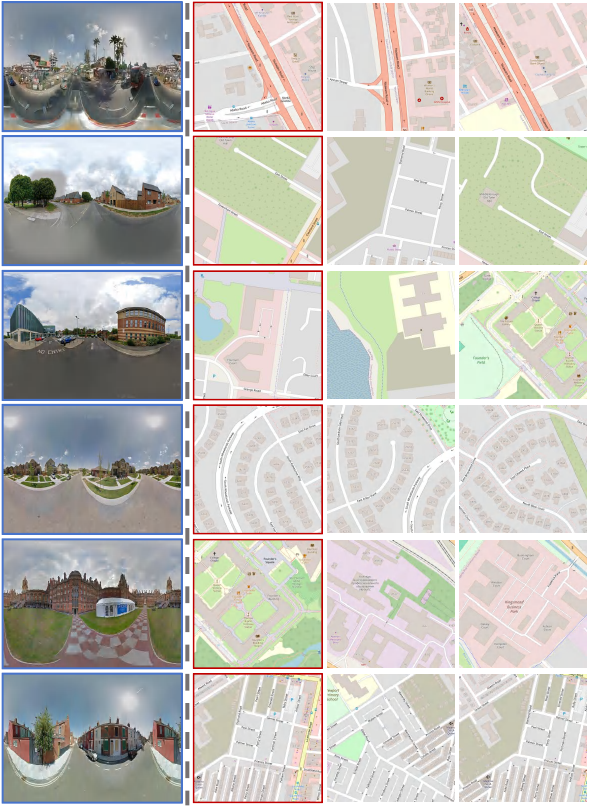}
    }
    \hspace{0.0\textwidth}
    \subfloat[Street to Satellite\label{sub2}]{%
        \includegraphics[width=0.4\textwidth]{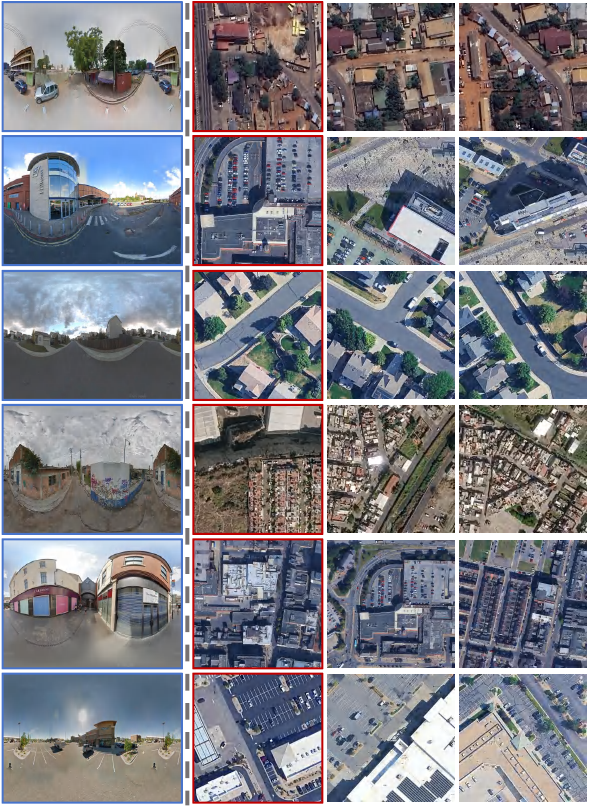}
    }
    \caption{Qualitative results of GeoBridge++ on GeoLoc-MM. Blue boxes denote queries, and red boxes indicate truth matches. The six rows from top to bottom correspond to spatial extents ranging from 80 to 200~\(m\).}
    \label{MM-t}
    \vspace{-13pt}
\end{figure*}

\begin{figure*}[t]
    \centering
    \subfloat[Map to Drone\label{sub1}]{%
        \includegraphics[width=0.405\textwidth]{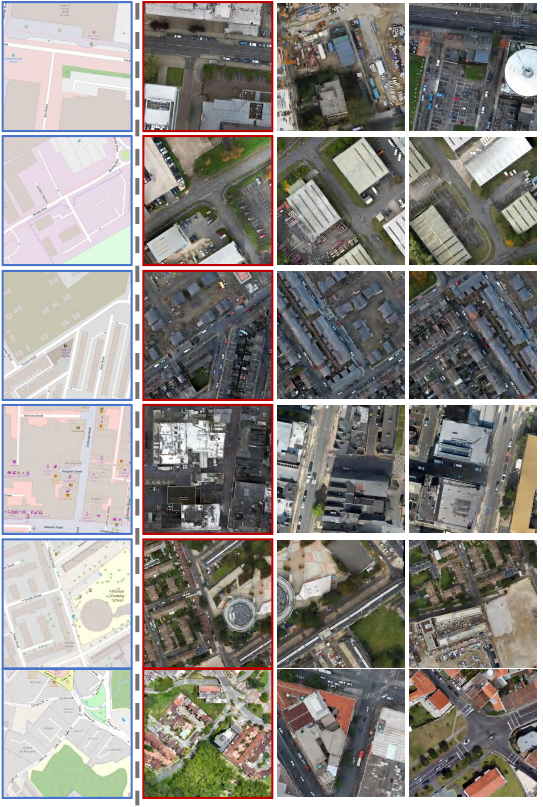}
    }
    \hspace{0.0\textwidth}
    \subfloat[Map to Street\label{sub2}]{%
        \includegraphics[width=0.51\textwidth]{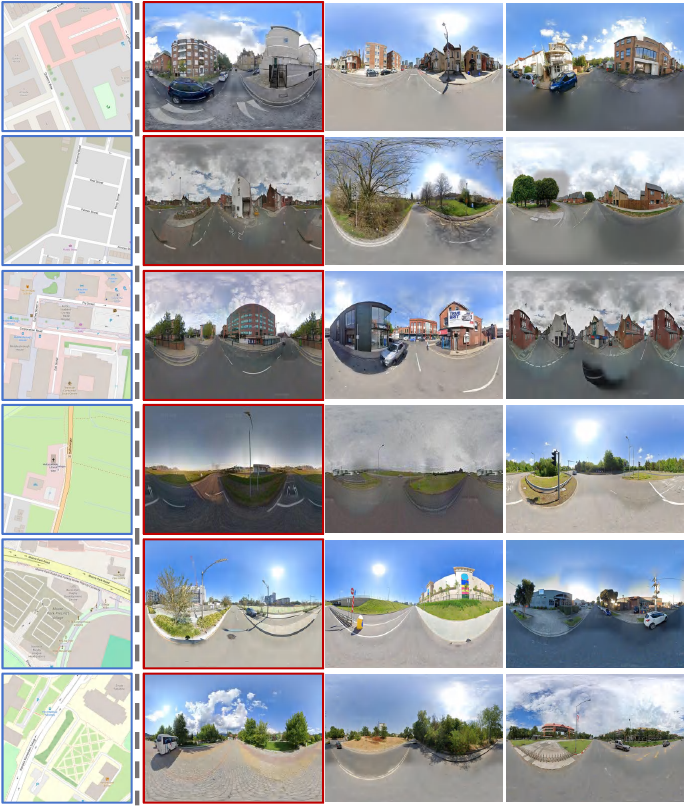}
    }
    \caption{Qualitative results of GeoBridge++ on GeoLoc-MM. Blue boxes denote queries, and red boxes indicate truth matches. The six rows from top to bottom correspond to spatial extents ranging from 80 to 200~\(m\).}
    \label{MM-m}
    \vspace{-20pt}
\end{figure*}

We further visualize cross-view image retrieval results of GeoBridge++ on the GeoLoc-MM dataset under different spatial scales. The results are shown in Figure~\ref{MM-d}, ~\ref{MM-s}, ~\ref{MM-t}, and~\ref{MM-m}. Given an image from one view as the query, GeoBridge++ retrieves geographically corresponding images from another view across drone, satellite, street-view, and static-map observations. The examples cover multiple spatial extents, illustrating the substantial variations in visible content, resolution, and scene coverage caused by both viewpoint and scale changes. Despite these differences, GeoBridge++ consistently retrieves the correct locations.

\begin{figure*}[t]
  \centering
   \includegraphics[width=0.9\linewidth]{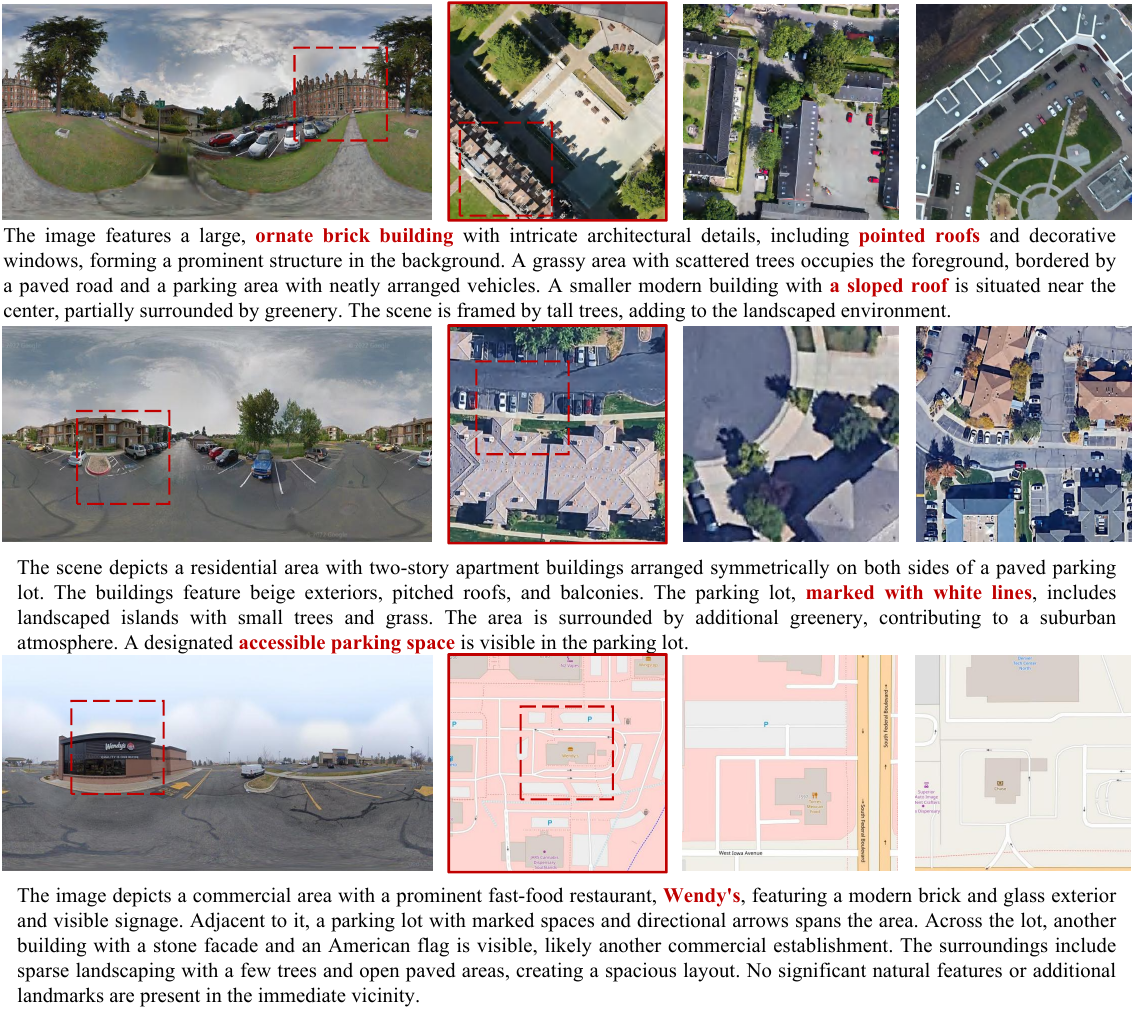}
   \caption{Qualitative cross-modal geo-localization results of GeoBridge++ on map-augmented GeoLoc. Descriptions generated from street-view observations are used to retrieve images from drone, satellite, and map perspectives. The top-three retrieved results are shown, with correct matches highlighted in red.}
   \label{des-t}
          \vspace{-13pt}
\end{figure*}

\begin{figure*}[t]
  \centering
   \includegraphics[width=0.9\linewidth]{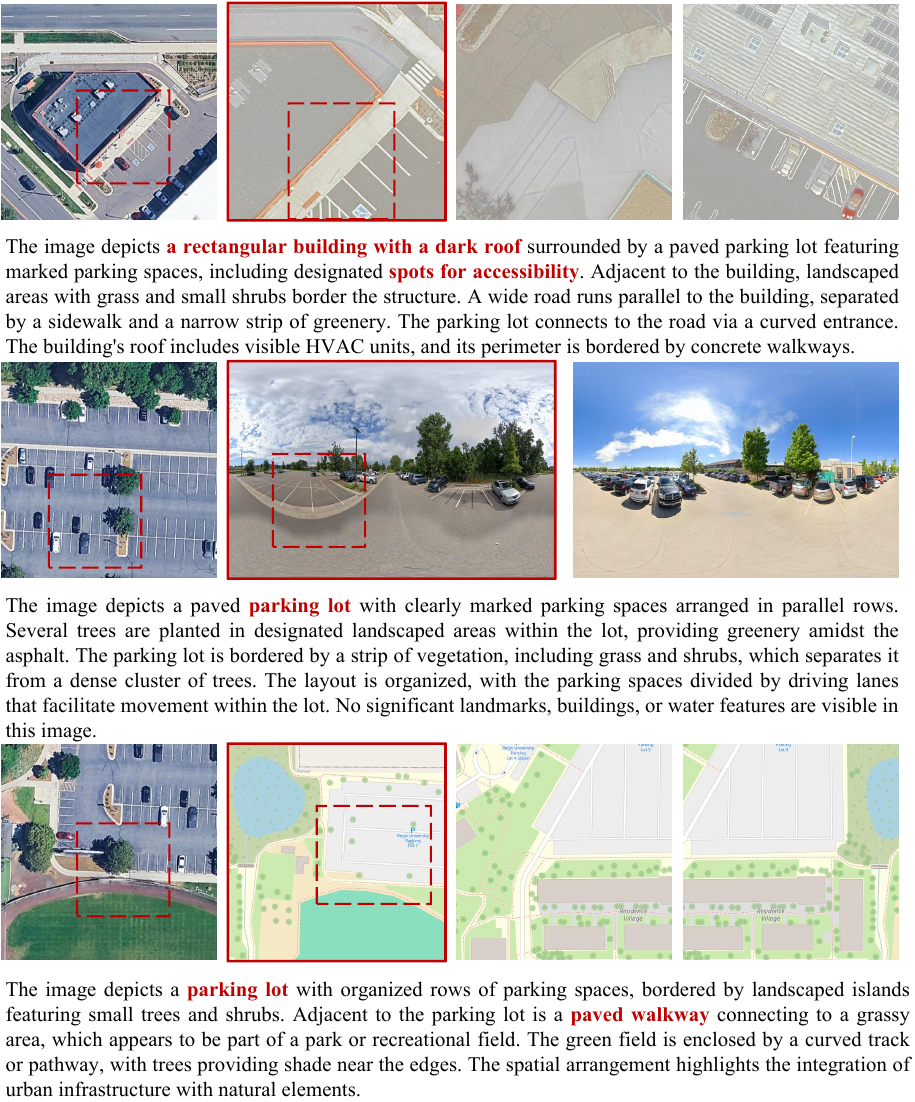}
   \caption{Qualitative cross-modal geo-localization results of GeoBridge++ on map-augmented GeoLoc. Descriptions generated from satellite observations are used to retrieve images from drone, street-view, and map perspectives. The top-ranked retrieved results are shown, with correct matches highlighted in red.}
   \label{des-s}
          \vspace{-13pt}
\end{figure*}

\begin{figure*}[t]
  \centering
   \includegraphics[width=0.9\linewidth]{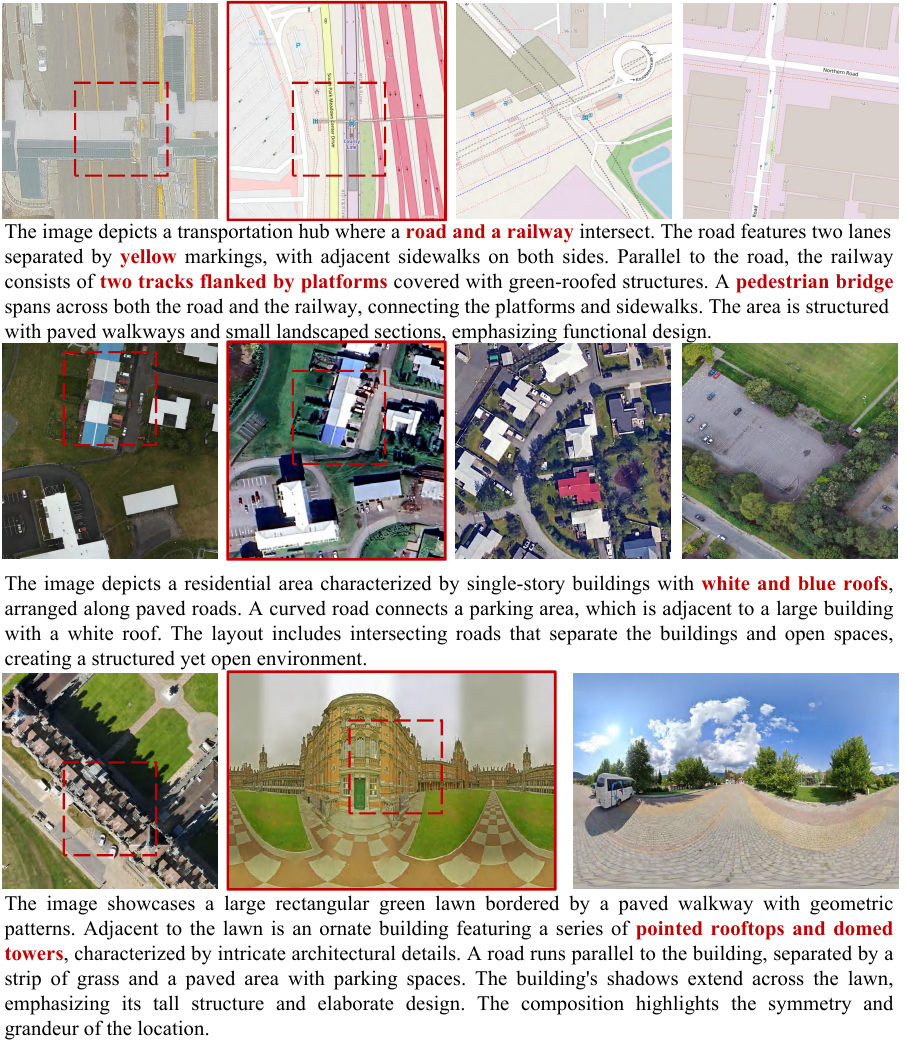}
   \caption{Qualitative cross-modal geo-localization results of GeoBridge++ on map-augmented GeoLoc. Descriptions generated from drone observations are used to retrieve images from map, satellite, and street-view perspectives. The top-ranked retrieved results are shown, with correct matches highlighted in red.}
   \label{des-d}
          \vspace{-13pt}
\end{figure*}

We visualize the cross-modal geo-localization results of GeoBridge++ on the map-augmented GeoLoc dataset. In this setting, a textual description generated from one observation is used to retrieve images from the other views. Fig.~\ref{des-t}, Fig.~\ref{des-s}, and Fig.~\ref{des-d} present the retrieval results using descriptions derived from street-view, satellite, and drone observations, respectively. Each description is used to retrieve the geographically corresponding observations from the remaining three views. The results demonstrate that GeoBridge++ effectively captures location-relevant semantics from textual descriptions and transfers them across heterogeneous viewpoints and representation forms.

\section{Datasheets}
\label{sec:datasheets}

In this section, we document essential details about the GeoLoc and GeoLoc-MM datasets following standard dataset documentation practices and the template provided by Gebru \textit{et al.} \cite{gebru2021datasheets}.

\subsection{Motivation}
The questions in this section are primarily intended to encourage dataset creators to clearly articulate their reasons for creating the dataset and to promote transparency about funding interests. The latter may be particularly relevant for datasets created for research purposes.
\begin{enumerate}
    \item \textit{``For what purpose was the dataset created?''}
    
    \textcolor{BurntOrange}{\textbf{A:}} 
    GeoLoc and GeoLoc-MM were created to facilitate unified cross-view geo-localization across heterogeneous viewpoints, representation forms, and spatial scales. Existing benchmarks are predominantly constructed around pairwise satellite-centric settings and therefore provide limited support for studying general correspondence among aerial, ground-level, symbolic, and language representations of the same geographic location. GeoLoc establishes a multi-view benchmark by jointly considering drone, satellite, and street-view observations together with textual descriptions. GeoLoc-MM further extends this setting by incorporating static-map observations, substantially increasing the dataset scale, and introducing multiple spatial extents for aerial and map-related views. Together, the two datasets support the study of cross-view retrieval, cross-modal geo-localization, multi-scale geographic representation learning, and generalization across heterogeneous geographic observations.
    
    \item \textit{``Who created the dataset (\textit{e.g.}, which team, research group) and on behalf of which entity?''}
    
    \textcolor{BurntOrange}{\textbf{A:}} The dataset was jointly created by all authors of this work.
    
    \item \textit{``Who funded the creation of the dataset?''}
    
    \textcolor{BurntOrange}{\textbf{A:}}
    The dataset creation was funded by the affiliations of the authors involved in this work.
\end{enumerate}

\subsection{Composition}
Most of the questions in this section are intended to provide dataset consumers with the information they need to make informed decisions about using the dataset for their chosen tasks. Some of the questions are designed to elicit information about compliance with the EU’s General Data Protection Regulation (GDPR) or comparable regulations in other jurisdictions. Questions that apply only to datasets that relate to people are grouped together at the end of the section. We recommend taking a broad interpretation of whether a dataset relates to people. For example, any dataset containing text that was written by people relates to people.
\begin{enumerate}
    \item \textit{``What do the instances that comprise our datasets represent (\textit{e.g.}, documents, photos, people, countries)?''}
    
    \textcolor{BurntOrange}{\textbf{A:}} 
     Each instance is organized around a real-world geographic location shared by observations from different viewpoints and representation forms. In GeoLoc, each location contains a drone-view image, a satellite image, a street-view panorama, and an associated textual description. GeoLoc-MM extends this organization by additionally incorporating a static-map view and observations at multiple spatial extents. All observations associated with the same location maintain geographic correspondence, while their visible content, spatial coverage, and representation form may differ substantially across views and scales. GeoLoc-MM further provides geographic, visual, and shared geo-visual factual descriptions that characterize complementary properties of each location.
    
    \item \textit{``How many instances are there in total (of each type, if appropriate)?''}
    
    \textcolor{BurntOrange}{\textbf{A:}} GeoLoc contains 52,679 geographically aligned location instances, including 47,328 instances for training and validation and 5,351 instances from geographically non-overlapping cities for evaluation. Each location is associated with aligned drone, satellite, and street-view observations, together with a textual description. GeoLoc-MM substantially extends this benchmark to 1,000,901 images organized into approximately 316,000 geographic groups. It comprises four observation types---drone, satellite, street-view, and static map---and spans six spatial scales of 80, 100, 120, 150, 180, and 200\,m, enabling systematic multi-scale evaluation.

    \item \textit{``Does the dataset contain all possible instances or is it a sample (not necessarily random) of instances from a larger set?''}
    
    \textcolor{BurntOrange}{\textbf{A:}} Both datasets are curated subsets of a substantially larger collection of geographically referenced imagery. Candidate locations are retained only when the required observations can be reliably acquired, geographically aligned, and validated according to the corresponding dataset construction criteria. GeoLoc focuses on reliable correspondence among drone, satellite, and street-view observations, whereas GeoLoc-MM applies an extended collection procedure to support four-view and multi-scale geographic correspondence. 
    
    \item \textit{``Is there a label or target associated with each instance?''}
    
    \textcolor{BurntOrange}{\textbf{A:}} Yes. The fundamental target in both datasets is geographic correspondence. Observations belonging to the same location share a common geographic identity and therefore constitute positive cross-view matches.
    
    \item \textit{``Is any information missing from individual instances?''}
    
    \textcolor{BurntOrange}{\textbf{A:}} No, each individual instance is complete.
    
    \item \textit{``Are relationships between individual instances made explicit (\textit{e.g.}, users’ movie ratings, social network links)?''}
    
    \textcolor{BurntOrange}{\textbf{A:}} Yes. The primary relationship is geographic correspondence. Observations assigned to the same location identity represent the same geographic region from different viewpoints, scales, or representation forms. Observations associated with different location identities are treated as geographically distinct samples. 
    
    \item \textit{``Are there recommended data splits (\textit{e.g.}, training, development/validation, testing)?''}
    
    \textcolor{BurntOrange}{\textbf{A:}} 
    Yes. We provide predefined training, validation, and evaluation protocols for both datasets. GeoLoc uses 47,328 instances for training and validation and reserves 5,351 instances from geographically non-overlapping cities for evaluation. GeoLoc-MM follows the same geographic split as GeoLoc, while extending each location with additional modalities and multi-scale observations.
    
    \item \textit{``Is the dataset self-contained, or does it link to or otherwise rely on external resources (\textit{e.g.}, websites, tweets, other datasets)?''}
    
    \textcolor{BurntOrange}{\textbf{A:}} The dataset draws on external sources for the street-view and satellite imagery, specifically Google Street View and Google Maps Satellite. 
    
    \item \textit{``Does the dataset contain data that might be considered confidential (\textit{e.g.}, data that is protected by legal privilege or by doctor–patient confidentiality, data that includes the content of individuals’ non-public communications)?''}
    
    \textcolor{BurntOrange}{\textbf{A:}} No. It does not include personal or confidential information. The Google Street View and Google Satellite images used in the dataset are subject to Google’s terms of service and licensing policies, and their use follows the corresponding Google usage agreements.
    
    \item \textit{``Does the dataset contain data that, if viewed directly, might be offensive, insulting, threatening, or might otherwise cause anxiety?''}
    
    \textcolor{BurntOrange}{\textbf{A:}} No, GeoLoc and GeoLoc-MM do not contain any data with negative information.
\end{enumerate}

\subsection{Collection Process}
In addition to the goals outlined in the previous section, the questions in this section are designed to elicit information that may help researchers and practitioners create alternative datasets with similar characteristics. Again, questions that apply only to datasets that relate to people are grouped together at the end of the section.
\begin{enumerate}
    \item \textit{``How was the data associated with each instance acquired?''}
    
    \textcolor{BurntOrange}{\textbf{A:}} 
    Both datasets are constructed around geo-tagged locations. Drone-view images with geographic metadata are first associated with their corresponding locations. Street-view panoramas and satellite images are then acquired according to the same geographic coordinates. GeoLoc-MM additionally obtains static-map observations and constructs aerial and map-related observations at multiple spatial extents. GeoLoc provides textual descriptions for the aligned visual observations, whereas GeoLoc-MM further derives geographic, visual, and shared geo-visual facts from complementary geographic and visual sources.
    
    \item \textit{``What mechanisms or procedures were used to collect the data (\textit{e.g.}, hardware apparatuses or sensors, manual human curation, software programs, software APIs)?''}
    
    \textcolor{BurntOrange}{\textbf{A:}} Drone imagery was obtained from OpenAerialMap, while street-view and satellite observations were acquired through Google mapping services. Static-map observations in GeoLoc-MM were derived from OpenStreetMap. The pipeline also included manual curation and automated procedures for cropping, alignment, filtering, and GPT-4.0–based text generation.
    
    \item \textit{``If the dataset is a sample from a larger set, what was the sampling strategy (\textit{e.g.}, deterministic, probabilistic with specific sampling probabilities)?''} 
    
    \textcolor{BurntOrange}{\textbf{A:}} The datasets are constructed using a curated rather than purely random sampling strategy. Candidate locations are selected according to data availability, geographic alignment, image quality, spatial overlap, and modality-specific requirements. Detailed sampling, filtering, and construction procedures are provided in the corresponding dataset sections of the main paper.
\end{enumerate}

\subsection{Preprocessing, Cleaning, and Labeling}
The questions in this section are intended to provide dataset
consumers with the information they need to determine whether the “raw” data has been processed in ways that are compatible with their chosen tasks. For example, text that has been converted into a ``bag-of-words" is not suitable for tasks involving word order.
\begin{enumerate}
    \item \textit{``Was any preprocessing/cleaning/labeling of the data done (\textit{e.g.}, discretization or bucketing, tokenization, part-of-speech tagging, SIFT feature extraction, removal of instances, processing of missing values)?''}
    
    \textcolor{BurntOrange}{\textbf{A:}} Yes. Both datasets undergo systematic preprocessing and quality control to ensure reliable geographic correspondence. The processing pipeline includes image-quality filtering, removal of invalid or heavily degraded samples, spatial deduplication, coordinate-based alignment, and view-specific cropping. GeoLoc retains only locations for which drone, satellite, and street-view observations can be reliably aligned. GeoLoc-MM extends the same principle to static-map observations and multiple spatial extents, while maintaining a common geographic identity across views and scales. For semantic annotation, GeoLoc constructs textual descriptions under a controlled language-model instruction protocol. GeoLoc-MM further constructs geographic facts from geographic topology, visual facts from observable scene attributes, and shared geo-visual facts from information consistently supported by complementary sources.

    \item \textit{``Was the `raw' data saved in addition to the preprocessed/cleaned/labeled data (\textit{e.g.}, to support unanticipated future uses)?''} 
    
    \textcolor{BurntOrange}{\textbf{A:}} Yes, raw data is accessible.
    
    \item \textit{``Is the software that was used to preprocess/clean/label the data available?''} 
    
    \textcolor{BurntOrange}{\textbf{A:}} Yes, the necessary software used to preprocess and clean the data is publicly available.
\end{enumerate}

\subsection{Uses}
The questions in this section are intended to encourage dataset creators to reflect on tasks for which the dataset should and should not be used. By explicitly highlighting these tasks, dataset creators can help dataset consumers make informed decisions, thereby avoiding potential risks or harms.
\begin{enumerate}
    \item \textit{``Has the dataset been used for any tasks already?''} 
    
    \textcolor{BurntOrange}{\textbf{A:}} 
    No.
    
    \item \textit{``Is there a repository that links to any or all papers or systems that use the dataset?''} 
    
    \textcolor{BurntOrange}{\textbf{A:}} Yes. The GeoLoc dataset and its associated resources are available at \url{https://github.com/MiliLab/GeoBridge}. The repository for GeoLoc-MM, together with the corresponding benchmark resources and related links, will be released after publication.
    
    \item \textit{``What (other) tasks could the dataset be used for?''} 
    
    \textcolor{BurntOrange}{\textbf{A:}} Beyond cross-view geo-localization, GeoLoc and GeoLoc-MM can support research on multi-view representation learning, cross-modal retrieval, vision--language grounding, multi-scale scene understanding, aerial--ground correspondence, map--image matching, geographic semantic representation learning, UAV navigation, urban understanding, and multimodal geographic foundation models. GeoLoc-MM is particularly suitable for studying representation gaps between natural imagery and symbolic maps, as well as consistency across different spatial extents.
    
    \item \textit{``Is there anything about the composition of the dataset or the way it was collected and preprocessed/cleaned/labeled that might impact future uses?''} 
    
    \textcolor{BurntOrange}{\textbf{A:}} No.
    
    \item \textit{``Are there tasks for which the dataset should not be used?''} 
    
    \textcolor{BurntOrange}{\textbf{A:}} N/A.
\end{enumerate}

\subsection{Distribution}
Dataset creators should provide answers to these questions prior to distributing the dataset either internally within the entity on behalf of which the dataset was created or externally to third parties.
\begin{enumerate}
    \item \textit{``Will the dataset be distributed to third parties outside of the entity (\textit{e.g.}, company, institution, organization) on behalf of which the dataset was created?''} 
    
    \textcolor{BurntOrange}{\textbf{A:}} No. The datasets will be made publicly accessible to the research community.
    
    \item \textit{``How will the dataset be distributed (\textit{e.g.}, tarball on website, API, GitHub)?''} 
    
    \textcolor{BurntOrange}{\textbf{A:}} We will provide in the GitHub and the Huggingface repository.
    
    \item \textit{``When will the dataset be distributed?''} 
    
    \textcolor{BurntOrange}{\textbf{A:}} GeoLoc is already publicly available, while GeoLoc-MM will be released through the project repository after the official publication of this work.
    
    \item \textit{``Will the dataset be distributed under a copyright or other intellectual property (IP) license, and/or under applicable terms of use (ToU)?''} 
    
    \textcolor{BurntOrange}{\textbf{A:}} Yes, the dataset will be released under the Creative Commons Attribution-NonCommercial-ShareAlike 4.0 International License.
    
    \item \textit{``Have any third parties imposed IP-based or other restrictions on the data associated with the instances?''} 
    
    \textcolor{BurntOrange}{\textbf{A:}} Yes. The street-view and satellite images included in the dataset originate from Google Street View and Google Maps Satellite, which are subject to Google’s terms of service and usage policies. These modalities must be used in compliance with Google’s licensing restrictions. 
    
    \item \textit{``Do any export controls or other regulatory restrictions apply to the dataset or to individual instances?''} 
    
    \textcolor{BurntOrange}{\textbf{A:}} No.    
\end{enumerate}

\subsection{Maintenance}
As with the questions in the previous section, dataset creators should provide answers to these questions prior to distributing the dataset. The questions in this section are intended to encourage dataset creators to plan for dataset maintenance and communicate this plan to dataset consumers.
\begin{enumerate}
    \item \textit{``Who will be supporting/hosting/maintaining the dataset?''} 
    
    \textcolor{BurntOrange}{\textbf{A:}} The authors of this work serve to support, host, and maintain the datasets.
    
    \item \textit{``How can the owner/curator/manager of the dataset be contacted (\textit{e.g.}, email address)?''} 
    
    \textcolor{BurntOrange}{\textbf{A:}} The curators can be contacted via the email addresses listed on our paper or webpage.
    
    \item \textit{``Is there an erratum?''} 
    
    \textcolor{BurntOrange}{\textbf{A:}} There is no explicit erratum; updates and known errors will be specified in future versions.
    
    \item \textit{``Will the dataset be updated (\textit{e.g.}, to correct labeling errors, add new instances, delete instances)?''} 
    
    \textcolor{BurntOrange}{\textbf{A:}} Future updates (if any) will be posted on the dataset website.
    
    \item \textit{``Will older versions of the dataset continue to be supported/hosted/maintained?''} 
    
    \textcolor{BurntOrange}{\textbf{A:}} 
    Yes.
    
    \item \textit{``If others want to extend/augment/build on/contribute to the dataset, is there a mechanism for them to do so?''} 
    
    \textcolor{BurntOrange}{\textbf{A:}} Yes, we will provide detailed instructions for future extensions.
\end{enumerate}

\section{Limitation and Potential Societal Impact}
\label{sec:app-limitation}
In this section, we discuss the limitations and potential societal impact of this work.

\subsection{Potential Limitations}

Despite their scale and multi-view design, GeoLoc and GeoLoc-MM have several inherent limitations. First, their geographic coverage depends on the availability of street-view, satellite, and mapping services, which may result in uneven representation across geographic regions and underrepresentation of areas with limited mapping infrastructure. Second, observations from different modalities may be acquired at different times, introducing temporal discrepancies caused by changes in buildings, vegetation, roads, or other environmental conditions. Third, the quality-control and filtering procedures remove severely degraded or invalid observations, which improves correspondence reliability but may reduce the diversity of extremely challenging conditions. Fourth, the drone imagery is collected under a finite set of acquisition configurations and therefore cannot fully represent the variability of UAV platforms, sensors, altitudes, and trajectories encountered in practical applications. In GeoLoc-MM, differences in spatial scale and representation form further increase the heterogeneity among observations, particularly between natural imagery and symbolic maps. Finally, textual descriptions and factual annotations are constructed with language-model assistance under controlled instructions. Although the construction process is designed to emphasize observable and geographically relevant information, residual inaccuracies, stylistic regularities, or semantic biases may remain.

\subsection{Potential Negative Societal Impact}

GeoLoc and GeoLoc-MM, similar to other geo-referenced vision datasets, may introduce risks if used outside their intended research scope. Because the datasets associate visual observations with real-world geographic locations, models trained on them could potentially be adapted for unauthorized geo-identification, invasive tracking, or surveillance-related applications. The combination of aerial, street-level, satellite, and map observations in GeoLoc-MM may further expose structural or environmental information about a location that is not apparent from a single modality alone. In addition, geographic coverage inherited from external mapping services may introduce regional biases that could affect model performance when deployed in underrepresented areas. We therefore encourage responsible use of both datasets for academic research, with careful consideration of privacy, geographic sensitivity, local regulations, and the licensing requirements of all underlying data providers.
\ifCLASSOPTIONcaptionsoff
  \newpage
\fi

\end{document}